\documentclass{article}

\usepackage[preprint]{neurips_2026}

\usepackage[utf8]{inputenc} 
\usepackage[T1]{fontenc}    
\usepackage{hyperref}       
\usepackage{url}            
\usepackage{booktabs}       
\usepackage{amsfonts}       
\usepackage{nicefrac}       
\usepackage{microtype}      
\microtypesetup{nopatch=footnote}
\usepackage{xcolor}         

\usepackage{linguex}
\usepackage{fontawesome}
\usepackage{amsmath}
\usepackage{marvosym}
\usepackage{minibox}
\usepackage{longtable}

\usepackage{pdflscape}

\usepackage{graphicx,tabularx,enumitem,array}
\usepackage{adjustbox,multicol}

\usepackage{wrapfig}

\title{Global PIQA: Evaluating Commonsense Reasoning Across 100+ Languages and Cultures 
}

\usepackage{authblk}

\makeatletter
\renewcommand\AB@affilsepx{, \protect\Affilfont}

\newcommand{\affilbreak}[1]{%
    \renewcommand\AB@affilsepx{\\\protect\Affilfont}
    #1
    \renewcommand\AB@affilsepx{, \protect\Affilfont}
}
\makeatother

\author[1*]{Tyler A. Chang}
\author[2*]{Catherine Arnett}
\author[3]{\authorcr Authors at the 5th Multilingual Representation Learning (MRL) Workshop}

\affil[1]{UC San Diego}
\affilbreak{\affil[2]{EleutherAI}}
\affilbreak{\affil[3]{For full authorship list, see \S\ref{app:authors}.}}
\affil[*]{Equal contribution}

\usepackage{tcolorbox}

\definecolor{blueframe}{HTML}{B6CBFF}
\definecolor{bluebg}{HTML}{E6EDFF}

\tcbset{badge/.style={
  on line, arc=3pt, colback=white, colframe=blueframe,
  boxsep=1pt, left=3pt, right=3pt, top=1pt, bottom=1pt, 
  fontupper=\ttfamily
}}

\begin{document}

\maketitle
\vspace{-1em}

\begin{abstract}
To date, there exist almost no culturally-specific evaluation benchmarks for large language models (LLMs) that cover a large number of languages and cultures.
In this paper, we present \textbf{Global PIQA}, a participatory commonsense reasoning benchmark for over 100 languages, constructed by hand by over 350 researchers from over 65 countries around the world.
The 141 language varieties in Global PIQA cover five continents, 19 language families, and 24 writing systems. 
In the \textbf{non-parallel} split of Global PIQA, over 50\% of examples reference local foods, customs, traditions, or other culturally-specific elements.
In the \textbf{parallel} split, we translate more ``culturally agnostic'' commonsense reasoning questions into 131 language varieties, for direct cross-lingual comparisons.
In both splits, all examples have been verified by native speakers of the languages.
We find that state-of-the-art LLMs perform well on Global PIQA in aggregate, but they exhibit weaker performance in lower-resource languages (e.g. up to a 68\% accuracy gap between languages in the parallel split).
Global PIQA highlights that in many languages and cultures, everyday knowledge remains an area for improvement in LLMs, alongside more widely-discussed capabilities such as complex reasoning and expert knowledge.
Beyond its uses for LLM evaluation, Global PIQA provides a glimpse into the wide diversity of cultures in which human language is embedded.
\end{abstract}

\vspace{-1em}
\begin{center}
   \raisebox{-2.1pt}{\includegraphics[scale=0.09]{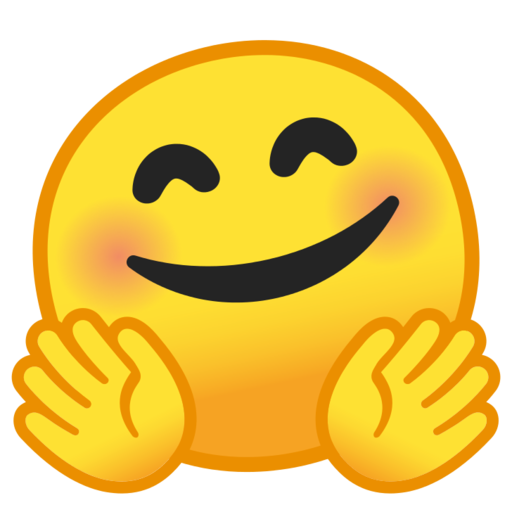}} 
   Global PIQA:
    \tcbox[badge]{%
    \href{https://huggingface.co/datasets/mrlbenchmarks/global-piqa-nonparallel}{non-parallel}}
    \tcbox[badge]{%
  \href{https://huggingface.co/datasets/mrlbenchmarks/global-piqa-parallel}{parallel}}
     
     \faGithub~\href{https://github.com/mrlbenchmarks}{\texttt{mrlbenchmarks}}
\end{center}
\vspace{-0.5em}

\section{Introduction}
\label{sec:intro}

Nearly all prominent multilingual benchmarks for large language models (LLMs) translate existing English datasets into other languages (e.g. XNLI, XCOPA, Belebele, XStoryCloze, MGSM, and Global MMLU; \citealp{conneau-etal-2018-xnli, ponti-etal-2020-xcopa, bandarkar-etal-2024-belebele, lin-etal-2022-shot, shi2023language, singh-etal-2025-global}).
As a result, the vast majority of the world's languages lack culturally-specific evaluation datasets that cover local customs, traditions, and everyday life for speakers of the language.
The culturally-specific datasets that do exist generally still rely heavily on translation or are limited to a relatively small number of languages (e.g. Global MMLU and BLEnD; \citealp{singh-etal-2025-global}; \citealp{blend_2024_myung}).

This lack of culturally-specific datasets is particularly relevant in the domain of commonsense reasoning, where LLMs are evaluated for physical, social, and world knowledge that is broadly known by the majority of people in a community.
Commonsense reasoning capabilities have long been a desirable property of LLM-based systems, evaluated through popular benchmarks such as HellaSwag \citep{zellers-etal-2019-hellaswag} and PIQA \citep{bisk_2020_piqa}.
Because commonsense reasoning focuses on everyday physical and social activities, and it has its basis in community knowledge, it differs greatly across languages and cultures.
Unfortunately, culturally-specific commonsense reasoning evaluation datasets do not exist for the vast majority of the world's languages.

To fill this gap, we present \textbf{Global PIQA}\footnote{We release Global PIQA under a \href{https://creativecommons.org/licenses/by-sa/4.0/deed.en}{CC BY-SA 4.0} license. Global PIQA is intended only for LLM evaluation. We do not allow training of AI systems on Global PIQA, or on synthetic data that uses Global PIQA as a seed.}, a culturally-specific commonsense reasoning benchmark created by native speakers of over 100 language varieties across the globe.
In contrast to previous multilingual benchmarks, examples in the \textbf{non-parallel} split of Global PIQA are written directly in each language, largely by NLP researchers who speak the language, involving very little or no translation.
Authors were given flexibility to determine the topics and domains for their examples, in order to develop ``target-language original prompts'' \citep{kreutzer2025dj} that are appropriate for each linguistic and cultural context.
The resulting non-parallel split covers 136 language varieties.
For more direct comparisons across languages, we also release a \textbf{parallel} split of more ``culturally agnostic'' commonsense reasoning questions translated into 131 language varieties.

\setlength{\belowcaptionskip}{-0.2cm}
\begin{figure}[t!]
    \centering
    \includegraphics[clip, trim=0.6cm 0.1cm 0.4cm 0.3cm, width=14.1cm]{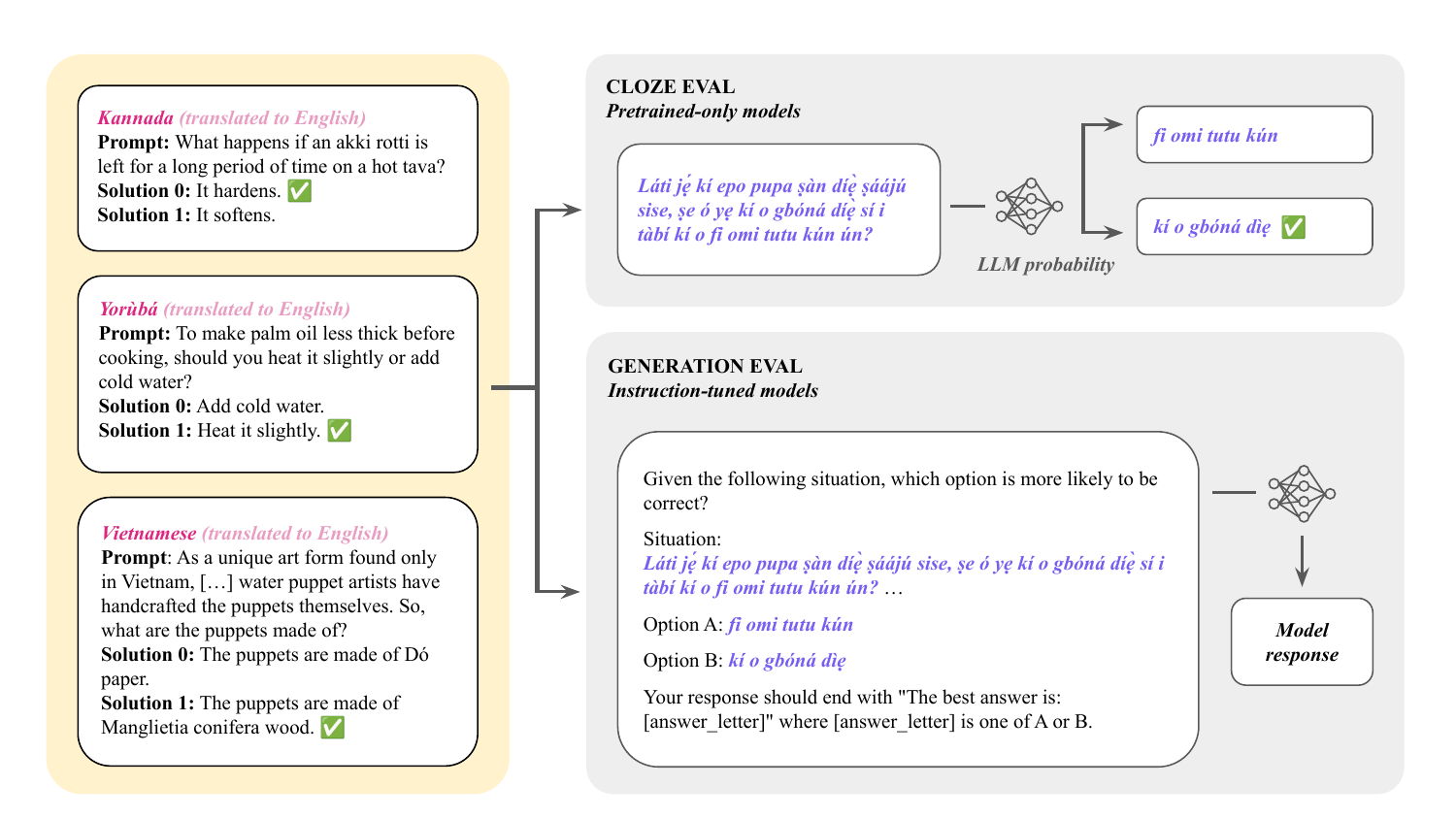}
    \caption{The format of Global PIQA examples. Each example can be used either in a cloze setting (completion probabilities, to evaluate pretrained-only models) or a generation setting (prompted, to evaluate instruction-tuned models). Evaluation method details are in \S\ref{sec:eval-format}.}
    \label{fig:example}
\end{figure}
\setlength{\belowcaptionskip}{0.0cm}

We then evaluate state-of-the-art LLMs on Global PIQA.
We find that proprietary models perform well in aggregate, with some models achieving over 90\% accuracy averaged across languages.
However, Global PIQA highlights disparities between high- and low-resource languages, along with uneven region coverage.
For example, on the parallel split, the best-performing open-weight model evaluated has over 25\% lower average scores for Sub-Saharan African languages than for European or East Asian languages.
Even the best-performing proprietary LLM has a 68\% accuracy gap between the best- and worst-performing languages (\S\ref{app:acc-gap}).
We hope that Global PIQA will enable researchers to measure and ultimately close this multilingual performance gap both across languages and between open and proprietary models.
More broadly, Global PIQA provides a glimpse into a wide variety of global cultures, through commonsense examples describing everyday life in over 100 languages.

\section{Background and Related Work} \label{sec:related_work}

\paragraph{Multilingual evaluation datasets.}
Most multilingual evaluations for standard LLM tasks (e.g. question answering and mathematical reasoning) are the product of translation from English (e.g. EU20, mArenaHard, Okapi, MMLU-ProX, and MGSM; \citealp{thellmann2024towards}; \citealp{dang2024aya}; \citealp{lai-etal-2023-okapi}; \citealp{xuan2025mmlu}; \citealp{shi2023language}).
In some cases, the translations are automatic without any human verification, which can lead to unnatural examples and low-quality datasets due to artifacts from machine translation \citep{singh-etal-2025-global}.
In other cases, benchmarks are professionally translated or use human-verified translations (e.g. Belebele, MMMLU, IrokoBench, Global MMLU, and XQuAD; \citealp{bandarkar-etal-2024-belebele}; \citealp{openai-mmmlu}; \citealp{adelani-etal-2025-irokobench}; \citealp{singh-etal-2025-global}; \linebreak \citealp{artetxe-etal-2020-cross}).
These benchmarks are less likely to suffer from quality issues related to machine translation, but they are still not necessarily culturally relevant for the target languages. Benchmarks translated from English have been found to propagate Anglocentric perspectives and values \citep{singh-etal-2025-global,kreutzer2025dj}.

\paragraph{Culturally-specific evaluation.}
Culturally-specific evaluation is critical for designing models that align with values other than those from higher-resourced countries for LLM research \citep{nigatu-etal-2024-zenos}.
Culturally-specific benchmarks have been constructed for a variety of languages (e.g. INCLUDE, TyDi QA, CulturalBench, MultiLoKo, DOSA, and BLEnD; \citealp{romanou2025include}; \citealp{clark2020tydi}; \citealp{chiu-etal-2025-culturalbench}; \citealp{hupkes2025multiloko}; \citealp{seth-etal-2024-dosa}; \citealp{blend_2024_myung}), and datasets such as MMLU \citep{hendrycks2021measuring} have been localized to other languages (e.g. CMMLU, KMMLU, ArabicMMLU, TurkishMMLU, and IndoMMLU; \citealp{li-etal-2024-cmmlu}; \citealp{son-etal-2025-kmmlu}; \citealp{koto-etal-2024-arabicmmlu}; \citealp{yuksel-etal-2024-turkishmmlu}; \citealp{koto-etal-2023-large}).
Results from these localized benchmarks correlate more strongly with human judgments of model quality than results from translated or non-localized benchmarks \citep{wu2025bitter}.
However, these datasets often focus on challenging knowledge questions in localized topics, rather than commonsense cultural knowledge which is often widely known in the community but not documented on the web.

\paragraph{Physical Interaction: Question Answering (PIQA).}
To define the task format and scope for Global PIQA, we take inspiration from English PIQA \citep{bisk_2020_piqa}.
PIQA aims to measure physical commonsense reasoning, which we note in \S\ref{sec:intro} is likely to vary substantially across languages and cultures.
In Global PIQA, we consider commonsense reasoning more broadly, defined as a broad collection of related tasks relying on knowledge of object properties, affordances (types of actions an agent can perform with an object; \citealp{gibson2014ecological, jones2022distrubutional}), physical and temporal relations, and basic world knowledge.
Each example in PIQA consists of a ``goal'' (or prompt) and two possible solutions, one correct and one incorrect (e.g. Figure~\ref{fig:example}). Prompt-solution pairs can consist of sentence beginnings and completions, questions and answers, or goals (e.g. making specific food dishes) and solutions.
Even five years after its initial release, PIQA is still being used in evaluations, e.g. reported in technical reports for releases such as Gemma 3 \citep{team2025gemma} and Llama 3 \citep{meta2024llama3}.
Despite its broad usage as a benchmark for English, PIQA has not been translated or broadly adapted as a multilingual benchmark, much less extended to massively multilingual and culturally-specific settings.\footnote{\citet{ustun-etal-2024-aya} machine-translate PIQA into 93 languages to train Aya, but the translations are not human verified. Translations also exist on Hugging Face for Catalan and Basque. Due to translation, these are not culturally specific.}

\section{Global PIQA: Non-Parallel Split}
\label{sec:non-parallel} 

Thus, we construct Global PIQA, a commonsense reasoning benchmark for 141 language varieties.
The first split of Global PIQA, covering 136 language varieties, is non-parallel (i.e. not translated across languages) to allow authors to write culturally-specific examples for their languages.
Following the PIQA dataset (\citealp{bisk_2020_piqa}; \S\ref{sec:related_work}), each example consists of a prompt and two candidate solutions, one correct and one incorrect.
Each example can be used to evaluate either a pretrained-only model (Figure~\ref{fig:example}, top) or an instruction-tuned model (Figure~\ref{fig:example}, bottom).
Determining the correct solution is designed to require some form of commonsense reasoning, such as physical reasoning, temporal reasoning, cultural knowledge, or basic world knowledge.

\subsection{Organizing a Global and Participatory Benchmark}
\label{sec:organizing}

For the non-parallel split of Global PIQA, authors contributed datasets following the task format described above (details in \S\ref{sec:construction-methods}).
Authors provided their datasets with short dataset descriptions,\footnote{Dataset descriptions ranged from single paragraphs to full length papers. Individual dataset descriptions that individual authors have decided to publicly release are on our \href{https://github.com/mrlbenchmarks/global-piqa}{GitHub}. Brief summaries are in \S\ref{app:descriptions}.} and all authors of included datasets were offered co-authorship on this paper.
To date, the Global PIQA project has involved over 350 contributors across over 65 countries and over 180 university or company affiliations.
Our authors range from early career undergraduate researchers to professors at major global universities.
In this setup, researchers benefit from co-authorship on a large benchmark paper, and they have both the domain expertise and motivation to write high quality examples. Participation is entirely voluntary. This contrasts with benchmarks where external annotators are paid to create datasets, with little incentive to create high quality examples \citep{fort-etal-2011-last,fort2011crowdsourcing}.
We recruited a diverse group of researchers through large online communities, low-resource NLP grassroots organizations, social media, and personal connections (\S\ref{app:organizing}).

\subsection{Dataset Construction Methods}
\label{sec:construction-methods}

We asked authors to construct at least 100 examples in their language, all manually checked by a native speaker of the language. Translated examples directly from the English PIQA dataset are not included in the non-parallel split of Global PIQA.
Authors were asked to construct examples (\textit{prompt}, \textit{solution0}, \textit{solution1}) where (1) the correct solution relates to physical properties of one or more objects, and (2) an average person who speaks the language natively would likely know the answer.
Because we found that many submissions did not cover \textit{physical} commonsense in a strict sense, we expanded our scope to cover commonsense reasoning more generally, as detailed in \S\ref{app:annotation}.

We also encouraged authors to include culturally-specific examples that might not be easily translatable into English, or that might require regional or cultural commonsense knowledge. Specifically, in the guidelines sent to all authors, we encouraged examples based on ``local foods, places, everyday objects, customs, traditions, religions, literature, folklore, or art forms''.
We asked authors to vary the length of their examples (e.g. to include some examples greater than 25 words long), make the two candidate solutions as similar as possible (while still having one be unambiguously correct and the other unambiguously incorrect), and avoid having the incorrect solution be ``so absurd that it is extremely obvious''. Full guidelines sent to authors are on our \href{https://github.com/mrlbenchmarks/global-piqa}{GitHub}.

Aside from these guidelines, authors were provided substantial flexibility in creating the datasets for their languages.
This is a benefit of having researchers construct their own datasets; as native speakers and researchers working in each language, they themselves are experts who can ensure the quality of their respective datasets.
This flexibility also allowed each author to construct a dataset that was culturally specific to their language and dialect, in the way that they believed was best.
Method descriptions for individual languages are in \S\ref{app:descriptions}, and we highly encourage readers to explore these individual dataset descriptions.

\paragraph{Diverse methods.}
Indeed, authors used a wide variety of methods to brainstorm and construct examples.
A total of 146 groups of authors contributed datasets.
We encouraged authors to manually write examples, and 128 out of 146 groups drafted their examples entirely manually (i.e. without the help of LLMs).
Some authors (29 groups) wrote examples motivated by content on websites or other resources in their language, such as recipe blogs, DIY pages, question forums, or how-to books.
Many groups brainstormed examples based on specific topic categories, such as food, home, clothing, transportation, hobbies, or religion.
The vast majority of groups (141 out of 146 groups) explicitly reported making their datasets at least partially culturally-specific, covering local foods, clothing, traditions, everyday life, and/or customs.

In line with the task description (\S\ref{sec:construction-methods}), all groups reported writing examples based on everyday topics.
For example, one group spent one month adapting naturally-occurring sentences spoken by family and friends, and another group read examples aloud to their parents and grandparents to verify ``colloquial [language] usage, cultural appropriateness, and everyday realism''.
All groups had their examples written or checked by at least one native speaker, and most groups (92 groups) had multiple native speakers check each example. 
Brief method details for all individual groups are in \S\ref{app:descriptions}.

A small number of groups (5 out of 146 groups) used LLMs to generate topic ideas, but not to generate examples themselves.
Another 16 groups used LLMs to initially generate examples, before filtering, editing, and manual verification by the authors. In these cases, LLMs had to be prompted carefully so as not to generate easy and generic examples; for example, one group reported that ``our preliminary attempts involved using state-of-the-art Large Language Models (LLMs) to generate question candidates. However, we found these outputs to be consistently inadequate'' (for Tamil). Another group reported that LLMs ``produced poor quality samples; no such items were included in the final dataset'' (for Azerbaijani).
The 16 groups that used LLMs to generate initial examples reported needing to filter the resulting datasets heavily for quality (e.g. keeping only 14.6\% and 22.0\% of examples in the two independent groups who reported the proportions of examples kept).

\subsection{Compiling the Non-Parallel Dataset}
\label{sec:compiling}

The next step in constructing the Global PIQA non-parallel split was to run quality checks and compile the dataset for each language.
For each language, we standardized column names, added unique example IDs, and normalized language codes to use ISO 639-3 individual language codes (e.g. \texttt{cmn} for Mandarin Chinese, c.f. macrolanguage codes; language code details in \S\ref{app:lang-codes}) with ISO 15924 script codes (e.g. \texttt{latn} for Latin script).
In cases where a dataset used a specific dialect within an individual language code, we appended an optional four-letter region code; for example, the Global PIQA language code for Brazilian Portuguese is \texttt{por\_latn\_braz}.
Finally, to inspect the data more easily, we generated initial machine translations into English using Gemini 2.5 Pro (details in \S\ref{app:cleaning}).  

\paragraph{Additional manual annotation and cultural specificity.}
Based on the LLM-generated English translations, we dropped examples that did not fit the task description (e.g. a small number of abstract logic puzzles). We also dropped examples that seemed trivially easy based on the English translations.
Finally, we annotated examples as ``culturally-specific'' if they met at least one of three criteria: (1) the example uses words that do not translate well into English, e.g. specific food dishes or local brands, (2) the example describes specific holidays, folklore, traditions, or sayings, or (3) the correct solution likely varies by region, e.g. involving local norms, laws, or customs.
Annotation details, along with motivations for our operationalization of cultural specificity, are in \S\ref{app:annotation}.
For contributions where all examples were non-culturally-specific, or where dropping trivial and off-task examples led to a dataset with under 100 examples in the language, we worked with those authors to reach the 100 example threshold and to increase their number of culturally-specific examples if possible.

\paragraph{Subsampling.}
After cleaning but before any subsampling, the full dataset consists of 29.1K examples covering 136 language codes (\S\ref{app:lang-codes}).
Because this full dataset is highly skewed across languages and often overwhelmed by non-culturally-specific examples, we subsampled to an \textit{official non-parallel split} of exactly 100 examples per language to use for model evaluations.
First, where possible (i.e. when this did not reduce our sample size to less than 100 examples for a given language), we filtered out examples where the two candidate solutions differed in length by more than 25 bytes, when normalized to English byte equivalents \citep{arnett2024bit}.
We also filtered out examples whose non-stopword tokens overlapped by more than 50\% with another example in the dataset, using the per-language tokenizers from Goldfish \citep{chang-etal-2024-goldfish}.\footnote{Due to lack of available resources for many low-resource languages in our dataset, we define stopword tokens as tokens that appear in at least 25\% of examples in the Global PIQA dataset for that language. Dataset filtering details are in \S\ref{app:sampling}.}
This aimed to ensure diversity across topics for the official Global PIQA dataset for each language.
Subsampling details are in \S\ref{app:sampling}.

Then, we sampled 100 examples from this filtered subset for each language. We sampled culturally-specific examples before non-culturally specific examples (as annotated above), and within each of these categories, we first sampled examples that did not use any LLMs in the creation process.
In the resulting official non-parallel split, 59.9\% of examples are annotated as culturally-specific, and only 4.1\% of examples are written with the help of LLMs.

\paragraph{Secondary validation and English translations.}
Finally, to make examples readable by a broader audience, authors corrected the machine translations into English for the official non-parallel split, for 126 of the 136 language varieties.
Authors followed the same translation correction guidelines as for the parallel split later in \S\ref{sec:parallel} (details in \S\ref{app:english-corrections}). As part of this process, authors were also asked to verify the correctness of each example in the original language.
As a result, 7.3\% of examples were modified in the source language during this final stage of correction, and 48.7\% of the machine translations to English were modified.
Lastly, because not all authors are native speakers of English, we used Gemini 3.0 Flash with manual validation by a native English speaker to correct any grammatical errors in the English translations, working with the dataset authors to ensure that the English translations remained semantically accurate (\S\ref{app:english-corrections}).
We note that the vast majority of our authors are not professional translators, and thus the English translations are intended only to convey the general meaning of each example to a broader audience.

\begin{figure}
    \centering
    \includegraphics[width=\linewidth]{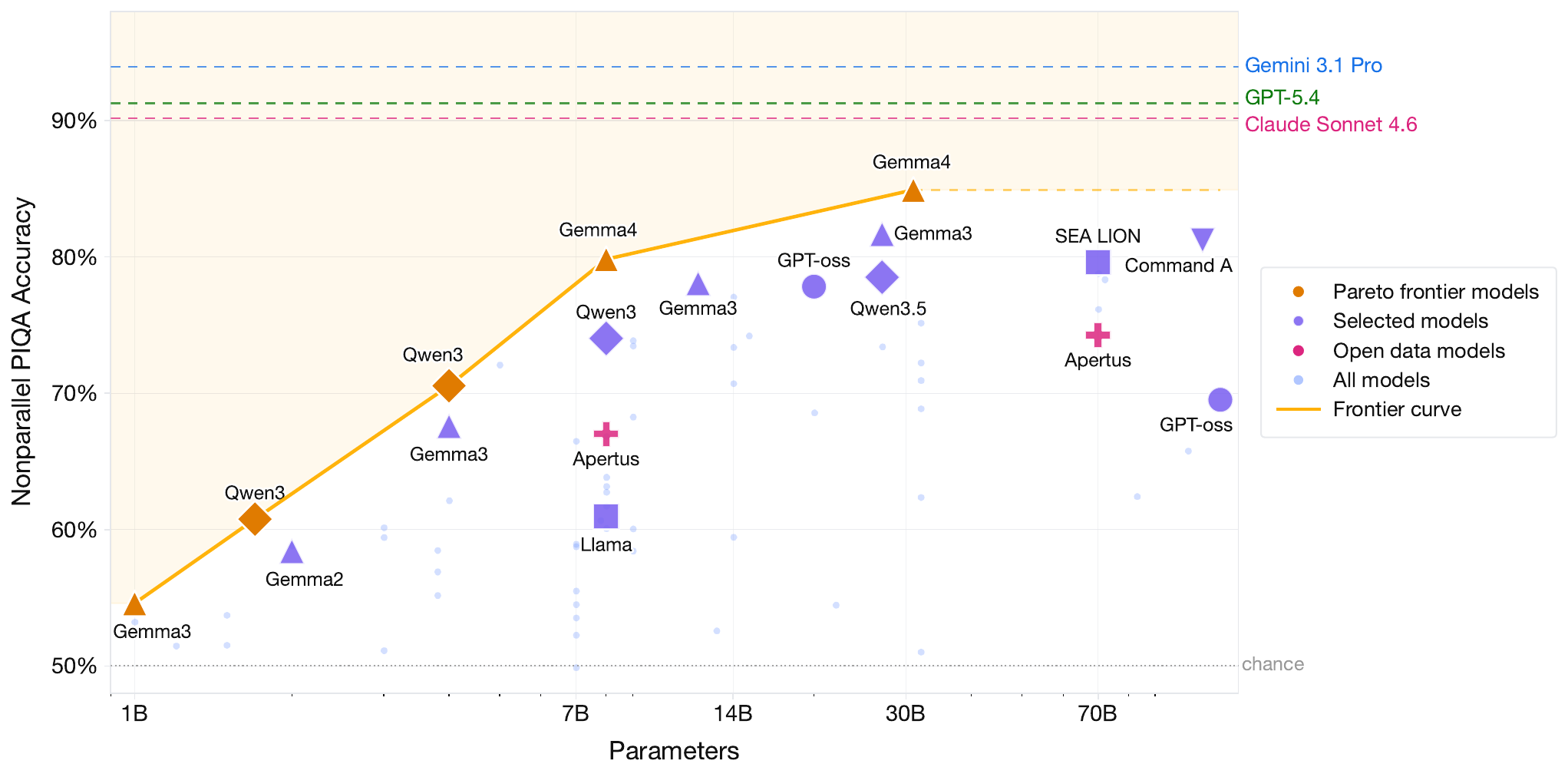}
    \vspace{-0.2cm}
    \caption{Accuracy averaged across all languages vs. parameter count for open-weight models, on the non-parallel split of Global PIQA. All evaluations here use the generation-style task format. We display names of top-performing models. Shape indicates model family, and color indicates openness (open-weight in purple vs. fully open in pink, including open data). All other models are plotted as dots. Chance performance (50\%) and performance of top closed systems are plotted as dashed lines. The analogous figure for the parallel split is in \S\ref{app:pareto-frontier-parallel}.} 
    \vspace{-0.2cm}
    \label{fig:pareto-frontier}\end{figure}

\subsection{Official Non-Parallel Split}
\label{sec:official-nonparallel}

The resulting official non-parallel split of Global PIQA contains 100 examples per language for 136 language codes.
When excluding region codes, the Global PIQA non-parallel split covers 124 language-script combinations and 113 unique ISO 639-3 language codes.
These languages cover five continents, 18 language families, and 24 scripts (writing systems).
The full list of languages is in \S\ref{app:lang-codes}.
All examples have been manually validated by at least one native speaker of the respective language, and 97.8\% of examples have been validated multiple times by native speakers; 92.6\% of examples include manually-verified English translations.
Even outside of LLM evaluation, we have anecdotally found that simply reading the English translations in the Global PIQA non-parallel split can provide a unique glimpse into a wide variety of traditions and cultures across the globe.

\section{Global PIQA: Parallel Split}
\label{sec:parallel}

The non-parallel split of Global PIQA enables culturally-specific evaluations of LLMs in 136 language varieties.
However, its non-parallel nature makes direct comparisons between languages difficult.
Thus, we also construct a parallel split of Global PIQA, consisting of 103 ``culturally agnostic'' commonsense reasoning questions translated from English into 131 language varieties.

\paragraph{Example construction.}
Examples in the parallel split differ from the non-parallel split in that (1) they have four candidate solutions per prompt instead of two, and (2) they only consist of prompts and solutions that are questions and answers, with no examples formatted as sentence beginnings with candidate completions.
To construct these examples, two native English speakers wrote 109 examples in English, with inspiration drawn from the original English PIQA \citep{bisk_2020_piqa}, EWoK \citep{ewok_2025_ivanova}, TRAM \citep{wang-zhao-2024-tram}, PROST \citep{aroca-ouellette-etal-2021-prost}, \citet{glenberg2000symbol}, HellaSwag \citep{zellers-etal-2019-hellaswag}, and difficult examples from the non-parallel split of Global PIQA.
Examples were written to require knowledge of object properties or affordances, object interactions, spatial or temporal reasoning, or basic counting (\S\ref{app:parallel-construction}).
The authors wrote all examples to be as ``culturally agnostic'' as possible, with minimal references to local dishes, customs, or traditions, to facilitate translation to a large number of languages.
After review by two native English speakers, each example was machine translated into all 131 language varieties in the parallel split.
Machine translations used either Gemini 2.5 Pro or Gemini 3.0 Flash (\S\ref{app:parallel-translation}).

\paragraph{Translation corrections.}

The machine translations were then sent for correction to authors who were native speakers of each language.
Translation correction guidelines are in \S\ref{app:parallel-corrections}, and each translation was corrected or verified by at least one native speaker of the target language.
The mean character edit distance per example between the corrected and uncorrected translations was 24.9 characters (mean 12.9\% of characters), with all edit distances and uncorrected machine translations available in the example supplements on Hugging Face.
The languages with the highest mean character edit distances were Ekpeye (273.7 characters per example), Idoma (209.0 characters per example), and Urhobo (131.6 characters per example).
Common issues that arose during translation correction are described in \S\ref{app:parallel-translation} and \S\ref{app:parallel-corrections}.

All authors were also asked to verify that each question and correct solution remained valid after translation.
In many cases, native speakers were able to use common loan words or approximate translations to substitute for words without direct equivalents in their language.
For example, in Urhobo, loan words for blue and yellow were ``borrowed and Urhobonized'', as our Urhobo-speaking author described that ``there are only three colors in the Urhobo language: red, white, and black''; this is not uncommon crosslinguistically \citep{wals-132}.
Still, six examples were dropped from all languages after translation correction, either due to ambiguities in the original English example, or due to ambiguities arising from translation.
For instance, in an example involving melting vs. dissolving sugar, we found that the verb for ``dissolve'' is the same as ``melt'' in at least 25 of the language varieties in the parallel split.
Additionally, two examples related to cardinal directions had to be dropped only from Ekpeye because, as reported by our Ekpeye-speaking author, ``there is no term to designate north, south, east, or west'' in Ekpeye; this is also not uncommon crosslinguistically \citep{brown1983cardinal}. These cases highlight the limitations of translated benchmarks, particularly for languages that are typologically distinct from and less culturally similar to most high-resource languages.

\subsection{Official parallel split}
In total, the official parallel split of Global PIQA contains 103 human-verified parallel examples in 131 language varieties (except for Ekpeye, which has 101 examples).
The full list of languages is in \S\ref{app:lang-codes}.
Although the parallel split of Global PIQA does not facilitate culturally-specific evaluations, it enables more direct comparisons across languages.
The parallel and non-parallel splits of Global PIQA together allow researchers to determine the extent to which commonsense reasoning performance differences across languages are due to (1) differences in models' ``culturally agnostic'' commonsense reasoning capabilities in different languages, as evaluated in the parallel split, vs. (2) differences in how well the models perform in different cultural contexts, as evaluated in the non-parallel split.

\section{Results for State-of-the-Art LLMs}
\label{sec:results-high-level-section}

Finally, we evaluate existing LLMs on Global PIQA. We find that proprietary models perform well when averaged across languages, but performance is substantially worse for some languages and regions. 
Dense open-weight models generally under-perform relative to closed models.

\subsection{Evaluation Format}
\label{sec:eval-format}
We evaluate models in one of two formats: cloze or generation (Figure~\ref{fig:example}). All examples in Global PIQA are amenable to either format, and both formats are implemented in the LM Evaluation Harness (\citealp{eval-harness}; \S\ref{app:evaluation-details}).
For most use cases, we recommend the generation format, as the cloze format can be difficult without few-shot prompting for pretrained-only models \citep{brown-etal-2020-language}.

\textbf{Cloze-style evaluation}: For models that are not tuned to follow instructions (i.e. pretrained-only or ``base'' models), we compute the log-probability from the LLM for each candidate solution given the prompt, normalized by the length of each solution in bytes: $\textrm{log}(P(\textit{solution } | \textit{ prompt})) / \textrm{ len}(\textit{solution})$. If the correct solution has a higher normalized probability than the incorrect solution(s), then we mark the model correct for that example.

\textbf{Generation-style evaluation}: For models that are tuned to follow instructions (e.g. the vast majority of proprietary models, and instruction-tuned and RL-tuned open models), we prompt the LLM with the prompt template in Figure~\ref{fig:example}. For the parallel split, we modify the prompt template slightly (\S\ref{app:evaluation-details}), to accommodate the four-choice question-answer task format. We sample up to 2048 tokens, then score the responses using string matching. Evaluation method details are in \S\ref{app:evaluation-details}.
In the main text here, we report generation-style evaluation results unless otherwise noted.

\subsection{Models}

We evaluate a wide range of open, open-weight, and proprietary (closed) systems on Global PIQA.
As noted in \S\ref{sec:eval-format}, we focus on instruction-tuned models (including proprietary models), which we evaluate with the generation-style format.
Evaluated models include Apertus \citep{hernandez2025apertus}, Qwen 2.5, 3, and 3.5 \citep{yang2024qwen2_5,qwen3technicalreport}, Llama 3.1 and 3.2 \citep{meta2024llama3}, Gemma 2, 3, and 4 \citep{team2024gemma,team2025gemma,gemma4_2025}, Aya, Command A, and Command R \citep{dang2024aya, cohere2025command}, GPT-5.4 (Regular, Mini, and Nano; \citealp{openai2025gpt5_4systemcard}), Claude Sonnet 4.6 \citep{anthropic2026claude_sonnet46_system_card}, Gemini 3.1 (Pro and Flash-Lite; \citealp{google2025gemini3_1_pro_modelcard, google2025gemini3_1_flash_lite}), and Gemini 3 Flash \citep{google2025gemini3_flash}.
We also evaluate a wide variety of open-weight models that are trained to focus on one language or region.
The full list of models we evaluate is in \S\ref{app:model_list}. 
Proprietary systems are evaluated with thinking on, with a 1024-token thinking budget for Gemini and Claude, and ``medium'' thinking for GPT-5.4 (details in \S\ref{app:evaluation-details}).
The open-weight models range from 300M to 120B parameters.

\setlength{\belowcaptionskip}{-0.3cm}
\begin{figure}[t]
    \centering
    \includegraphics[width=\linewidth]{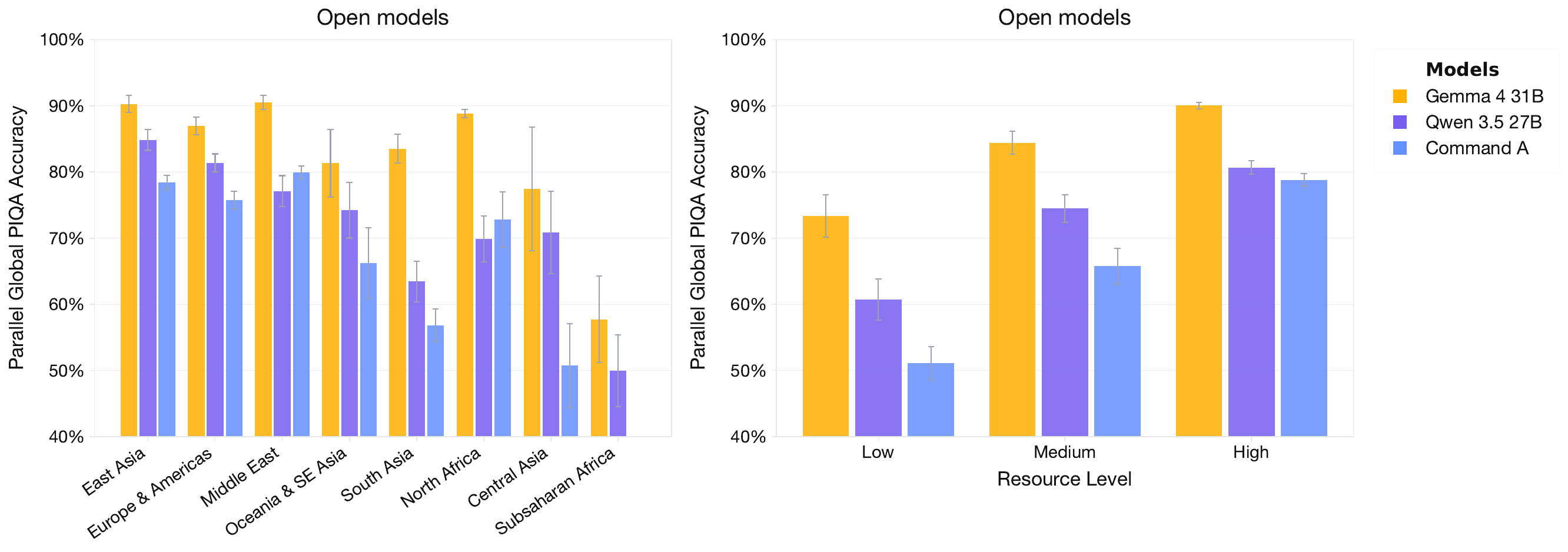}
    \vspace{0.5em}
    \includegraphics[width=\linewidth]{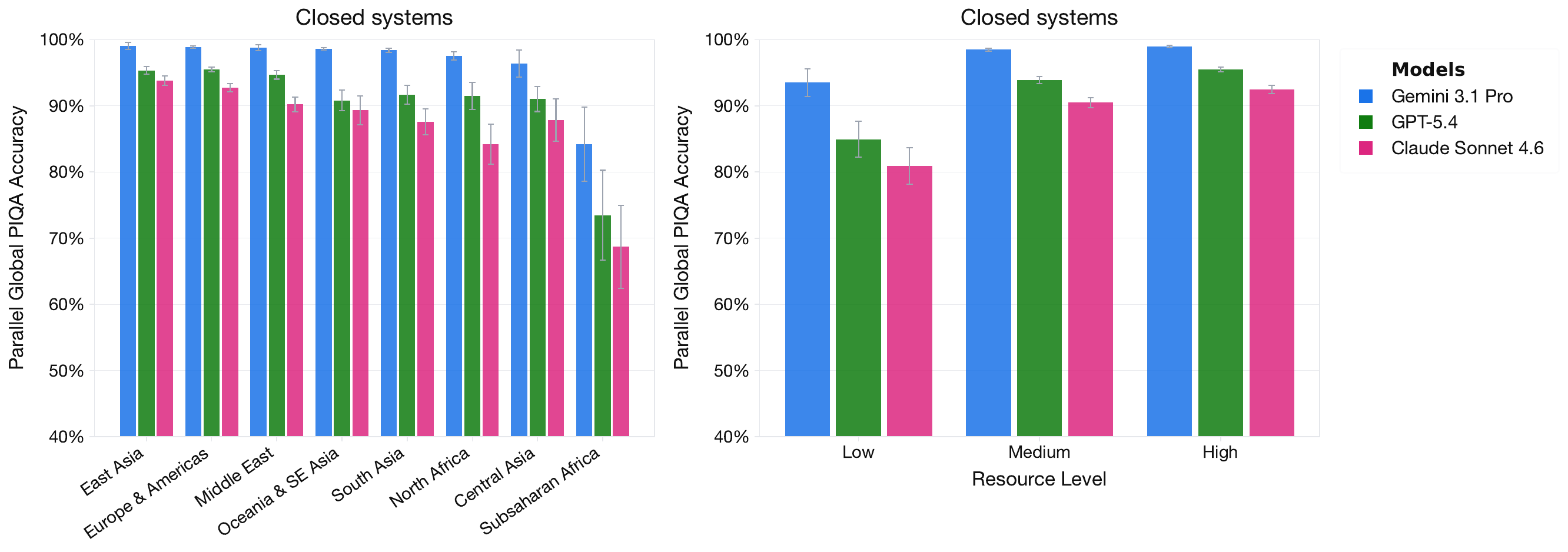}
    \vspace{-0.5em}
    \caption{Parallel split accuracies for top open-weight models and closed systems, aggregating languages by geographic region and resource level (\citealp{joshi-etal-2020-state}; \S\ref{app:by_region}). Error bars indicate one standard error of the mean. The corresponding figure for the non-parallel split is in \S\ref{app:by_region}.}
    \label{fig:regions_resource}
\end{figure}
\setlength{\belowcaptionskip}{-0.0cm}

\subsection{Results}
\label{sec:results}

In Figure~\ref{fig:pareto-frontier}, we show the top-performing open-weight models' performance on the non-parallel split of Global PIQA by model size; a similar plot for the parallel split is in \S\ref{app:pareto-frontier-parallel}.
Following \citet{michaelov2026open}, we do not directly compare open models and proprietary LLM-based systems.
Instead, closed systems (GPT-5.4, Claude, and Gemini) are plotted as horizontal dashed lines as a ``skyline'' for model performance; some of these systems achieve over 90\% accuracy on both the parallel and non-parallel splits of Global PIQA.
For open-weight models, we observe a steady performance increase as parameter counts increase, but performance begins to plateau around 30-40B parameters, and there remains a gap between the top closed systems and the strongest open-weight models that we evaluate.
The best-performing open-weight model we evaluate is Gemma 4 31B, with a mean accuracy of 82.4\% on the parallel split and 84.9\% on the non-parallel split.
We hope that Global PIQA will help direct progress towards closing this gap between smaller open-weight models and proprietary systems.

\paragraph{Disparaties across regions and languages.}
Global PIQA also highlights languages and regions for which state-of-the-art LLMs under-perform.
In Figure~\ref{fig:regions_resource}, we report mean performance on the parallel split for the top open-weight models and closed systems, grouped by region and by resource level.
Regions are defined in \S\ref{app:by_region}, and resource levels are taken from \citet{joshi-etal-2020-state}.
Similar figures for the non-parallel split are in \S\ref{app:by_region}.
In the best-performing open-weight model (Gemma 4 31B), mean accuracy for European languages is 88.1\%, but mean accuracy for Sub-Saharan African languages is only 60.5\%.
When grouping languages by resource level, Gemma 4 31B achieves 91.0\% mean accuracy for high-resource languages compared to only 75.0\% for low-resource languages.
Similar trends are observed for other models evaluated, including the proprietary LLMs (Figure~\ref{fig:regions_resource}).

Furthermore, for the parallel and non-parallel splits respectively, there are 6 and 12 languages for which the best-performing LLM achieves less than 90\% accuracy, even when including closed LLM systems.
There are 4 and 8 languages for which the best models achieve less than 80\% accuracy: Ekpeye (33\% parallel / 65\% non-parallel), Idoma (37\% / 75\%), Isoko (60\% / 96\%), Urhobo (66\% / 89\%), Lingala (92\% / 72\%), Plateau Malagasy (95\% / 76\%), Zarma (-- / 79\%), Sundanese (-- / 70\%), Meitei Manipuri (-- / 63\%), and Burushaski (-- / 59\%).
Lingala and Plateau Malagasy show the largest drop in performance from parallel to non-parallel for the best-performing models (20\% and 19\% accuracy drops), indicating weaker cultural knowledge relative to linguistic ability.
However, we note qualitatively that the non-parallel split varies substantially in difficulty across languages due to its non-parallel nature.
All results per language and model are reported in our \href{https://github.com/mrlbenchmarks/global-piqa}{GitHub}.

\section{Discussion and Conclusion}
\label{sec:discussion}

In this paper, we have presented Global PIQA, a commonsense reasoning benchmark covering 141 language varieties.
Unlike previous benchmarks, Global PIQA is a participatory benchmark, constructed by hand by over 350 researchers across over 65 countries.
This enables the construction of a culturally-specific non-parallel split, where over 50\% of examples reference local foods, clothing, customs, traditions, or other culturally-specific elements. For direct comparisons between languages, we also construct a parallel split of more ``culturally agnostic'' commonsense reasoning questions translated into over 100 languages.
We find that proprietary LLMs perform well overall on Global PIQA, but there are still significant disparities between some languages and regions.
Open-weight models generally have lower accuracies than proprietary models, and Global PIQA allows researchers to clearly quantify the gap between open and proprietary models in multilingual settings.
Notably, Global PIQA measures commonsense knowledge, demonstrating that in many languages, areas for improvement can be as simple as everyday reasoning. This contrasts with complex logical reasoning and expert knowledge, which have been the focus of many recent LLM benchmarks.

\paragraph{Limitations.}
Of course, Global PIQA has several limitations.
First, the sample size per language is only 100 examples; in the future, we hope that our participatory approach to benchmark construction will facilitate the construction of larger datasets.
Second, we note that while Global PIQA contains culturally-specific examples, these examples are snapshots specific to our authors and researchers, not necessarily representative of entire cultures.
Cultural stereotypes may be present in the dataset, although all examples are constructed by native speakers of the languages.
For language coverage, we emphasize that while we have aimed to include as many languages as possible in Global PIQA, more languages is not necessarily better when constructing multilingual benchmarks; researchers should work with communities themselves to determine if and how they want their languages included.
In Global PIQA, we have sought to work together with native speakers as authors, giving authors flexibility and ownership over how they construct their datasets.

Finally, although large closed systems perform quite well on Global PIQA when averaging across languages, we believe that Global PIQA remains useful as a benchmark.
Global PIQA distinguishes between the performance of closed systems and open models, between different open models, and even between different closed systems (Figure~\ref{fig:regions_resource}).
It also measures variation in performance across languages within each of these models.
This variation across languages and models indicates that Global PIQA remains unsaturated as a benchmark \citep{akhtar2026ai}.
We see Global PIQA as particularly useful for tracking performance of frontier systems for low-resource languages, and for tracking the development of smaller, open models for under-studied languages.

\paragraph{Conclusion.} We close by noting that the scale of participation in this project has far exceeded the organizers' expectations.
The result is a manually-curated, culturally-specific evaluation dataset for over 100 languages.
Furthermore, the parallel split of Global PIQA provides a multi-parallel evaluation dataset with unprecedented language coverage.
We are excited to continue developing community-led open-source multilingual evaluations, and we believe that this is an extremely promising avenue for addressing the critical lack of benchmarks for the vast majority of the world's languages.

\begin{ack}
See \S\ref{app:authors} for author list and acknowledgments. Global PIQA would not be possible without the efforts of all of the authors.
\end{ack}

\bibliographystyle{apalike}
\bibliography{custom}

\clearpage

\appendix

\section{Author Contributions}
\label{app:authors}

Global PIQA would not be possible without the efforts of all of the authors. We intentionally do not list authors by contributed language. This is to preserve privacy, as some authors would prefer not to be contacted by unaffiliated projects that require expertise in their language. Correspondence should be sent to the lead authors (\texttt{tachang@ucsd.edu} and \texttt{catherine@eleuther.ai}) or to \texttt{mrl.benchmarks@gmail.com}. Global PIQA is a community effort, and it does not necessarily reflect the opinions or views of the authors' affiliated organizations.

\medskip


\begin{multicols}{2}
{\raggedright

\textbf{Co-Leads} \\
Tyler A. Chang*, UC San Diego \\
Catherine Arnett*, EleutherAI \vspace{0.2cm} \\
\phantom{\_\_} *Equal contribution.
\smallskip

\setlength{\parskip}{0cm}
\textbf{Contributors (Alphabetical)}

\hangindent=0.5cm
\hangafter=1
Abdelrahman Sadallah, Università della Svizzera italiana (USI)

\hangindent=0.5cm
\hangafter=1
Abdelrahman Eldesokey, King Abdullah University of Science and Technology (KAUST)

\hangindent=0.5cm
\hangafter=1
Abeer Kashar, University of Waterloo

\hangindent=0.5cm
\hangafter=1
Abolade Daud, Masakhane

\hangindent=0.5cm
\hangafter=1
Abosede Grace Olanihun, Obafemi Awolowo University

\hangindent=0.5cm
\hangafter=1
Adamu Labaran Mohammed, Independent

\hangindent=0.5cm
\hangafter=1
Adeyemi Praise, Tonative

\hangindent=0.5cm
\hangafter=1
Adhikarimayum Meerajita Sharma, Banasthali Vidyapith

\hangindent=0.5cm
\hangafter=1
Aditi Gupta, International Institute of Information Technology Hyderabad

\hangindent=0.5cm
\hangafter=1
Adril Putra Merin, Institut Teknologi Bandung

\hangindent=0.5cm
\hangafter=1
Adwoa Bremang, Ashesi University

\hangindent=0.5cm
\hangafter=1
Afitab Iyigun, Boston University

\hangindent=0.5cm
\hangafter=1
Afonso Simplício, NOVA School of Science and Technology, NOVA University Lisbon

\hangindent=0.5cm
\hangafter=1
Ahmed Essouaied, Independent

\hangindent=0.5cm
\hangafter=1
Aicha Chorana, The University of Sydney

\hangindent=0.5cm
\hangafter=1
Akhil Eppa, Independent

\hangindent=0.5cm
\hangafter=1
Akintunde Oladipo, The African Research Collective

\hangindent=0.5cm
\hangafter=1
Akriti Kuri, Banasthali Vidyapith

\hangindent=0.5cm
\hangafter=1
Akshay Ramesh, Vellore Institute of Technology - Chennai

\hangindent=0.5cm
\hangafter=1
Aleksei Dorkin, University of Tartu

\hangindent=0.5cm
\hangafter=1
Alfred Malengo Kondoro, Hanyang University

\hangindent=0.5cm
\hangafter=1
Alham Fikri Aji, Mohamed bin Zayed University of Artificial Intelligence (MBZUAI)

\hangindent=0.5cm
\hangafter=1
Ali Eren Çetintaş, Middle East Technical University

\hangindent=0.5cm
\hangafter=1
Allan Hanbury, TU Wien

\hangindent=0.5cm
\hangafter=1
Alou Dembele, RobotsMali

\hangindent=0.5cm
\hangafter=1
Alp Niksarli, Davidson College

\hangindent=0.5cm
\hangafter=1
Álvaro Arroyo, University of Oxford

\hangindent=0.5cm
\hangafter=1
Amin Bajand, Linköping University

\hangindent=0.5cm
\hangafter=1
Amol Khanna, CrowdStrike

\hangindent=0.5cm
\hangafter=1
Ana Chkhaidze, University of California San Diego \& Free University of Tbilisi

\hangindent=0.5cm
\hangafter=1
Ana Carolina Condez, NOVA School of Science and Technology, NOVA University Lisbon

\hangindent=0.5cm
\hangafter=1
Anamaria-Roberta Hartl, Johannes Kepler University Linz

\hangindent=0.5cm
\hangafter=1
Andiswa Mkhonto, Independent

\hangindent=0.5cm
\hangafter=1
Andrew Hoblitzell, Purdue University

\hangindent=0.5cm
\hangafter=1
Andrew Tran, Independent

\hangindent=0.5cm
\hangafter=1
Angelos Poulis, Boston University

\hangindent=0.5cm
\hangafter=1
Anirban Majumder, Amazon Science (work done independently, outside of their role at Amazon)

\hangindent=0.5cm
\hangafter=1
Anjali Chaudhary, Banasthali Vidyapith

\hangindent=0.5cm
\hangafter=1
Anna Vacalopoulou, Institute for Language and Speech Processing, Athena Research Center

\hangindent=0.5cm
\hangafter=1
Annette Kuuipolani Kanahele Wong, University of Hawai`i at Mānoa

\hangindent=0.5cm
\hangafter=1
Annika Simonsen, University of Iceland

\hangindent=0.5cm
\hangafter=1
Anton Kovalev, University of Massachusetts Lowell

\hangindent=0.5cm
\hangafter=1
Anupam Nayak, Carnegie Mellon University

\hangindent=0.5cm
\hangafter=1
Ashvanth S, Cohere Labs Community

\hangindent=0.5cm
\hangafter=1
Ayodeji Lana, Independent

\hangindent=0.5cm
\hangafter=1
Ayu Purwarianti, Institut Teknologi Bandung

\hangindent=0.5cm
\hangafter=1
Bashar Alhafni, Mohamed bin Zayed University of Artificial Intelligence (MBZUAI)

\hangindent=0.5cm
\hangafter=1
Benedict Busole, Independent

\hangindent=0.5cm
\hangafter=1
Bernard Ghanem, King Abdullah University of Science and Technology (KAUST)

\hangindent=0.5cm
\hangafter=1
Bharti Nathani, Banasthali Vidyapith

\hangindent=0.5cm
\hangafter=1
Biljana Stojanovska Đurić, University of Rijeka

\hangindent=0.5cm
\hangafter=1
Blessing Ogundipe, University of Ibadan

\hangindent=0.5cm
\hangafter=1
Bolaotan Agbonile, Zabbot LLC

\hangindent=0.5cm
\hangafter=1
Bragi Bergsson, Independent

\hangindent=0.5cm
\hangafter=1
Bruce Torres Fischer, University of Hawai`i at Hilo

\hangindent=0.5cm
\hangafter=1
Burak Tutar, Middle East Technical University

\hangindent=0.5cm
\hangafter=1
Burcu Çınar, Middle East Technical University

\hangindent=0.5cm
\hangafter=1
Cade Kane, University of Hawai`i at Mānoa

\hangindent=0.5cm
\hangafter=1
Can Udomcharoenchaikit, Vidyasirimedhi Institute of Science and Technology

\hangindent=0.5cm
\hangafter=1
Chadi Helwe, Lebanese American University

\hangindent=0.5cm
\hangafter=1
Chaithra Reddy Nerella, International Institute of Information Technology Hyderabad

\hangindent=0.5cm
\hangafter=1
Chen Cecilia Liu, Independent

\hangindent=0.5cm
\hangafter=1
Chiamaka Nwokolo, University of Ibadan

\hangindent=0.5cm
\hangafter=1
Christopher Homan, Rochester Institute of Technology

\hangindent=0.5cm
\hangafter=1
Clément Sampebgo, Ashesi University

\hangindent=0.5cm
\hangafter=1
Cristina España-Bonet, German Research Center for Artificial Intelligence (DFKI) \& Barcelona Supercomputing Center (BSC)

\hangindent=0.5cm
\hangafter=1
Cynthia Amol, Maseno University \& Tonative

\hangindent=0.5cm
\hangafter=1
Daeyoep Lee, KT Corporation

\hangindent=0.5cm
\hangafter=1
Dan Saattrup Smart, The Alexandra Institute

\hangindent=0.5cm
\hangafter=1
Dana Arad, Technion – Israel Institute of Technology

\hangindent=0.5cm
\hangafter=1
Daniil Dzenhaliou, École Polytechnique Fédérale de Lausanne (EPFL)

\hangindent=0.5cm
\hangafter=1
Dasol Choi, AIM Intelligence

\hangindent=0.5cm
\hangafter=1
David Liu, Boston University

\hangindent=0.5cm
\hangafter=1
David Semedo, NOVA School of Science and Technology, NOVA University Lisbon

\hangindent=0.5cm
\hangafter=1
David Anugraha, Stanford University

\hangindent=0.5cm
\hangafter=1
Deborah Popoola, Tonative

\hangindent=0.5cm
\hangafter=1
Deividas Mataciunas, M11 Labs

\hangindent=0.5cm
\hangafter=1
Delphine Nyaboke, Independent

\hangindent=0.5cm
\hangafter=1
Dennis Owusu, Ashesi University

\hangindent=0.5cm
\hangafter=1
Dhyuthy Krishna Kumar, Independent

\hangindent=0.5cm
\hangafter=1
Diogo Tavares, NOVA School of Science and Technology, NOVA University Lisbon

\hangindent=0.5cm
\hangafter=1
Diogo Glória-Silva, NOVA School of Science and Technology, NOVA University Lisbon

\hangindent=0.5cm
\hangafter=1
Divyanshu Goyal, Adobe Inc.

\hangindent=0.5cm
\hangafter=1
DongGeon Lee, Pohang University of Science and Technology

\hangindent=0.5cm
\hangafter=1
E. Kelly Buchanan, Stanford University

\hangindent=0.5cm
\hangafter=1
Ebele Nwamaka Anajemba, Nnamdi Azikiwe University, Awka

\hangindent=0.5cm
\hangafter=1
Egonu Ngozi Grace, Alvan Ikoku Federal University of Education Owerri, Imo State, Nigeria

\hangindent=0.5cm
\hangafter=1
Elena Mickel, Independent

\hangindent=0.5cm
\hangafter=1
Elias Herranen, Independent

\hangindent=0.5cm
\hangafter=1
Eliza Acharya, Independent

\hangindent=0.5cm
\hangafter=1
Eman Nisar, Independent

\hangindent=0.5cm
\hangafter=1
Emile Anand, Georgia Institute of Technology

\hangindent=0.5cm
\hangafter=1
Emmanuel Habumuremyi, Rwanda Journalists Association

\hangindent=0.5cm
\hangafter=1
Emuobonuvie Maria Ajiboye, Delta State University, Abraka

\hangindent=0.5cm
\hangafter=1
Eryawan Presma Yulianrifat, Universitas Indonesia

\hangindent=0.5cm
\hangafter=1
Esther Adenuga, The African Research Collective

\hangindent=0.5cm
\hangafter=1
Ewa Rudnicka, Wrocław University of Science and Technology

\hangindent=0.5cm
\hangafter=1
Faith Itiola, Texas State University

\hangindent=0.5cm
\hangafter=1
Faran Taimoor Butt, Independent

\hangindent=0.5cm
\hangafter=1
Fareeha Fayyaz Sheikh, Independent

\hangindent=0.5cm
\hangafter=1
Fathima Thekkekara, Independent

\hangindent=0.5cm
\hangafter=1
Fatima Haouari, University of Sheffield

\hangindent=0.5cm
\hangafter=1
Faustin Nsengiyumva, Independent

\hangindent=0.5cm
\hangafter=1
Fenal Ashokbhai Ilasariya, Stevens Institute of Technology

\hangindent=0.5cm
\hangafter=1
Filbert Aurelian Tjiaranata, Universitas Indonesia

\hangindent=0.5cm
\hangafter=1
Firas Laakom, King Abdullah University of Science and Technology (KAUST)

\hangindent=0.5cm
\hangafter=1
Francesca Grasso, University of Turin

\hangindent=0.5cm
\hangafter=1
Francesco Periti, Clario

\hangindent=0.5cm
\hangafter=1
Francesco Orabona, King Abdullah University of Science and Technology (KAUST)

\hangindent=0.5cm
\hangafter=1
Gbenga Kayode Solomon, Adekunle Ajasin University

\hangindent=0.5cm
\hangafter=1
Genta Indra Winata, Capital One

\hangindent=0.5cm
\hangafter=1
Gia Nghia Ngo, True North International School

\hangindent=0.5cm
\hangafter=1
Gloria Udhedhe-oze, University of Port Harcourt

\hangindent=0.5cm
\hangafter=1
Gonçalo Vinagre, NOVA School of Science and Technology, NOVA University Lisbon

\hangindent=0.5cm
\hangafter=1
Gopi Naga Sai Ram Challagolla, Independent

\hangindent=0.5cm
\hangafter=1
Gorka Urbizu-Garmendia, Orai NLP Teknologiak

\hangindent=0.5cm
\hangafter=1
Gouthami Vadithya, University of New Haven

\hangindent=0.5cm
\hangafter=1
Guijin Son, Seoul National University

\hangindent=0.5cm
\hangafter=1
Gulnaz Abdykadyrova, Independent

\hangindent=0.5cm
\hangafter=1
Gyan Swaroop Mohapatra, Independent

\hangindent=0.5cm
\hangafter=1
Hafeez Ullah, University of Gwadar

\hangindent=0.5cm
\hangafter=1
Hafsteinn Einarsson, University of Iceland

\hangindent=0.5cm
\hangafter=1
Hai Hu, City University of Hong Kong

\hangindent=0.5cm
\hangafter=1
Hamidreza Saffari, Politecnico di Milano

\hangindent=0.5cm
\hangafter=1
Hamza Zaidi, University of Waterloo

\hangindent=0.5cm
\hangafter=1
Haopeng Zhang, University of Hawai`i at Mānoa

\hangindent=0.5cm
\hangafter=1
Harethah Abu Shairah, King Abdullah University of Science and Technology (KAUST)

\hangindent=0.5cm
\hangafter=1
Harry Vuong, Independent

\hangindent=0.5cm
\hangafter=1
Hele-Andra Kuulmets, University of Tartu

\hangindent=0.5cm
\hangafter=1
Hitesh Laxmichand Patel, Oracle

\hangindent=0.5cm
\hangafter=1
Houda Bouamor, Carnegie Mellon University in Qatar

\hangindent=0.5cm
\hangafter=1
Hwanjo Yu, Pohang University of Science and Technology

\hangindent=0.5cm
\hangafter=1
Iben Nyholm Debess, University of the Faroe Islands

\hangindent=0.5cm
\hangafter=1
İbrahim Ethem Deveci, Middle East Technical University

\hangindent=0.5cm
\hangafter=1
Ikhlasul Akmal Hanif, Mohamed bin Zayed University of Artificial Intelligence (MBZUAI)

\hangindent=0.5cm
\hangafter=1
Ikhyun Cho, University of Illinois Urbana-Champaign

\hangindent=0.5cm
\hangafter=1
Inês Vieira, NOVA School of Science and Technology, NOVA University Lisbon

\hangindent=0.5cm
\hangafter=1
Inês Calvo, NOVA School of Science and Technology, NOVA University Lisbon

\hangindent=0.5cm
\hangafter=1
Isaac Manzi, MbazaNLP

\hangindent=0.5cm
\hangafter=1
Ismael Illa Salifou, Independent

\hangindent=0.5cm
\hangafter=1
Ismail Daud, The African Research Collective

\hangindent=0.5cm
\hangafter=1
Ismail Yusuf, Obafemi Awolowo University

\hangindent=0.5cm
\hangafter=1
Itay Itzhak, Technion – Israel Institute of Technology

\hangindent=0.5cm
\hangafter=1
Ivan Zhelyazkov, Independent

\hangindent=0.5cm
\hangafter=1
Ivan Belashkin, Independent

\hangindent=0.5cm
\hangafter=1
Ivan Spada, Fondazione Bruno Kessler \& University of Trento

\hangindent=0.5cm
\hangafter=1
Jacob Brinton, Boston University

\hangindent=0.5cm
\hangafter=1
Jafar Isbarov, Virginia Tech

\hangindent=0.5cm
\hangafter=1
Jaka Čibej, University of Ljubljana

\hangindent=0.5cm
\hangafter=1
Jan Kocoń, Wrocław University of Science and Technology

\hangindent=0.5cm
\hangafter=1
Jan Cuhel, Independent

\hangindent=0.5cm
\hangafter=1
Jauza Krito, Universitas Gadjah Mada

\hangindent=0.5cm
\hangafter=1
Jebish Purbey, Zerograd.ai

\hangindent=0.5cm
\hangafter=1
Jennifer Za, Independent

\hangindent=0.5cm
\hangafter=1
Jennifer Mickel, EleutherAI Community

\hangindent=0.5cm
\hangafter=1
Jenny Kunz, Linköping University

\hangindent=0.5cm
\hangafter=1
Jessica Ratovondranto, Clemson University

\hangindent=0.5cm
\hangafter=1
Jeyarajalingam Varsha, University of Jaffna

\hangindent=0.5cm
\hangafter=1
Jihae Jeong, Pohang University of Science and Technology

\hangindent=0.5cm
\hangafter=1
Jimena Tena Dávalos, Independent

\hangindent=0.5cm
\hangafter=1
Jinu Lee, University of Illinois Urbana-Champaign

\hangindent=0.5cm
\hangafter=1
João Magalhães, NOVA School of Science and Technology, NOVA University Lisbon

\hangindent=0.5cm
\hangafter=1
John Seon Keun Yi, Boston University

\hangindent=0.5cm
\hangafter=1
Jongin Kim, Boston University

\hangindent=0.5cm
\hangafter=1
Joseph Chataignon, University of Bern

\hangindent=0.5cm
\hangafter=1
Joseph Marvin Imperial, National University Philippines \& University of Bath

\hangindent=0.5cm
\hangafter=1
Jubeerathan Thevakumar, University of Moratuwa

\hangindent=0.5cm
\hangafter=1
Judith Land, Independent

\hangindent=0.5cm
\hangafter=1
Julia Alekseenko, University of Strasbourg \& Centre national de la recherche scientifique (CNRS) \& INSERM

\hangindent=0.5cm
\hangafter=1
Junchen Jiang, Shanghai Jiao Tong University

\hangindent=0.5cm
\hangafter=1
Jungwhan Kim, NAVER Cloud

\hangindent=0.5cm
\hangafter=1
Kairit Sirts, University of Tartu

\hangindent=0.5cm
\hangafter=1
Kamesh R, Independent

\hangindent=0.5cm
\hangafter=1
Kamesh V, Sathyabama Institute of Science and Technology

\hangindent=0.5cm
\hangafter=1
Kanda Tshinu, Tshwane University of Technology

\hangindent=0.5cm
\hangafter=1
Kätriin Kukk, Linköping University

\hangindent=0.5cm
\hangafter=1
Kaustubh Ponkshe, École Polytechnique Fédérale de Lausanne (EPFL)

\hangindent=0.5cm
\hangafter=1
Kavsar Huseynova, Baku Higher Oil School

\hangindent=0.5cm
\hangafter=1
Ke He, Shanghai Jiao Tong University

\hangindent=0.5cm
\hangafter=1
Kenneth Enevoldsen, Aarhus University

\hangindent=0.5cm
\hangafter=1
Kent Joshua Alvarez, Independent

\hangindent=0.5cm
\hangafter=1
Kerem Zaman, University of North Carolina at Chapel Hill

\hangindent=0.5cm
\hangafter=1
Khalil Mrini, Oracle

\hangindent=0.5cm
\hangafter=1
Kian Kyars, Independent

\hangindent=0.5cm
\hangafter=1
Komal Gour, Banasthali Vidyapith

\hangindent=0.5cm
\hangafter=1
Krishnakumar Lainitha, University of Moratuwa

\hangindent=0.5cm
\hangafter=1
Krister Kruusmaa, Tallinn University \& Institute of the Estonian Language

\hangindent=0.5cm
\hangafter=1
Kunal Mukherjee, Virginia Tech

\hangindent=0.5cm
\hangafter=1
Kusum Chouhan, Banasthali Vidyapith

\hangindent=0.5cm
\hangafter=1
Laura Castro, Centro Singular de Investigación en Tecnoloxías Intelixentes (CiTIUS-USC), University of Santiago de Compostela

\hangindent=0.5cm
\hangafter=1
Laura M. Porrino-Moscoso, Universidad Alfonso X El Sabio

\hangindent=0.5cm
\hangafter=1
Lenny Sivi Za Nzambi, United World College of the Adriatic

\hangindent=0.5cm
\hangafter=1
Leshem Choshen, IBM Research \& MIT-IBM Watson AI Lab \& Massachusetts Institute of Technology (MIT)

\hangindent=0.5cm
\hangafter=1
Levent Sencan, Boston University

\hangindent=0.5cm
\hangafter=1
Lilja Øvrelid, University of Oslo

\hangindent=0.5cm
\hangafter=1
Lisa Alazraki, Imperial College London

\hangindent=0.5cm
\hangafter=1
Loretta Oma Jones, University of Benin

\hangindent=0.5cm
\hangafter=1
Lovina Ehimen-Ugbede, Independent

\hangindent=0.5cm
\hangafter=1
Luheerathan Thevakumar, Independent

\hangindent=0.5cm
\hangafter=1
Luxshan Thavarasa, University of Moratuwa

\hangindent=0.5cm
\hangafter=1
Mahnoor Malik, NED University of Engineering and Technology (Karachi, Pakistan)

\hangindent=0.5cm
\hangafter=1
Mamadou K. Keita, Rochester Institute of Technology

\hangindent=0.5cm
\hangafter=1
Mansi Jangid, Banasthali Vidyapith

\hangindent=0.5cm
\hangafter=1
Marco De Santis, University of Udine

\hangindent=0.5cm
\hangafter=1
Marcos Garcia, Centro Singular de Investigación en Tecnoloxías Intelixentes (CiTIUS-USC), University of Santiago de Compostela

\hangindent=0.5cm
\hangafter=1
Marek Šuppa, Comenius University in Bratislava \& Cisco

\hangindent=0.5cm
\hangafter=1
Mariam D'Ciofalo, Independent

\hangindent=0.5cm
\hangafter=1
Marii Ojastu, University of Tartu

\hangindent=0.5cm
\hangafter=1
Marium Attaullah, Independent

\hangindent=0.5cm
\hangafter=1
Maryam Sikander, Cohere Labs Community

\hangindent=0.5cm
\hangafter=1
Mausami Narayan, Independent

\hangindent=0.5cm
\hangafter=1
Maximos Skandalis, Laboratoire d'Informatique, de Robotique et de Microélectronique de Montpellier (LIRMM) \& Centre national de la recherche scientifique (CNRS) \& University of Montpellier

\hangindent=0.5cm
\hangafter=1
Mehak Mehak, Independent

\hangindent=0.5cm
\hangafter=1
Mehmet İlteriş Bozkurt, Middle East Technical University

\hangindent=0.5cm
\hangafter=1
Melaku Bayu, Addis Ababa University

\hangindent=0.5cm
\hangafter=1
Menan Velayuthan, University of Jaffna

\hangindent=0.5cm
\hangafter=1
Mhasilenuo Vizo, Banasthali Vidyapith

\hangindent=0.5cm
\hangafter=1
Michael Leventhal, RobotsMali

\hangindent=0.5cm
\hangafter=1
Michał Marcińczuk, CodeNLP (Gdańsk, Poland)

\hangindent=0.5cm
\hangafter=1
Mina Almasi, Aarhus University

\hangindent=0.5cm
\hangafter=1
Mirna Potočnjak, Independent

\hangindent=0.5cm
\hangafter=1
Mithil Bangera, University of New Haven

\hangindent=0.5cm
\hangafter=1
Mohammadamin Shafiei, University of Milan

\hangindent=0.5cm
\hangafter=1
Mohiba Ansari, Banasthali Vidyapith

\hangindent=0.5cm
\hangafter=1
Mridul Sharma, Institute for Research and Innovation in Intelligent Systems (IRIIS)

\hangindent=0.5cm
\hangafter=1
Mrityunjaya Indoria, Banasthali Vidyapith

\hangindent=0.5cm
\hangafter=1
Mughees Ur Rehman, Virginia Tech

\hangindent=0.5cm
\hangafter=1
Muhammad Ravi Shulthan Habibi, Universitas Indonesia

\hangindent=0.5cm
\hangafter=1
Murat Kolić, Independent

\hangindent=0.5cm
\hangafter=1
Murat Barkın Kınay, Robert College

\hangindent=0.5cm
\hangafter=1
Nada Galant, Čakavski sabor

\hangindent=0.5cm
\hangafter=1
Naina Singh Rathore, Banasthali Vidyapith

\hangindent=0.5cm
\hangafter=1
Naphat Permpredanun, Independent

\hangindent=0.5cm
\hangafter=1
Narada Maugin, Sorbonne University

\hangindent=0.5cm
\hangafter=1
Nathalie Norman, Copenhagen University

\hangindent=0.5cm
\hangafter=1
Nicholas Kluge Corrêa, University of Bonn

\hangindent=0.5cm
\hangafter=1
Nikola Ljubešić, Jožef Stefan Institute

\hangindent=0.5cm
\hangafter=1
Nirmal Thomas, Pratham International

\hangindent=0.5cm
\hangafter=1
Nisansa de Silva, University of Moratuwa

\hangindent=0.5cm
\hangafter=1
Nisheeth Joshi, Banasthali Vidyapith

\hangindent=0.5cm
\hangafter=1
Nitish Ponkshe, University of Minnesota Twin Cities

\hangindent=0.5cm
\hangafter=1
Nizar Habash, New York University (NYU) Abu Dhabi

\hangindent=0.5cm
\hangafter=1
Nneoma Udeze, Northwestern University

\hangindent=0.5cm
\hangafter=1
Noel Thomas, Mohamed bin Zayed University of Artificial Intelligence (MBZUAI)

\hangindent=0.5cm
\hangafter=1
Noémi Ligeti-Nagy, Eötvös Loránd University (ELTE), Research Centre for Linguistics

\hangindent=0.5cm
\hangafter=1
Nouhoum Coulibaly, RobotsMali

\hangindent=0.5cm
\hangafter=1
Odunayo Ogundepo, The African Research Collective

\hangindent=0.5cm
\hangafter=1
Odunayo Kareemat Buliaminu, University of Benin

\hangindent=0.5cm
\hangafter=1
Oghojafor Godswill Fejiro, Delta State University, Abraka

\hangindent=0.5cm
\hangafter=1
Okechukwu God'spraise, Tonative

\hangindent=0.5cm
\hangafter=1
Olanrewaju Samuel, Stony Brook University

\hangindent=0.5cm
\hangafter=1
Olaoye Deborah Oluwaseun, University of Ilorin

\hangindent=0.5cm
\hangafter=1
Olasoji Akindejoye, University of Ibadan

\hangindent=0.5cm
\hangafter=1
Olga Snissarenko, Kazakhstan Branch of Lomonosov Moscow State University (MSU)

\hangindent=0.5cm
\hangafter=1
Onyinye Anulika Chiemezie, Nnamdi Azikiwe University, Awka

\hangindent=0.5cm
\hangafter=1
Orkun Kınay, University of Edinburgh

\hangindent=0.5cm
\hangafter=1
Osman Tursun, Queensland University of Technology

\hangindent=0.5cm
\hangafter=1
Oyelade Oluwafemi Joshua, University of Ilorin \& Linguistics Island

\hangindent=0.5cm
\hangafter=1
Oyesanmi Fiyinfoluwa, University of Johannesburg

\hangindent=0.5cm
\hangafter=1
Pablo Rodríguez, Centro Singular de Investigación en Tecnoloxías Intelixentes (CiTIUS-USC), University of Santiago de Compostela

\hangindent=0.5cm
\hangafter=1
Pablo Gamallo, Centro Singular de Investigación en Tecnoloxías Intelixentes (CiTIUS-USC), University of Santiago de Compostela

\hangindent=0.5cm
\hangafter=1
Palak Arora, DIT University (Dehradun, Uttarakhand, India)

\hangindent=0.5cm
\hangafter=1
Pedro Valente, NOVA School of Science and Technology, NOVA University Lisbon

\hangindent=0.5cm
\hangafter=1
Peter Rupnik, Jožef Stefan Institute

\hangindent=0.5cm
\hangafter=1
Philip Oghenesuowho Ekiugbo, University of Benin

\hangindent=0.5cm
\hangafter=1
Prakhar Agarwal, University of Washington

\hangindent=0.5cm
\hangafter=1
Pramit Sahoo, Independent

\hangindent=0.5cm
\hangafter=1
Prokopis Prokopidis, Institute for Language and Speech Processing, Athena Research Center

\hangindent=0.5cm
\hangafter=1
Pua Niau-Puhipau, University of Hawai`i at Mānoa

\hangindent=0.5cm
\hangafter=1
Quadri Yahya, University of Abuja \& Linguistics Island

\hangindent=0.5cm
\hangafter=1
Rachele Mignone, University of Turin

\hangindent=0.5cm
\hangafter=1
Raghav Singhal, École Polytechnique Fédérale de Lausanne (EPFL)

\hangindent=0.5cm
\hangafter=1
Rahul Raja, Carnegie Mellon University

\hangindent=0.5cm
\hangafter=1
Ram Mohan Rao Kadiyala, Cohere Labs Community

\hangindent=0.5cm
\hangafter=1
Raphael Merx, The University of Melbourne

\hangindent=0.5cm
\hangafter=1
Rasmus Larsen, The Alexandra Institute

\hangindent=0.5cm
\hangafter=1
Ratnavel Rajalakshmi, Vellore Institute of Technology - Chennai

\hangindent=0.5cm
\hangafter=1
Rishav Ghosh, LMU Munich

\hangindent=0.5cm
\hangafter=1
Romina Oji, Linköping University

\hangindent=0.5cm
\hangafter=1
Ron Kekeha Solis, University of Hawai`i at Mānoa

\hangindent=0.5cm
\hangafter=1
Rui Guerra, NOVA School of Science and Technology, NOVA University Lisbon

\hangindent=0.5cm
\hangafter=1
Rushikesh Zawar, Independent

\hangindent=0.5cm
\hangafter=1
Sa'ad Nasir Bashir, Linguistics Island

\hangindent=0.5cm
\hangafter=1
Saeed Alzaabi, New York University (NYU) Abu Dhabi

\hangindent=0.5cm
\hangafter=1
Sahil Sandeep, Vellore Institute of Technology - Chennai

\hangindent=0.5cm
\hangafter=1
Sai Pavan Batchu, Independent

\hangindent=0.5cm
\hangafter=1
Sai Sandeep Kantareddy, Independent

\hangindent=0.5cm
\hangafter=1
Saleha Muzammil, University of Virginia

\hangindent=0.5cm
\hangafter=1
Salsabila Zahirah Pranida, Mohamed bin Zayed University of Artificial Intelligence (MBZUAI)

\hangindent=0.5cm
\hangafter=1
Sam Buchanan, University of California Berkeley

\hangindent=0.5cm
\hangafter=1
Samuel Rutunda, Digital Umuganda \& MbazaNLP

\hangindent=0.5cm
\hangafter=1
Sander Land, Writer Inc.

\hangindent=0.5cm
\hangafter=1
Sarah Sulollari, University of Vienna

\hangindent=0.5cm
\hangafter=1
Sardar Ali, Independent

\hangindent=0.5cm
\hangafter=1
Saroj Sapkota, Institute for Research and Innovation in Intelligent Systems (IRIIS)

\hangindent=0.5cm
\hangafter=1
Sarveswaran Kengatharaiyer, University of Jaffna

\hangindent=0.5cm
\hangafter=1
Saulius Tautvaisas, Independent

\hangindent=0.5cm
\hangafter=1
Sayambhu Sen, Independent

\hangindent=0.5cm
\hangafter=1
Sayantani Banerjee, University of Kashmir

\hangindent=0.5cm
\hangafter=1
Sebastien Diarra, RobotsMali

\hangindent=0.5cm
\hangafter=1
Segun Afolayan, University of Ilorin

\hangindent=0.5cm
\hangafter=1
Senthilnathan M, Independent

\hangindent=0.5cm
\hangafter=1
Sewoong Lee, University of Illinois Urbana-Champaign

\hangindent=0.5cm
\hangafter=1
Shaan Shah, University of California San Diego

\hangindent=0.5cm
\hangafter=1
Shankar Venkitachalam, Independent

\hangindent=0.5cm
\hangafter=1
Sharifa Djurabaeva, Independent

\hangindent=0.5cm
\hangafter=1
Sharon Ibejih, Tonative

\hangindent=0.5cm
\hangafter=1
Shivanya Shomir Dutta, Vellore Institute of Technology - Chennai

\hangindent=0.5cm
\hangafter=1
Siddhant Gupta, IIT Roorkee

\hangindent=0.5cm
\hangafter=1
Silvia Paniagua Suárez, Centro Singular de Investigación en Tecnoloxías Intelixentes (CiTIUS-USC), University of Santiago de Compostela

\hangindent=0.5cm
\hangafter=1
Sina Ahmadi, University of Zurich

\hangindent=0.5cm
\hangafter=1
Sivasuthan Sukumar, University of Moratuwa

\hangindent=0.5cm
\hangafter=1
Siyuan Song, University of Texas at Austin

\hangindent=0.5cm
\hangafter=1
Snegha A, IIT Bombay

\hangindent=0.5cm
\hangafter=1
Sokratis Sofianopoulos, Institute for Language and Speech Processing, Athena Research Center

\hangindent=0.5cm
\hangafter=1
Sona Elza Simon, IIT Bombay

\hangindent=0.5cm
\hangafter=1
Sonja Benčina, Independent

\hangindent=0.5cm
\hangafter=1
Sophie Gvasalia, Lightcast

\hangindent=0.5cm
\hangafter=1
Sphurti More, Independent

\hangindent=0.5cm
\hangafter=1
Spyros Dragazis, Boston University

\hangindent=0.5cm
\hangafter=1
Stefan Milosavljević, University of Graz

\hangindent=0.5cm
\hangafter=1
Stephan P. Kaufhold, University of California San Diego

\hangindent=0.5cm
\hangafter=1
Suba S, Independent

\hangindent=0.5cm
\hangafter=1
Sultan Alrashed, King Abdullah University of Science and Technology (KAUST)

\hangindent=0.5cm
\hangafter=1
Surangika Ranathunga, Massey University

\hangindent=0.5cm
\hangafter=1
Taiga Someya, The University of Tokyo

\hangindent=0.5cm
\hangafter=1
Taja Kuzman Pungeršek, Jožef Stefan Institute

\hangindent=0.5cm
\hangafter=1
Tal Haklay, Technion – Israel Institute of Technology

\hangindent=0.5cm
\hangafter=1
Tasi'u Jibril, Linguistics Island

\hangindent=0.5cm
\hangafter=1
Tatsuya Aoyama, Georgetown University

\hangindent=0.5cm
\hangafter=1
Tea Abashidze, Independent

\hangindent=0.5cm
\hangafter=1
Terenz Jomar Dela Cruz, Independent

\hangindent=0.5cm
\hangafter=1
Terra Blevins, Northeastern University

\hangindent=0.5cm
\hangafter=1
Themistoklis Nikas, Boston University

\hangindent=0.5cm
\hangafter=1
Theresa Idoko, Benue State University

\hangindent=0.5cm
\hangafter=1
Thu Mai Do, The Dewey Schools Hanoi

\hangindent=0.5cm
\hangafter=1
Tilek Chubakov, Independent

\hangindent=0.5cm
\hangafter=1
Tina Munda, Jožef Stefan Institute \& University of Ljubljana

\hangindent=0.5cm
\hangafter=1
Tobiloba Owoeye, Independent

\hangindent=0.5cm
\hangafter=1
Tommaso Gargiani, Independent

\hangindent=0.5cm
\hangafter=1
Uma Rathore, Banasthali Vidyapith

\hangindent=0.5cm
\hangafter=1
Uni Johannesen, University of the Faroe Islands

\hangindent=0.5cm
\hangafter=1
Uwuma Ugwu, Ignatius Ajuru University of Education

\hangindent=0.5cm
\hangafter=1
Vallerie Alexandra Putra, Bina Nusantara University

\hangindent=0.5cm
\hangafter=1
Vanya Bannihatti Kumar, Independent

\hangindent=0.5cm
\hangafter=1
Varvara Arzt, TU Wien

\hangindent=0.5cm
\hangafter=1
Vasily Konovalov, Core Language Technologies, MIRAI

\hangindent=0.5cm
\hangafter=1
Vasudevan Nedumpozhimana, Trinity College Dublin

\hangindent=0.5cm
\hangafter=1
Viktoria Ondrejova, Comenius University in Bratislava \& Cisco

\hangindent=0.5cm
\hangafter=1
Viktoryia Horbik, Independent

\hangindent=0.5cm
\hangafter=1
Vishnu Vardhan Reddy Kummitha, Independent

\hangindent=0.5cm
\hangafter=1
Vuk Dinić, Jožef Stefan Institute

\hangindent=0.5cm
\hangafter=1
Walelign Sewunetie, African Institute of Mathematical Sciences (AIMS) Research and Innovation Centre (RIC)

\hangindent=0.5cm
\hangafter=1
Winston Wu, University of Hawai`i at Hilo

\hangindent=0.5cm
\hangafter=1
Xiaojing Zhao, The Hong Kong Polytechnic University

\hangindent=0.5cm
\hangafter=1
Yacouba Diarra, RobotsMali

\hangindent=0.5cm
\hangafter=1
Yaniv Nikankin, Technion – Israel Institute of Technology

\hangindent=0.5cm
\hangafter=1
Yash Mathur, Independent

\hangindent=0.5cm
\hangafter=1
Yash Bagla, Independent

\hangindent=0.5cm
\hangafter=1
Yeshil Bangera, University of New Haven

\hangindent=0.5cm
\hangafter=1
Yixi Chen, Indiana University Bloomington

\hangindent=0.5cm
\hangafter=1
Yiyuan Li, University of North Carolina at Chapel Hill

\hangindent=0.5cm
\hangafter=1
Yolanda Xavier, Research Center for Linguistics at NOVA University Lisbon

\hangindent=0.5cm
\hangafter=1
Yonatan Belinkov, Technion – Israel Institute of Technology \& Kempner Institute, Harvard University

\hangindent=0.5cm
\hangafter=1
Zaid Alyafeai, King Abdullah University of Science and Technology (KAUST)

\hangindent=0.5cm
\hangafter=1
Zhargal Batozargalova, Independent

\hangindent=0.5cm
\hangafter=1
Zhengyang Shan, Boston University

\hangindent=0.5cm
\hangafter=1
Zhi Rui Tam, National Taiwan University

\hangindent=0.5cm
\hangafter=1
Zilu Tang, Boston University

\hangindent=0.5cm
\hangafter=1
Zuzana Nadova, Universidad del País Vasco

\medskip
\textbf{Evaluation Infrastructure} \\
Baber Abbasi, EleutherAI \\
Stella Biderman, EleutherAI
\medskip

\textbf{Workshop Organizers} \\

\hangindent=0.5cm
\hangafter=1
Catherine Arnett, EleutherAI

\hangindent=0.5cm
\hangafter=1
David Stap, NXAI

\hangindent=0.5cm
\hangafter=1
Duygu Ataman, \newline
Middle East Technical University

\hangindent=0.5cm
\hangafter=1
Fabian Schmidt, Cohere

\hangindent=0.5cm
\hangafter=1
Hila Gonen, University of British Columbia

\hangindent=0.5cm
\hangafter=1
Jiayi Wang, University College London (UCL)

\hangindent=0.5cm
\hangafter=1
Tyler A. Chang, \newline
University of California San Diego

\hangindent=0.5cm
\hangafter=1
David Ifeoluwa Adelani, \newline
McGill University \& Mila

}
\setlength{\parskip}{\baselineskip}
\end{multicols}

\textbf{Acknowledgments}: We also thank several anonymous contributors who preferred not to be authors on this paper. The research of Yolanda Xavier is supported by Portuguese national funding through the FCT – Portuguese Foundation for Science and Technology, I.P., as part of the project UID/03213/2025 – Research Center for Linguistics at NOVA University Lisbon (CLUNL) (\url{https://doi.org/10.54499/UID/03213/2025}) and by the Doctoral Grant (FCT PhD grant) number 2022.13977.BD from the same funder (\url{https://doi.org/10.54499/2022.13977.BD}). We would also like to thank Natalia Xavier for helping with the some examples. Nicholas Kluge Corrêa is supported by the state of North Rhine-Westphalia as part of the Lamarr Institute for Machine Learning and Artificial Intelligence. Groups 0022, 0136, and 0144 are supported by the following grants: LLM4DH (ARIS GC-
0002), DARIAH-SI (ARIS I0-E007), DIHUR (ARIS P6-0436), and Datavysts (Horizon Europe 101169037). The research of Jaka Čibej is supported by the research programme Language Resources and Technologies for Slovene (P6-0411), funded by the Slovenian Research and Innovation Agency (ARIS). Group 0025 is supported by the following grants: CLARIN-PL (POIR.04.02.00-00C002/19, FENG.02.04-IP.040004/24, 2024/WK/01), DARIAH-PL (POIR.04.02.00-00-D006/20, KPOD.01.18-IW.03-0013/23). Annika Simonsen was funded by the European Commission under grant agreement no. 101135671. CEB has been partially funded by the German ministry for education and research (BMBF) through the TRAILS project (grant number 01IW24005). Group 0070 is supported by funding from King Abdullah University of Science and Technology (KAUST) - Center of Excellence for Generative AI, under award number 5940. Group 0079 would like to thank Mr. Sudhir R. Narayana for help with correction and verification of items in their dataset. Sina Ahmadi gratefully acknowledges support from the University of Zurich (UZH) Postdoc Grant (reference number 269093). This work was supported by the AMALIA project inserted in measure RE-C05-i08 of the Portuguese national “Programa de Recuperação e Resiliência”, by the Fundação para a Ciência e Tecnologia (FCT), the FCT project Ref. 2024.07383.IACDC for public administration, and by the NOVA LINCS project (UID/04516/2025). We would also like to thank Moldir Baidildinova for translation verification for Kazakh. Group 0133 would like to thank the MbazaNLP community, including students from the University of Rwanda, School of Art and Languages. Zuzana Nadova would like to acknowledge the predoctoral research training contract PIF22/141, awarded by the University of the Basque Country (UPV/EHU), and the grant awarded to the Research Group Language and Speech by the Basque Government (IT1965-26), which allowed for the participation in this study. We thank the Danish Foundation Models project, as part of the Danish Research Reserve, for funding the annotation of the Danish part of the dataset. We would also like to thank Yonatan Bisk for useful insights into the original PIQA dataset.

\clearpage

\section{Language Codes and Included Languages}
\label{app:lang-codes}

We normalize all language codes in Global PIQA to use ISO 639-3 individual language codes (three letters), ISO 15924 script codes (four letters), and an optional custom four-letter region code for dialects within an individual language code. For example, the code for Mexican Spanish is \texttt{spa\_latn\_mexi}, and the code for Peninsular Spanish (as spoken in Spain) is \texttt{spa\_latn\_spai}.
In the non-parallel split, when the language code was unclear for an individual dataset based on the description from the authors, we worked with authors to identify the specific ISO 639-3 and ISO 15924 codes that would best reflect their dataset.
For clarity, we note:
\begin{itemize}[leftmargin=0.5cm,itemsep=0.05cm,topsep=0.02cm]
\item ISO 639-3 macrolanguage codes are often used in other work for some languages. We use individual language codes for more precision, and here, we show mappings from commonly-used macrolanguage codes to the ISO 639-3 individual codes used in Global PIQA:
\begin{itemize}
    \item[$\circ$] Mandarin Chinese: \texttt{zho} $\to$ \texttt{cmn} \\
    Cantonese Chinese: \texttt{zho} $\to$ \texttt{yue} \\
    Standard Estonian: \texttt{est} $\to$ \texttt{ekk} \\
    Norwegian Bokmål: \texttt{nor} $\to$ \texttt{nob} \\
    Norwegian Nynorsk: \texttt{nor} $\to$ \texttt{nno} \\
    Nepali: \texttt{nep} $\to$ \texttt{npi} \\
    Iranian Persian (Farsi): \texttt{fas} $\to$ \texttt{pes} \\
    Swahili (Kiswahili): \texttt{swa} $\to$ \texttt{swh} \\
    Northern Uzbek: \texttt{uzb} $\to$ \texttt{uzn} \\
    Standard Malay: \texttt{msa} $\to$ \texttt{zsm} \\
    Central Kurdish: \texttt{kur} $\to$ \texttt{ckb} \\
    Odia: \texttt{ori} $\to$ \texttt{ory} \\
    North Azerbaijani: \texttt{aze} $\to$ \texttt{azj} \\
    Classical Sanskrit: \texttt{san} $\to$ \texttt{cls} \\
    Plateau Malagasy: \texttt{mlg} $\to$ \texttt{plt}
\end{itemize}
\item Dialects of Arabic are often separate individual ISO 639-3 language codes. In Global PIQA, we have:
\begin{itemize}
    \item[$\circ$] Iraqi Arabic (Gelet): \texttt{acm\_arab} \\
    Yemeni Arabic: \texttt{acq\_arab} \\
    Tunisian Arabic: \texttt{aeb\_arab} \\
    Gulf Arabic: \texttt{afb\_arab} \\
    Levantine Arabic (Jordan): \texttt{apc\_arab\_jord} \\
    Levantine Arabic (Lebanon): \texttt{apc\_arab\_leba} \\
    Levantine Arabic (Palestine): \texttt{apc\_arab\_pale} \\
    Levantine Arabic (Syria): \texttt{apc\_arab\_syri} \\
    Modern Standard Arabic: \texttt{arb\_arab} \\
    Algerian Arabic: \texttt{arq\_arab} \\
    Najdi (Saudi) Arabic: \texttt{ars\_arab} \\
    Moroccan Arabic (Darija): \texttt{ary\_arab} \\
    Egyptian Arabic: \texttt{arz\_arab} \\
\end{itemize}
\item A Filipino dataset (language code \texttt{fil}) separate from the Tagalog dataset (language code \texttt{tgl}) is not included, despite the two being considered separate individual language codes in ISO 639-3.
This is because native speakers of Tagalog often refer to the two languages interchangeably; Filipino is the standardized national language of the Philippines, but it draws influence primarily from Tagalog.
\end{itemize}

Using these language, script, and optional region codes, Global PIQA contains 141 unique language varieties. This includes 129 unique ISO language-script combinations, 118 unique ISO 639-3 language codes, and 24 unique ISO 15924 script codes. Language counts per region and language family are shown in Table \ref{tab:by_family_region}.
Language families use the top-level families from Glottolog \citep{hammarstrom-etal-2021-glottolog}; we note that the Indo-European family is a large family including the Armenic, Balto-Slavic, Celtic, Germanic, Indo-Aryan, and Iranian sub-families (among others), with languages ranging from English and Spanish to Hindi and Persian (Farsi).
Regions are defined in \S\ref{app:by_region}.
We also note that all of the languages in the North America and South America regions in Global PIQA are originally European languages that are now spoken in the Americas; if possible, we hope to include more indigenous languages of the Americas in future benchmarks.

The full list of languages in Global PIQA is in Table \ref{tab:lang_results}.

\setlength{\LTcapwidth}{\linewidth}
\begin{footnotesize}
\begin{longtable}{p{2.2cm}llrrrr}
\caption{List of all languages in Global PIQA. This includes the language code, language name,
language family, resource level \citep{joshi-etal-2020-state}, percentage of culturally-specific items in the
non-parallel split, and the best LLM accuracy for both the parallel and non-parallel splits.}
\label{tab:lang_results} \\
\toprule
\textbf{Code} & \textbf{Language} & \textbf{Family} & \textbf{Res.} & \textbf{Cultural} &
\textbf{Parallel} & \textbf{Nonpar.} \\
& & & \textbf{level} & \textbf{percent} &
\textbf{acc.} & \textbf{acc.} \\
\midrule
\endfirsthead

\multicolumn{7}{c}{\tablename\ \thetable{}, continued.} \\ \\
\toprule
\textbf{Code} & \textbf{Language} & \textbf{Family} & \textbf{Res.} & \textbf{Cultural} &
\textbf{Parallel} & \textbf{Nonpar.} \\
& & & \textbf{level} & \textbf{percent} &
\textbf{acc.} & \textbf{acc.} \\
\midrule
\endhead

\midrule
\multicolumn{7}{r}{\textit{Continued on next page.}} \\
\endfoot

\bottomrule \\
\multicolumn{7}{r}{\textit{End of \tablename\ \thetable{}.}} \\
\endlastfoot
\texttt{acm\_arab} & Iraqi Arabic (Gelet) & Afro-Asiatic & 5 & 21\% & 96\% & 96\% \\
\texttt{acq\_arab} & Yemeni Arabic & Afro-Asiatic & 5 & 80\% & 93\% & 95\% \\
\texttt{aeb\_arab} & Tunisian Arabic & Afro-Asiatic & 5 & 100\% & 98\% & 100\% \\
\texttt{afb\_arab} & Gulf Arabic & Afro-Asiatic & 5 & 98\% & 99\% & 91\% \\
\texttt{als\_latn} & Northern Tosk Albanian & Indo-European & 1 & 37\% & 98\% & 95\% \\
\texttt{amh\_ethi} & Amharic & Afro-Asiatic & 2 & 100\% & 98\% & 93\% \\
\texttt{apc\_arab\_jord} & Levantine Arabic (Jordan) & Afro-Asiatic & 5 & 76\% & 100\% & 99\% \\
\texttt{apc\_arab\_leba} & Levantine Arabic (Lebanon) & Afro-Asiatic & 5 & 39\% & 100\% & 95\% \\
\texttt{apc\_arab\_pale} & Levantine Arabic (Palestine) & Afro-Asiatic & 5 & 100\% & 100\% & 95\% \\
\texttt{apc\_arab\_syri} & Levantine Arabic (Syria) & Afro-Asiatic & 5 & 100\% & 98\% & 95\% \\
\texttt{arb\_arab} & Modern Standard Arabic & Afro-Asiatic & 5 & 100\% & 99\% & 96\% \\
\texttt{arq\_arab} & Algerian Arabic & Afro-Asiatic & 5 & 100\% & 96\% & 96\% \\
\texttt{ars\_arab} & Najdi (Saudi) Arabic & Afro-Asiatic & 5 & 85\% & 97\% & 92\% \\
\texttt{ary\_arab} & Moroccan Arabic (Darija) & Afro-Asiatic & 5 & 100\% & 97\% & 90\% \\
\texttt{arz\_arab} & Egyptian Arabic & Afro-Asiatic & 5 & 100\% & 99\% & 95\% \\
\texttt{asm\_beng} & Assamese & Indo-European & 1 & 100\% & -- & 98\% \\
\texttt{azj\_latn} & North Azerbaijani & Turkic & 1 & 19\% & 98\% & 98\% \\
\texttt{bam\_latn} & Bambara & Mande & 1 & 17\% & 89\% & 85\% \\
\texttt{bcc\_arab} & Southern Balochi & Indo-European & 0 & 67\% & 93\% & 94\% \\
\texttt{bel\_cyrl} & Belarusian & Indo-European & 1 & 38\% & 97\% & 99\% \\
\texttt{ben\_beng} & Bengali & Indo-European & 3 & 100\% & 99\% & 98\% \\
\texttt{ben\_latn} & Bengali & Indo-European & 3 & 70\% & 99\% & 100\% \\
\texttt{bgc\_deva} & Haryanvi & Indo-European & 0 & 44\% & 98\% & 97\% \\
\texttt{bho\_deva} & Bhojpuri & Indo-European & 1 & 16\% & 99\% & 96\% \\
\texttt{bos\_latn} & Bosnian & Indo-European & 3 & 35\% & -- & 100\% \\
\texttt{bra\_deva} & Braj & Indo-European & 0 & 82\% & 100\% & 95\% \\
\texttt{bsk\_arab} & Burushaski & Isolate & 0 & 77\% & -- & 59\% \\
\texttt{btx\_latn} & Batak Karo & Austronesian & 0 & 34\% & -- & 98\% \\
\texttt{bul\_cyrl} & Bulgarian & Indo-European & 3 & 12\% & 98\% & 100\% \\
\texttt{bxr\_cyrl} & Russian Buryat & Mongolic & 0 & -- & 89\% & -- \\
\texttt{cat\_latn} & Catalan & Indo-European & 4 & 58\% & 99\% & 99\% \\
\texttt{ceb\_latn} & Cebuano & Austronesian & 3 & -- & 98\% & -- \\
\texttt{ces\_latn} & Czech & Indo-European & 4 & 100\% & 98\% & 96\% \\
\texttt{ckb\_arab} & Central Kurdish & Indo-European & 1 & 92\% & -- & 91\% \\
\texttt{ckm\_latn} & Chakavian & Indo-European & 0 & 24\% & 91\% & 86\% \\
\texttt{cls\_deva} & Classical Sanskrit & Indo-European & 2 & 47\% & 97\% & 100\% \\
\texttt{cmn\_hans} & Mandarin Chinese & Sino-Tibetan & 5 & 100\% & 100\% & 94\% \\
\texttt{cmn\_hant} & Mandarin Chinese & Sino-Tibetan & 5 & 94\% & 100\% & 90\% \\
\texttt{dan\_latn} & Danish & Indo-European & 3 & 46\% & 100\% & 95\% \\
\texttt{deu\_latn} & German & Indo-European & 5 & 87\% & 99\% & 98\% \\
\texttt{dhd\_deva} & Dhundari & Indo-European & 0 & 17\% & 99\% & 93\% \\
\texttt{dje\_latn} & Zarma & Nilo-Saharan & 0 & 42\% & -- & 79\% \\
\texttt{ekk\_latn} & Estonian & Uralic & 3 & 86\% & 100\% & 99\% \\
\texttt{ekp\_latn} & Ekpeye & Atlantic-Congo & 1 & 67\% & 32\% & 70\% \\
\texttt{ell\_grek} & Greek & Indo-European & 3 & 100\% & 98\% & 90\% \\
\texttt{eng\_latn} & English & Indo-European & 5 & 13\% & 100\% & 96\% \\
\texttt{eus\_latn} & Basque & Isolate & 4 & 90\% & 99\% & 96\% \\
\texttt{fao\_latn} & Faroese & Indo-European & 1 & 41\% & 100\% & 99\% \\
\texttt{fin\_latn} & Finnish & Uralic & 4 & 36\% & 100\% & 100\% \\
\texttt{fra\_latn\_cana} & French (Canada) & Indo-European & 5 & 3\% & 100\% & 98\% \\
\texttt{fra\_latn\_fran} & French (France) & Indo-European & 5 & 46\% & 100\% & 98\% \\
\texttt{glg\_latn} & Galician & Indo-European & 3 & 44\% & 99\% & 99\% \\
\texttt{guj\_gujr} & Gujarati & Indo-European & 1 & 35\% & 100\% & 95\% \\
\texttt{hau\_latn} & Hausa & Afro-Asiatic & 2 & 75\% & 99\% & 98\% \\
\texttt{haw\_latn} & Hawaiian ('ōlelo Hawai'i) & Austronesian & 1 & 25\% & 99\% & 93\% \\
\texttt{heb\_hebr} & Hebrew & Afro-Asiatic & 3 & 90\% & 99\% & 95\% \\
\texttt{hin\_deva} & Hindi & Indo-European & 4 & 100\% & 98\% & 97\% \\
\texttt{hin\_latn} & Hindi & Indo-European & 4 & 100\% & 100\% & 97\% \\
\texttt{hrv\_latn} & Croatian & Indo-European & 4 & 8\% & 99\% & 100\% \\
\texttt{hun\_latn} & Hungarian & Uralic & 4 & 15\% & 100\% & 97\% \\
\texttt{hye\_armn} & Eastern Armenian & Indo-European & 1 & 12\% & 98\% & 97\% \\
\texttt{ibo\_latn} & Igbo & Atlantic-Congo & 1 & 100\% & 95\% & 93\% \\
\texttt{idu\_latn} & Idoma & Atlantic-Congo & 0 & 83\% & 37\% & 75\% \\
\texttt{ilo\_latn} & Iloko (Ilocano) & Austronesian & 1 & -- & 98\% & -- \\
\texttt{ind\_latn} & Indonesian & Austronesian & 3 & 100\% & 99\% & 97\% \\
\texttt{isl\_latn} & Icelandic & Indo-European & 2 & 82\% & 98\% & 97\% \\
\texttt{iso\_latn} & Isoko & Atlantic-Congo & 0 & 40\% & 60\% & 96\% \\
\texttt{ita\_latn} & Italian & Indo-European & 4 & 100\% & 98\% & 98\% \\
\texttt{jav\_latn} & Javanese & Austronesian & 1 & 62\% & 98\% & 93\% \\
\texttt{jpn\_jpan} & Japanese & Japonic & 5 & 36\% & 97\% & 97\% \\
\texttt{kan\_knda} & Kannada & Dravidian & 1 & 71\% & 99\% & 97\% \\
\texttt{kan\_latn} & Kannada & Dravidian & 1 & 71\% & 99\% & 96\% \\
\texttt{kat\_geor} & Georgian & Kartvelian & 3 & 29\% & 99\% & 96\% \\
\texttt{kaz\_cyrl} & Kazakh & Turkic & 3 & 82\% & -- & 93\% \\
\texttt{kin\_latn} & Kinyarwanda & Atlantic-Congo & 1 & 67\% & 97\% & 98\% \\
\texttt{kir\_cyrl} & Kyrgyz & Turkic & 1 & 16\% & 98\% & 100\% \\
\texttt{kor\_hang} & Korean & Koreanic & 4 & 100\% & 99\% & 94\% \\
\texttt{lin\_latn} & Lingala & Atlantic-Congo & 1 & 53\% & 92\% & 72\% \\
\texttt{lit\_latn} & Lithuanian & Indo-European & 3 & 98\% & 99\% & 98\% \\
\texttt{luo\_latn} & Luo & Nilotic & 0 & 29\% & -- & 93\% \\
\texttt{mag\_deva} & Magahi & Indo-European & 0 & -- & 99\% & -- \\
\texttt{mal\_mlym} & Malayalam & Dravidian & 1 & 100\% & 99\% & 94\% \\
\texttt{mar\_deva} & Marathi & Indo-European & 2 & 84\% & 99\% & 96\% \\
\texttt{mkd\_cyrl} & Macedonian & Indo-European & 1 & 45\% & 99\% & 100\% \\
\texttt{mni\_beng} & Manipuri & Sino-Tibetan & 1 & 17\% & 93\% & 91\% \\
\texttt{mni\_mtei} & Manipuri & Sino-Tibetan & 1 & 100\% & -- & 64\% \\
\texttt{nag\_latn} & Nagamese & Pidgin & 0 & 17\% & 96\% & 90\% \\
\texttt{nld\_latn} & Dutch & Indo-European & 4 & 59\% & 99\% & 94\% \\
\texttt{nno\_latn} & Norwegian Nynorsk & Indo-European & 1 & 15\% & 100\% & 94\% \\
\texttt{nob\_latn} & Norwegian Bokmål & Indo-European & 1 & 67\% & 100\% & 97\% \\
\texttt{npi\_deva} & Nepali & Indo-European & 1 & 20\% & 100\% & 99\% \\
\texttt{ory\_orya} & Odia & Indo-European & 1 & 100\% & 99\% & 99\% \\
\texttt{pan\_guru} & Eastern Panjabi & Indo-European & 2 & 17\% & 99\% & 96\% \\
\texttt{pcm\_latn} & Nigerian Pidgin (Naijá) & Pidgin & 0 & 61\% & 96\% & 96\% \\
\texttt{pes\_arab} & Western Farsi & Indo-European & 4 & 57\% & 100\% & 95\% \\
\texttt{plt\_latn} & Plateau Malagasy & Austronesian & 1 & 91\% & 95\% & 76\% \\
\texttt{pol\_latn} & Polish & Indo-European & 4 & 80\% & 100\% & 99\% \\
\texttt{por\_latn\_braz} & Portuguese (Brazil) & Indo-European & 4 & 35\% & 100\% & 99\% \\
\texttt{por\_latn\_port} & Portuguese (Portugal) & Indo-European & 4 & 54\% & 100\% & 96\% \\
\texttt{ron\_latn} & Romanian & Indo-European & 3 & 35\% & 99\% & 96\% \\
\texttt{rus\_cyrl} & Russian & Indo-European & 4 & 54\% & 99\% & 97\% \\
\texttt{rwr\_deva} & Marwari & Indo-European & 0 & 52\% & 100\% & 98\% \\
\texttt{sin\_latn} & Sinhala & Indo-European & 0 & 77\% & 99\% & 95\% \\
\texttt{sin\_sinh} & Sinhala & Indo-European & 0 & 77\% & 99\% & 95\% \\
\texttt{slk\_latn} & Slovak & Indo-European & 3 & 13\% & 100\% & 100\% \\
\texttt{slk\_latn\_sari} & Šariš Slovak & Indo-European & 3 & 50\% & 98\% & 97\% \\
\texttt{slv\_latn} & Slovenian & Indo-European & 3 & 29\% & 100\% & 100\% \\
\texttt{slv\_latn\_cerk} & Slovenian (Cerkno) & Indo-European & 3 & 20\% & 93\% & 98\% \\
\texttt{slv\_latn\_prle} & Slovenian (Prlekija) & Indo-European & 3 & 43\% & 99\% & 93\% \\
\texttt{snd\_arab} & Sindhi & Indo-European & 1 & 80\% & 98\% & 100\% \\
\texttt{snd\_deva} & Sindhi & Indo-European & 1 & 17\% & 98\% & 94\% \\
\texttt{spa\_latn\_mexi} & Spanish (Mexico) & Indo-European & 5 & 36\% & 97\% & 99\% \\
\texttt{spa\_latn\_peru} & Spanish (Peru) & Indo-European & 5 & 20\% & 99\% & 99\% \\
\texttt{spa\_latn\_spai} & Spanish (Peninsular) & Indo-European & 5 & 100\% & 98\% & 99\% \\
\texttt{srp\_cyrl} & Serbian & Indo-European & 4 & 5\% & 100\% & 99\% \\
\texttt{srp\_cyrl\_torl} & Torlak (Serbian Torlak) & Indo-European & 4 & 61\% & 98\% & 96\% \\
\texttt{srp\_latn} & Serbian & Indo-European & 4 & 5\% & 100\% & 100\% \\
\texttt{srp\_latn\_torl} & Torlak (Serbian Torlak) & Indo-European & 4 & 59\% & 100\% & 95\% \\
\texttt{sun\_latn} & Sundanese & Austronesian & 1 & 6\% & -- & 70\% \\
\texttt{swe\_latn} & Swedish & Indo-European & 4 & 71\% & 98\% & 100\% \\
\texttt{swh\_latn} & Swahili & Atlantic-Congo & 2 & 16\% & 98\% & 90\% \\
\texttt{swv\_deva} & Shekhawati & Indo-European & 0 & 44\% & 96\% & 96\% \\
\texttt{tam\_latn} & Tamil & Dravidian & 3 & 100\% & 99\% & 89\% \\
\texttt{tam\_taml} & Tamil & Dravidian & 3 & 100\% & 99\% & 91\% \\
\texttt{tel\_latn} & Telugu & Dravidian & 1 & 81\% & 96\% & 96\% \\
\texttt{tel\_telu} & Telugu & Dravidian & 1 & 81\% & 99\% & 96\% \\
\texttt{tgl\_latn} & Tagalog / Filipino & Austronesian & 3 & 98\% & 99\% & 97\% \\
\texttt{tha\_thai} & Thai & Tai-Kadai & 3 & 49\% & 99\% & 99\% \\
\texttt{tur\_latn} & Turkish & Turkic & 4 & 100\% & 100\% & 96\% \\
\texttt{uig\_arab} & Uighur & Turkic & 1 & 57\% & 99\% & 97\% \\
\texttt{ukr\_cyrl} & Ukrainian & Indo-European & 3 & 47\% & 99\% & 97\% \\
\texttt{urd\_arab} & Urdu & Indo-European & 3 & 83\% & 99\% & 100\% \\
\texttt{urd\_latn} & Urdu & Indo-European & 3 & 44\% & 100\% & 100\% \\
\texttt{urh\_latn} & Urhobo & Atlantic-Congo & 0 & 32\% & 66\% & 89\% \\
\texttt{uzn\_latn} & Northern Uzbek & Turkic & 3 & 43\% & 98\% & 94\% \\
\texttt{vie\_latn} & Vietnamese & Austroasiatic & 4 & 93\% & 99\% & 93\% \\
\texttt{xho\_latn} & Xhosa & Atlantic-Congo & 2 & -- & 97\% & -- \\
\texttt{yor\_latn} & Yoruba & Atlantic-Congo & 2 & 100\% & 93\% & 90\% \\
\texttt{yue\_hant} & Yue Chinese (Cantonese) & Sino-Tibetan & 1 & 85\% & 99\% & 94\% \\
\texttt{zsm\_latn} & Standard Malay & Austronesian & 3 & 95\% & 97\% & 96\% \\
\texttt{zul\_latn} & Zulu & Atlantic-Congo & 2 & 17\% & 94\% & 91\% \\
\end{longtable}\end{footnotesize}
\begin{table}[h!]
\caption{Number of languages in Global PIQA per region (left) and per language family (right). Language families use the top-level families from Glottolog \citep{hammarstrom-etal-2021-glottolog}, and regions are defined in \S\ref{app:by_region}.}
\label{tab:by_family_region}
\centering
\begin{minipage}{0.35\textwidth}
\centering
\begin{tabular}{rr}
\toprule
Region & \# Langs \\ \midrule
South Asia & 36 \\
Eastern Europe & 28 \\
Subsaharan Africa & 18 \\
Western Europe & 17 \\
Middle East & 13 \\
Southeast Asia & 10 \\
Central Asia & 5 \\
East Asia & 5 \\
North Africa & 4 \\
North America & 2 \\
South America & 2 \\
Oceania & 1 \\
\bottomrule
\end{tabular}
\end{minipage}
\hspace{3em}
\begin{minipage}{0.35\textwidth}
\centering
\begin{tabular}{rr}
\toprule
Language Family & \# Langs \\ \midrule
Indo-European & 70 \\
Afro-Asiatic & 16 \\
Atlantic-Congo & 11 \\
Austronesian & 10 \\
Dravidian & 7 \\
Turkic & 6 \\
Sino-Tibetan & 5 \\
Uralic & 3 \\
Isolate & 2 \\
Pidgin & 2 \\
Mande & 1 \\
Mongolic & 1 \\
Nilotic & 1 \\
Tai-Kadai & 1 \\
Japonic & 1 \\
Austroasiatic & 1 \\
Koreanic & 1 \\
Kartvelian & 1 \\
Nilo-Saharan & 1 \\
\bottomrule
\end{tabular}
\end{minipage}
\vspace{1em}
\end{table}

\section{Organizing a Global Participatory Benchmark}
\label{app:organizing}

As described briefly in \S\ref{sec:organizing}, Global PIQA was organized as a global participatory benchmark involving over 350 contributors across over 65 countries and over 180 university or company affiliations.
All contributors were offered authorship on this paper, and the vast majority chose to be authors.
Here, we detail additional procedures that made the collaboration successful.

\begin{itemize}[leftmargin=0.5cm,itemsep=0.02cm,topsep=0.02cm]
\item \textbf{Recruiting.} We recruited a diverse group of contributors through large online communities, low-resource NLP community organizations, social media, and personal connections.
For example, we publicized the Global PIQA task through announcements on the Eleuther AI Discord, the LINGUIST List, Masakhane, X/Twitter, BlueSky, and LinkedIn.
We also identified NLP researchers with experience constructing benchmarks and language models for specific languages or language families, and we contacted them directly to broaden our reach.
We maintained a spreadsheet of interested volunteers (with contact information and languages spoken) to keep volunteers informed throughout the process.
We encouraged existing contributors to recruit additional volunteers for languages that were missing from the benchmark.
\item \textbf{Early feedback.} We allowed authors to send initial examples and preliminary versions of their datasets for feedback well before the dataset submission deadline.
This contrasts with traditional shared tasks at NLP conferences, where participants have minimal interaction with the organizers prior to submitting.
Furthermore, we held FAQ meetings one month before the deadline, held at multiple times to accommodate different time zones, and we maintained a consistently-updated set of slides with instructions and FAQs for creating the Global PIQA datasets.
\item \textbf{Data quantity.} We required a minimum of 100 examples per language for each submitted dataset. We found that this quantity was doable so as not to discourage researchers from participating, but large enough to ensure that researchers put significant thought into creating their datasets.
\item \textbf{Timeline and acceptances.} The shared task was publicly announced in late June 2025, with a submission deadline of September 15, 2025. This allowed almost three months to recruit contributors and for groups to develop datasets. The timeline was short enough, however, that no momentum was lost.
After the dataset submission deadline, we also continued to allow submissions for languages and dialects that were still missing from the benchmark. We individually reached out to volunteers who had signed up for specific missing languages, and in many cases, we were able to work out later deadlines that were more amenable to those authors.
In cases where an initial dataset submission did not meet quality checks (\S\ref{sec:compiling}), the dataset was not simply rejected; instead, we worked with the authors to make improvements for the dataset to be accepted.
Datasets were only rejected if those contributors did not respond to feedback or decided not to make required changes (e.g. adding examples to meet the minimum 100 examples per language, after filtering out examples as in \S\ref{app:cleaning} and \S\ref{app:annotation}), or due to other non-technical organizational challenges (\S\ref{app:sdn}).
\end{itemize}

\section{Non-Parallel Split: Cleaning, Compilation, and Sampling Details}
\label{app:dataset-creation-details}

\subsection{Non-Parallel Split: Cleaning and Compilation}
\label{app:cleaning}

As described in \S\ref{sec:non-parallel}, authors contributed datasets to the Global PIQA non-parallel split for their own language(s).
At minimum, each dataset contributed to Global PIQA contained a \texttt{prompt}, \texttt{solution0}, \texttt{solution1}, and \texttt{label} column.
For each dataset, we first removed exact duplicate examples and invalid examples where the two solutions were identical.
We normalized column names, moved supplemental information (e.g. ``topic'' fields or other columns added by individual groups) to a \texttt{supplement} column, and we converted all text fields to use UTF-8 text encoding.
For transparency, we annotated any examples that used LLMs to initially generate the example; this is a relatively small number of examples (9.6\% before subsampling, then 4.1\% in the official non-parallel split), and all examples are human validated before inclusion in Global PIQA (see method descriptions in \S\ref{app:descriptions}).
For several datasets, we found that sentence completion examples (i.e. examples where the prompt is an incomplete sentence, and the candidate solutions complete the sentence) contained prompts ending with ellipses (``\textit{...}'') or underscores (``\textit{\_\_\_}'', i.e. fill-in-the-blank). We removed these ending ellipses and underscores, as the completions are concatenated directly onto the prompts when fed into LLMs in the cloze evaluation setup (\S\ref{app:evaluation-details}).

As a preliminary check, we used Google Translate to translate a random subset of $\sim$20 examples per dataset, to identify any egregious errors (e.g. all examples far too easy, not following the task format, or large numbers of repetitive examples).
Based on this preliminary check, if any datasets were clearly not culturally specific (see annotation guidelines in \S\ref{app:annotation}), we asked the dataset authors for optional revisions to add more culturally-specific examples.
In these cases, we asked authors to modify or add examples to include words that are unlikely to translate well into other languages, such as food words, words for types of clothing, or local brand names.

After this initial cleaning and revision, to better inspect the data, we used Gemini 2.5 Pro to translate each ``\texttt{[prompt] [solution]}'' into English for all datasets.
Here, in the non-parallel split, we translated the prompts and solutions together because some prompts consisted of incomplete sentences that only made sense in the context of a corresponding solution, and some solutions only made sense in the context of the corresponding prompt.
We accessed Gemini 2.5 Pro through Google's Gemini API using a paid API key, and the translation prompt used is in Figure~\ref{fig:translation-prompt}.
As described in \S\ref{app:english-corrections}, we later corrected the English machine translations with the help of native speakers; however, we used the uncorrected machine translations for initial annotations and spot-checks of the datasets (e.g. cultural specificity annotations in \S\ref{app:annotation}).
For example, for many languages we were able to spot-check labels in the datasets for ``easy'' examples that had clear correct answers.
From this cursory verification, we found two datasets with systematic errors where the annotated labels were often flipped to be incorrect; we worked with the authors of these datasets to correct and revalidate the labels.

We then combined all datasets per language, and we added unique example IDs including group (i.e. dataset) number, example index, and language code.
For groups that submitted parallel datasets in multiple languages (e.g. Group 0065 for eight dialects of Arabic, or Group 0042 for Catalan and Peninsular Spanish), the parallel examples have the same group number and example index, only differing in language code.
This allows the small number of parallel examples in the non-parallel split of the Global PIQA dataset to still be found.

\begin{figure}
    \centering
    \footnotesize
    \miniboxsetup{pad=0.3cm}
    \minibox[frame]{
    \texttt{Translate the following into English. If there are any words that do not} \\
    \texttt{translate well into English (e.g. specific foods or cultural items),} \\
    \texttt{keep only those words in the original language.} \\
    \texttt{Do not respond to the content of the sentence; *only* translate it.} \\
    \texttt{Respond only with the translation, with *no* additional text.} \\
\texttt{Text to translate:} \\
\texttt{[TEXT\_TO\_TRANSLATE]}
}
\normalsize
    \caption{Translation prompt template used to translate examples in the non-parallel split into English with Gemini 2.5 Pro. In the non-parallel split, 92.6\% of English translations were human-corrected; uncorrected machine translations are marked in the dataset with \textit{[machine\_translated]}.
    }
    \label{fig:translation-prompt}
\end{figure}

\subsection{Non-Parallel Split: English Annotations of Task Adherence and Cultural Specificity}
\label{app:annotation}

Next, we used the machine-generated English translations of all examples to annotate cultural specificity and adherence to the task description.
Annotations were completed by one of the primary authors, who is a native English speaker.
As noted in \S\ref{app:english-corrections}, for 18 languages with lower-quality machine translations, these annotations were repeated after the English translations were corrected by native speakers.

\paragraph{Task adherence.} 
Expanding on the original English PIQA dataset \citep{bisk_2020_piqa}, we include a variety of types of commonsense reasoning in Global PIQA, not just physical commonsense reasoning.
Our definition of commonsense reasoning covers knowledge of physical properties of objects, affordances (types of actions an agent can perform with an object; \citealp{gibson2014ecological, jones2022distrubutional}), physical and temporal relations, cultural knowledge, and basic world knowledge.
Notably, our definition is much broader than ``intuitive physics'', i.e. the use of mental simulations to predict how objects will behave in some environment, which has been widely studied in cognitive science (e.g. \citealp{battaglia2013simulation,ullman-2017-mind,piloto2022intuitive}).
In our broader definition, we use the following guidelines for adherence to the task description:
\begin{enumerate}[leftmargin=0.5cm,itemsep=0.02cm,topsep=0.02cm]
\item Drop examples that consist of a complex or abstract logical problem, as these do not fit the task description of commonsense reasoning. For example, we drop complex logic puzzles and computer programming questions.
\item Drop examples that appear both generic and extremely easy based on the English translation. For example, we drop examples such as ``\textit{When you heat water, it becomes [hot/cold]}''.
\item Keep examples that query common knowledge about locations (e.g. locations of cities or famous monuments, or common events to observe in particular cities).
\item Keep examples that query social or cultural knowledge. These examples often describe regional customs, norms, and traditions.
\item Where possible, drop examples that query obscure historical factoids. In some languages, there are too few total examples to drop all such examples, so a small number of historical knowledge questions are still present in the dataset. These examples are generally apparent from their English translations (\S\ref{app:english-corrections}).
\end{enumerate}
Based on these guidelines, we dropped approximately 2K out of 31K examples in the submitted datasets, before subsampling for the official non-parallel split.
In cases where this filtering caused the number of examples in a language to drop below 100 examples, we worked directly with authors to reach the 100 example minimum.
We note that based on these guidelines, examples in Global PIQA cover a range of physical commonsense, social commonsense, cultural knowledge, and common knowledge.

\paragraph{Cultural specificity.}
Because cultural specificity is fairly subjective and perspective-dependent, we attempt to provide clear guidelines for when we annotated an example as ``culturally-specific''. Our definition of culturally-specific covers both culturally-\textit{specific} examples, i.e. examples that are only relevant in a specific region or language, and culturally-\textit{sensitive} examples, i.e. examples whose solution varies across regions or languages \citep{blend_2024_myung,singh-etal-2025-global}.
When we use the term ``culturally specific'', we refer to this broad definition.
We formulated the guidelines here in an attempt to reduce potential bias and the presence of stereotypes in our annotations of cultural specificity \citep{zhou-etal-2025-culture}.
We annotate examples for cultural specificity using these guidelines:
\begin{enumerate}[leftmargin=0.5cm,itemsep=0.02cm,topsep=0.02cm]
\item Some datasets have some examples marked as culturally specific by the dataset authors. We annotate these examples as culturally specific; this defers to the authors (members of the cultural communities) to choose examples that they believe reflect their culture, giving more ownership back to the communities themselves.
\item We annotate examples as culturally specific if they describe specific holidays, folklore, traditions, sayings, or aphorisms in the language.
\item We annotate an example as culturally specific if its solution likely varies by region. For example, traffic rules and social norms are likely to vary across regions.
\item If an example contains a word that does not translate well into English, then we annotate it as culturally specific. This can include words for local food dishes, traditional objects or articles of clothing, or local brands. We do \textit{not} count city names (or person names), as many examples that simply mention a city are not actually specific to that city. We acknowledge that some words are ambiguously ``English'' vs. borrowed from another language; in these cases, we use our best judgment based on how commonly the word is used in English.
\item We do \textit{not} count the presence of local ingredients or objects if they have widely used English words, such as corn, rice, beans, or many fruits and vegetables, even if these items vary in popularity across regions. In other words, we do not annotate an example as culturally specific solely based on the presence of these items. This guideline aims to reduce bias where some examples might otherwise be annotated as culturally specific based on stereotypical associations between specific foods and corresponding regions or cultures.
\item In cases where the English machine translation appears to be extremely low quality, such that the topic of the example is not clear, we use our best judgment based on the previous guidelines. We lean towards annotating cultural specificity in borderline cases, because we expect that machine translation systems are more likely to perform poorly in culturally-specific scenarios \citep{naveen2024overview,yao-etal-2024-benchmarking}.
\end{enumerate}
Through these annotations, we primarily aim to have a coarse filter for cultural specificity, such that we can up-sample culturally-specific examples in the following section.
In the full non-parallel dataset (i.e. before subsampling to the official split), 40.8\% of examples are annotated as culturally specific.
We note that even when marked as culturally specific, many examples do not actually require knowledge of the referenced culturally-specific item or tradition to correctly answer the prompt; in many cases, the culturally-specific element is referenced, but the correct answer can be inferred naively from the rest of the context.

\subsection{Additional Submission Selection Criteria} \label{app:sdn}

A prior version of this work listed co-authors at institutions on the US SDN list and used data provided by such individuals. Over the course of this project, policies at major machine learning publication venues were clarified, making it impossible to undergo peer review when collaborating with such individuals. Consequently, the current version of this work has been re-done without their contributions and they are not listed as authors. 

\subsection{Non-Parallel Split: Subsampling to the Official Split}
\label{app:sampling}

Before subsampling, the Global PIQA non-parallel split is highly skewed across languages.
For example, before subsampling, Hindi (\texttt{hin\_deva}) and Yoruba (\texttt{yor\_latn}) have 1.4K examples each, while many other languages have close to the minimum dataset submission requirement of 100 examples.
The full dataset before subsampling is available at \url{https://huggingface.co/datasets/mrlbenchmarks/global-piqa-nonparallel} in the \texttt{unsampled\_full} folder, containing 29.1K examples.
Due to the imbalance across languages, we select a subsample of 100 diverse and maximally culturally-specific examples in each language as the official non-parallel split of Global PIQA. This enables efficient evaluations of state-of-the-art LLMs across all languages in Global PIQA.

When filtering, we apply the following stages per language; we continue to the next stage unless that stage would cause the dataset for the language to fall below 100 examples. This allows us to maximize the quality and diversity of the examples for each language while still maintaining at least 100 examples per language. We note that the extremely low quality examples and off-task examples were already filtered out by the cleaning and annotations in \S\ref{app:cleaning} and \S\ref{app:annotation}.
We apply the following filtering stages in order (or until reaching 100 examples in the language):
\begin{enumerate}[leftmargin=0.5cm,itemsep=0.02cm,topsep=0.02cm]
\item We remove any duplicate prompts, i.e. examples that have the same prompt but different pairs of solutions. This is generally a very small number of examples (e.g. one or two examples), and zero examples for most datasets. This filtering step drops a total of 60 examples across all languages. Note that exact duplicate examples (i.e. same prompt and same solutions) were already removed in \S\ref{app:cleaning}.
\item  We filter out examples where the two candidate solutions differ in length by more than 25 English byte equivalents (this is roughly the same as English character equivalents, because most English characters are one byte in UTF-8). We compute English byte equivalents by computing the solution lengths first in raw UTF-8 bytes, then dividing by the language's byte premium \citep{arnett2024bit}, which is the estimated number of bytes used to encode text in the language compared to content-matched (parallel) text in English.
We perform this filtering step to attempt to minimize any length biases in the dataset, where longer solutions might be assigned systematically lower probabilities than shorter solutions by pretrained-only models, leading to a bias towards shorter solutions for those models.
This filtering step drops a total of 1.6K examples across all languages. 
\item We filter out examples whose non-stopword tokens overlap by more than 50\% with another example in the dataset.
Specifically, we tokenize all examples using the Goldfish tokenizer for the language \citep{chang-etal-2024-goldfish}. For the 24 Global PIQA languages not covered by the 350 languages in Goldfish, we use a simple space-based tokenizer after removing common punctuation symbols; all Global PIQA languages without a Goldfish tokenizer use scripts that separate words with spaces. Upon tokenizing all examples, we define stopword tokens as tokens that appear in at least 25\% of examples for the language.
Then, we sort examples by length (in order to give longer examples priority), and we loop through all examples, dropping any examples in which greater than 50\% of its non-stopword tokens are contained in another previously-encountered example.
This filtering step aims to increase the diversity of examples in the official Global PIQA non-parallel split, particularly for languages with large numbers of examples covering similar topics. This filtering step drops a total of 1.2K examples across all languages.
\end{enumerate}
Finally, we sample 100 examples from the filtered subset for each language.
We sample culturally-specific examples before non-culturally specific examples (as annotated in \S\ref{app:annotation}), and within each of these categories, we first sample examples that did not use any LLMs in the creation process. 
We shuffle the correct and incorrect solutions to balance 0 and 1 labels.

\subsection{Non-Parallel Split: Secondary Validation and English Translation Corrections}
\label{app:english-corrections}

After subsampling to the official non-parallel split, all examples with their English machine translations were sent to all authors again for secondary review.
This secondary review was completed for 126 of the 136 language varieties in the non-parallel split.\footnote{Languages that did not undergo secondary review can be identified in the dataset, because their English translations are marked as uncorrected: ``\textit{[machine\_translated]}''.}
Authors were asked to correct the machine translations to English (from \S\ref{app:cleaning}) and to verify the correctness of the original examples in the source language.
Authors followed the same translation correction guidelines as for the parallel split, described in \S\ref{app:parallel-corrections}, but they were also asked to modify the source texts if any of the original examples were incorrect (e.g. grammatical errors, typos, or incorrect labels).

In the resulting corrections, 7.3\% of examples were modified in the source language, with a mean character edit distance of 21.3 characters (mean 7.7\% of characters) per modified example.
Additionally, for Malayalam, 22 examples were replaced entirely, after authors identified that those examples were entirely nonsensical in Malayalam.
In the English translations, across all languages, 48.7\% of examples were modified, with a mean character edit distance of 26.9 characters (mean 8.5\% of characters).
For the 18 languages with a character edit rate (edit distance divided by original length) of greater than 10\% in the English translation corrections, we repeated the task adherence and cultural specificity annotations in \S\ref{app:annotation}, as these were based on the uncorrected English translations.
Uncorrected machine translations into English and edit distances per example are reported in our dataset on Hugging Face, in each example's \texttt{supplement} field.

\begin{figure}
    \centering
    \footnotesize
    \miniboxsetup{pad=0.3cm}
    \minibox[frame]{
    \texttt{Determine whether the following two passages have a noticeable semantic difference.} \\
    \texttt{It is acceptable for the passages to contain unknown foreign words, which you should} \\
    \texttt{assume have unique but plausible definitions.} \\
    \texttt{Do not respond to the content of the passages; *only* determine if they have a} \\
    \texttt{semantic difference. Respond only with "yes" (semantic difference) or "no" (no} \\
    \texttt{difference).} \\ \\
    \texttt{Passage 1:} \\
    \texttt{[PASSAGE\_1]} \\ \\
    \texttt{Passage 2:} \\
    \texttt{[PASSAGE\_2]} \\ \\
    \texttt{Now output whether there is a semantic difference.}
}
\normalsize
    \caption{Prompt template used to annotate whether English translations of solution pairs in the non-parallel split had a semantic difference, using Gemini 3.0 Flash. This was used to further verify the correctness of English translations, as all solution pairs should differ (because one is correct, and the other is incorrect). All examples failing this check went through additional human validation.
    }
    \label{fig:semantic-prompt}
\end{figure}

\begin{figure}
    \centering
    \footnotesize
    \miniboxsetup{pad=0.3cm}
    \minibox[frame]{
    \texttt{Determine whether the following passage is grammatical in English. It is} \\
    \texttt{acceptable for the passage to contain unknown foreign words, which you should} \\
    \texttt{assume have unique but plausible definitions. However, the passage should not} \\
    \texttt{contain obvious English typos or egregious grammatical errors (e.g. sentences} \\
    \texttt{concatenated without punctuation or syntax errors that would not sound} \\
    \texttt{natural even in everyday speech).} \\
    \texttt{Do not respond to the content of the passage; *only* determine if it has a} \\
    \texttt{grammatical error or typo.} \\
    \texttt{Respond only with "no" (no grammatical error) or "yes" (contains an error).} \\
    \texttt{Do *not* include any other text in your response.} \\ \\
    \texttt{Passage:} \\
    \texttt{[PASSAGE]} \\ \\
    \texttt{Now output either "no" (no errors) or "yes" (contains an error).}}
\normalsize
    \caption{Prompt template used to annotate whether English translations of examples in the non-parallel split were grammatical, using Gemini 3.0 Flash. All English translations failing this check went through additional manual editing.}
    \label{fig:grammatical-prompt}
\end{figure}

To further verify the correctness of the English translations, we used Gemini 3.0 Flash to annotate any English solution pairs with no ``noticeable semantic difference'' (prompt template in Figure~\ref{fig:semantic-prompt}), as all solution pairs should have some difference (because one is correct, and the other is incorrect).
Any examples failing this check (approximately 2\% of examples) went through additional human validation with native speakers of the original language.
Finally, because many of our authors are not native English speakers, we used Gemini 3.0 Flash to identify any English translations that were ungrammatical after correction (prompt template in Figure~\ref{fig:grammatical-prompt}).
Any examples failing this check were edited by a native English speaker, consulting with authors who spoke the original language of the example if needed.
This process was intended to make examples in the non-parallel split understandable to a broader audience, through accurate and grammatical English translations.

\section{Parallel Split: Construction and Correction Details}
\label{app:parallel-details}

\subsection{Parallel Split: Constructing Examples in English}
\label{app:parallel-construction}

As described in \S\ref{sec:parallel}, examples in the parallel split of Global PIQA were first written by two native English speakers.
Each example in the parallel split consists of a prompt (question) and four candidate solutions.
We chose the question-answer format because most modern LLMs are instruction-tuned for question answering rather than text completion \citep{ouyang2022training,wei2022finetuned}.
Perhaps more importantly, it is not always possible to directly translate prompt-completion examples to languages with different sentence structures, because information appears in different orders across languages.
For example, a subject-object prompt with a verb completion in an SOV language would have a very different format in a VSO language.
The question-answer format ensures that each prompt is a standalone question, facilitating easier translation.

We wrote 109 questions with candidate solutions, drawing inspiration from previous commonsense reasoning datasets: the original English PIQA (physical commonsense; \citealp{bisk_2020_piqa}), EWoK (basic world knowledge; \citealp{ewok_2025_ivanova}), TRAM (temporal reasoning; \citealp{wang-zhao-2024-tram}), PROST (physical reasoning; \citealp{aroca-ouellette-etal-2021-prost}), \citet{glenberg2000symbol} (object affordances), and HellaSwag (commonsense reasoning in physical situations; \citealp{zellers-etal-2019-hellaswag}).
The original English PIQA is released under an Academic Free License v. 3.0 license. TRAM and HellaSwag are released under an MIT license. PROST is released under an Apache-2.0 license. The stimuli from \citet{glenberg2000symbol} are not available publicly, and we did not use any items directly in the creation of the parallel split.

We also collected a small number of difficult examples from the non-parallel split of Global PIQA translated into English (\S\ref{app:english-corrections}), by filtering for examples that GPT-5, Gemini 2.5 Pro, or Claude Sonnet 4.5 answered incorrectly; in this filtering, we only considered examples with unambiguously correct answers.
Each example in the parallel split is annotated with the dataset from which it drew loose inspiration, or labeled as entirely novel.

When writing examples, we avoided writing extremely easy examples such as ``\textit{When you put water in a refrigerator, it becomes [hot/cold]}'', and we avoided including references to culturally-specific elements such as local foods or customs, to facilitate translation to a large number of languages.
We also wrote examples to cover a variety of types of commonsense reasoning: object properties and interactions (67 examples), temporal reasoning (13 examples), basic counting (13 examples), spatial reasoning (25 examples), and affordances (23 examples).
Some examples covered multiple commonsense reasoning types, and all examples are annotated with their reasoning types in the released parallel split.
In the end, we wrote 109 questions in English;
after translation corrections, these were filtered to 103 examples (or 101 examples for Ekpeye; \S\ref{sec:parallel}).

\subsection{Parallel Split: Machine Translations}
\label{app:parallel-translation}

\begin{figure}
    \centering
    \footnotesize
    \miniboxsetup{pad=0.3cm}
    \minibox[frame]{
    \texttt{Translate the following into [TARGET\_LANGUAGE].} \\
    \texttt{Do not respond to the content of the question or phrase; *only* translate it.} \\
    \texttt{Respond only with the translation, with *no* additional text.} \\
    \texttt{Text to translate:} \\ \\
    \texttt{[TEXT\_TO\_TRANSLATE]} \\
}
\normalsize
    \caption{Translation prompt template used to translate examples into each target language for the parallel split, using Gemini 2.5 Pro and Gemini 3.0 Flash. All translations for the parallel split were human-corrected.
    }
    \label{fig:parallel-translation-prompt}
\end{figure}

Next, we translated all examples written for the parallel split into the 131 language varieties covered in the split.
These languages were selected based on availability of the authors who spoke those languages, along with additional recruitment after releasing the non-parallel split. Additional recruitment for the parallel split was primarily done through the connections of authors for the non-parallel split. For each target language, the first 50 examples in the parallel split were translated using Gemini 2.5 Pro. The remaining examples were translated using Gemini 3.0 Flash.
The prompt template used is in Figure~\ref{fig:parallel-translation-prompt}. Each prompt and candidate solution was translated separately, as we found a greater number of hallucinations when translating prompts and candidate solutions together (e.g. the LLM correcting incorrect candidate solutions if the prompt was included).
However, we note that our approach of translating prompts and solutions separately introduced some inconsistencies between translations for prompts and solutions, e.g. for gender agreement and formatting of numerals in years (e.g. ``\textit{2026}'' vs. ``\textit{two thousand and twenty six}'').
Additionally, in some languages, the writing system used in translations for different examples was inconsistent.
These inconsistencies were largely fixed during the translation correction stage.

\subsection{Parallel Split: Translation Corrections}
\label{app:parallel-corrections}

Finally, we sent the machine-translated examples to all authors for correction.
Each source text with its machine translation was presented in a row in a Google Sheet (one sheet per language), and authors were asked to correct the machine translations.
To mark each row as corrected, authors were asked to verify that (1) the meaning of the translation was correct, (2) the text sounded natural in the target language, and (3) the initially-labeled ``correct'' solution (always listed first) was indeed the correct solution, with all other candidate solutions being incorrect, even after translation into the target language.
Full translation correction guidelines sent to authors are on our \href{https://github.com/mrlbenchmarks/global-piqa}{GitHub}.
To prevent duplicate work from multiple native speakers per example, each row was marked as corrected after at least one native speaker reviewed it.

When processing these translation corrections, we observed several common issues in the corrected sheets. First, sheet rows were sometimes swapped (e.g. by dragging rows in the Google Sheet), or corrected translations were pasted or typed into the incorrect cells.
In some cases, authors initially corrected the source texts to match the meaning of the machine translations, rather than correcting the machine translations to match the source texts.
We detected these issues using edit distances between texts before and after correction (e.g. a high edit distance to the uncorrected text, but a low edit distance to a different example, or edits only to the sources and not the translations).
We also identified a small number of texts which were accidentally deleted instead of corrected, or examples where two of the solutions were identical after correction.

As a sanity check, we also checked all occurrences of special characters that might be artifacts of LLM translation (e.g. `<', `>', `[', `]', `(', `)', and `*') and any substrings longer than 20 characters of the original translation prompts.
For Arabic dialects, we removed diacritics using the \texttt{dediac\_ar} function from CAMeL Tools \citep{alhafni-etal-2023-advancements}, based on feedback from authors who are native speakers of Arabic.
In all of the cases above, along with cases where the authors left a note in an optional notes column, we manually checked those examples, in consultation with the authors who spoke that target language if needed.

\section{Non-Parallel Split: \textit{Ad Hoc} Human Evaluations}
\label{app:human-eval}
We do not explicitly perform a human evaluation study due to the substantial resources that it would take to run a study for the large number of languages involved in Global PIQA. However, several groups reported human evaluations on their dataset contributions to the non-parallel split, where a native speaker was asked to choose correct solutions without access to the ``ground truth'' labels, or inter-annotator agreement percentages were reported (from which we can compute an analogy to human ``accuracy'' by treating the other annotator's labels as the ``ground truth'').
On top of this, we conducted \textit{ad hoc} human evaluation with one author (a native speaker of Mandarin Chinese) for the Mandarin Chinese datasets (simplified and traditional Chinese characters, \texttt{cmn\_hans} and \texttt{cmn\_hant}) in the Global PIQA official non-parallel split, after observing somewhat low scores in the language for some models (e.g. GPT-5 with less than 90\% accuracy, given that Mandarin Chinese is a high-resource language). Accuracies for individual human annotators for the 12 language varieties with available human results are shown in Table~\ref{tab:human-eval}.

\begin{table}[t]
    \caption{Ad hoc human evaluations, showing accuracies for individual human annotators for various languages, on individual dataset contributions to the Global PIQA non-parallel split. Details in \S\ref{app:human-eval}.}
    \label{tab:human-eval}
    \centering
    \footnotesize
    \begin{tabular}{lr|lr}
    \toprule
    Language & Acc. & Language & Acc. \\
    \midrule
    Slovenian (\texttt{slv\_latn}) & 97\% & Croatian (\texttt{hrv\_latn}) & 100\% \\
    Serbian (\texttt{srp\_latn}) & 97\% & Macedonian (\texttt{mkd\_cyrl}) & 92\% \\
    Catalan (\texttt{cat\_latn}) & 94\%, 95\%, 98\% & Estonian (\texttt{ekk\_latn}) & 95\% \\
    Tamil (\texttt{tam\_taml}) & 95\% & European Portuguese (\texttt{por\_latn\_port}) & 91\%, 95\% \\
    Algerian Arabic (\texttt{arq\_arab}) & 95\% & Moroccan Arabic (\texttt{ary\_arab}) & 95\% \\
    Mandarin (\texttt{cmn\_hans}) & 95\% & Mandarin (\texttt{cmn\_hant}) & 93\% \\
    \bottomrule \\
    \end{tabular}
    \normalsize
    \vspace{0.1cm}
\end{table}

In these ad hoc human evaluations, mean human annotator accuracy was 95.1\%, and none of the fifteen individual annotators had accuracy below 91\%.
Of course, we note that there is likely some sampling bias, where dataset authors who chose to run human evaluations were also more likely to construct high quality datasets in the first place.
That said, we even observe high accuracies for Mandarin Chinese (95\% and 93\%), in which we ran our ad hoc human evaluation after dataset submissions and compilation, independent of the dataset authors.
These results suggest that human accuracy on the Global PIQA non-parallel split is likely to be at least 90\%, and potentially as high as 95\%.
After running these ad hoc evaluations, examples were updated based on disagreeing labels.

\begin{figure}[t]
    \centering
    \footnotesize
    \miniboxsetup{pad=0.3cm}
    \minibox[frame]{
    \texttt{Given the following situation, which option is more likely to be correct?} \\ \\
\texttt{Situation:} \\
\texttt{[PROMPT]} ... \\ \\
\texttt{Option A: [SOLUTION0]} \\ \\
\texttt{Option B: [SOLUTION1]} \\ \\
\texttt{Your response should end with "The best answer is: [answer\_letter]" where} \\
\texttt{[answer\_letter] is one of A or B.}
    }

    \vspace{0.3cm}

    \minibox[frame]{
\texttt{[PROMPT]} \\ \\
\texttt{Option A: [SOLUTION0]} \\ \\
\texttt{Option B: [SOLUTION1]} \\ \\
\texttt{Option C: [SOLUTION2]} \\ \\
\texttt{Option D: [SOLUTION3]} \\ \\
\texttt{Your response should end with "The best answer is: [answer\_letter]" where} \\
\texttt{[answer\_letter] is one of A, B, C, or D.}
    }
    
\normalsize
    \caption{Prompt templates for the generation-style evaluation format (\S\ref{sec:eval-format}), for the non-parallel (top) and parallel (bottom) split of Global PIQA. In the parallel split, all prompts are formatted as questions, and there are four candidate solutions.
    }
    \label{fig:generation-prompt-templates}
\end{figure}

\section{Evaluation Details}
\label{app:evaluation-details}

As described in \S\ref{sec:eval-format}, we evaluate pretrained-only models with the cloze evaluation format and instruction-tuned models with the generation evaluation format.
For the cloze evaluations, we normalize solution log-probabilities by the length of each solution in bytes; we do not need to use English byte equivalents normalized per language (``byte premiums''; \citealp{arnett2024bit}), because for each example, this would only divide each log-probability by a constant, leaving the ranking of solutions unchanged.
Both the cloze and generation evaluation formats are implemented on the LM Evaluation Harness, with our implementation available at: \url{https://github.com/mrlbenchmarks/lm-eval-global-piqa} (to be merged into the official LM Evaluation Harness branch soon).

For the generation evaluations, we use the prompt templates in Figure~\ref{fig:generation-prompt-templates}, generating a maximum of 2048 response tokens per question.
For open and open-weight models, we sample using temperature $\tau=0.90$ and top-$p=0.80$; for closed models, we use temperature $\tau=0.90$.
For closed systems with ``thinking" models, we allow up to 1024 thinking tokens, with the remaining 1024 token budget allocated to the response.
Gemini 3.1 and Claude Sonnet 4.6 allow setting this directly, but GPT-5.4 only supports ``low'', ``medium'', and ``high'' thinking. We use ``medium'' thinking for GPT-5.4 with a total generation length of 2048 tokens, to be comparable to the other models.
For a small number of responses (about 1-3\% of responses, primarily in lower-resource languages), GPT-5.4 (Regular, Mini, and Nano) exceeded its token budget, and we reran it with ``low'' thinking.
After sampling responses, we used string matching (e.g. ``best answer is: A'' or ``best answer is: B'', as specified in the prompt template instructions) to mark answers as correct or incorrect.

All open-weight models were run using 8x A6000 or 8x A100 (80GB each) for a total of 1480.4 GPU hours (367.3 A100 hours and 1113.1 A6000 hours). 
Running Claude Sonnet 4.6 cost \$61.23 for 4.9M input tokens and 7.2M output tokens.
Running the OpenAI models cost \$97.04 for 11.0M input tokens and 25.9M output tokens.
Running the Gemini models cost \$155.45 for 14.6M input tokens and 75.7M output tokens.

\subsection{Full List of Models} \label{app:model_list}

We evaluate several large well-known LLMs on Global PIQA, along with a wide variety of open and open-weight models.
We prioritize models that were requested by the authors of the datasets, and we prioritize models pretrained from scratch over adapted and fine-tuned models. 
We did not evaluate Claude Opus because we estimated the cost at \$400, which is more than the cost of running all the other closed systems combined.
Due to hardware limitations, for open-weight models, we focus on dense models ranging from 300M to 120B parameters. All raw evaluation scores for each model (results by language and non-parallel vs. parallel split) are available on our \href{https://github.com/mrlbenchmarks/global-piqa}{GitHub}.
In total, we evaluate Global PIQA on 146 models, including 
seven closed models and 139 open-weight models:
\begin{itemize}[leftmargin=0.5cm,itemsep=0.0cm,topsep=0.02cm,nosep]
\item Claude Sonnet 4.6 \citep{anthropic2026claude_sonnet46_system_card}
\item Gemini 3.1 Pro \citep{google2025gemini3_1_pro_modelcard}, 3.0 Flash \citep{google2025gemini3_flash}, and 3.1 Flash-Lite \citep{google2025gemini3_1_flash_lite}
\item GPT-5.4 (Regular, Mini, and Nano; \citealp{openai2025gpt5_4systemcard}) \\
\item APT3 1B \citep{AzurroAPT3Base1B} (CC BY-NC 4.0 license)
\item Apertus 8B and 70B \citep{hernandez2025apertus} (Apache 2.0 license)
\item Aya Expanse \citep{dang2024aya} (CC BY-NC 4.0 license)
\item BLOOM 560M, 1.1B, 1.7B, 3B, and 7.1B \citep{workshop2022bloom} (BigScience RAIL License v1.0)
\item Babel 9B and 83B \citep{zhao2025babel} (SeaLLM license\footnote{\url{https://huggingface.co/SeaLLMs/SeaLLM-13B-Chat/blob/main/LICENSE}})
\item Bielik v3 1.5B and 4.5B \citep{ociepa2025bielikv3smalltechnical} (Apache 2.0 license)
\item Command R/R+ 7B and 32B \citep{cohere2025command}, Command A \citep{cohere2025commandaenterprisereadylarge} (CC BY-NC 4.0 license)
\item Croissant LLM v0.1 1B \citep{faysse2025croissantllm} (MIT license)
\item DeepSeek R1 Distill Qwen 1.5B, 7B 14B, and 32B \citep{deepseekai2025deepseekr1incentivizingreasoningcapability} (MIT license)
\item EXAONE 3.5 7.8B and 32B \citep{an2024exaone}, EXAONE 4 1.2B and 32B \citep{bae2025exaone} (Exaone license\footnote{\url{https://huggingface.co/LGAI-EXAONE/EXAONE-3.5-7.8B-Instruct/blob/main/LICENSE}})
\item EuroLLM 9B \citep{martins2025eurollm} (Apache 2.0 license)
\item Falcon 7B and 40B \citep{falcon40b} (Apache 2.0 license)
\item GPT-SW3 1.3B, 6.7B, and 20B \citep{ekgren-etal-2024-gpt} (AI Sweden's LLM AI Model License Agreement\footnote{\url{https://huggingface.co/AI-Sweden-Models/gpt-sw3-126m/blob/main/LICENSE}}) 
\item GPT-oss 20B and 120B \citep{agarwal2025gpt} (Apache 2.0 license)
\item Ganda Gemma\footnote{\url{https://huggingface.co/CraneAILabs/ganda-gemma-1b}} and Swahili Gemma\footnote{\url{https://huggingface.co/CraneAILabs/swahili-gemma-1b}} (Gemma license\footnote{\url{https://ai.google.dev/gemma/terms}})
\item Gemma 2 2B, 9B, and 27B \citep{team2024gemma}, Gemma 3 270M, 1B, 4B, 12B, and 27B \citep{team2025gemma}, Gemma 4 5B, 8B, and 31B \citep{gemma4_2025} (Gemma license\footnote{\url{https://ai.google.dev/gemma/terms}} and Apache 2.0 license for Gemma 4)
\item Gemma SEA-LION v3 9B and Llama SEA-LION 8B, 70B \citep{ng2025sea} (Gemma license\footnote{\url{https://ai.google.dev/gemma/terms}} and Llama 3.1 license\footnote{\url{https://huggingface.co/meta-llama/Llama-3.1-70B-Instruct/blob/main/LICENSE}})
\item Gromenauer 7B\footnote{\url{https://huggingface.co/bertin-project/Gromenauer-7B}} (Apache 2.0 license)
\item HyGPT 10B \citep{Gen2B2025HyGPT} (HyGPT Permissive Use License\footnote{\url{https://huggingface.co/Gen2B/HyGPT-10b/blob/main/LICENSE}})
\item HyperCLOVA X 500M and 1.5B \citep{yoo2024hyperclova} (HyperCLOVA X SEED Model License Agreement\footnote{\url{https://huggingface.co/naver-hyperclovax/HyperCLOVAX-SEED-Text-Instruct-0.5B/blob/main/LICENSE}})
\item InkubaLM \citep{tonja2024inkubalm} (CC BY-NC 4.0 license)
\item Kanana 1.5 2.1B \citep{bak2025kanana} (Apache 2.0 license) 
\item Komodo 7B \citep{owen2024komodo} (Llama 2 license\footnote{\url{https://huggingface.co/meta-llama/Llama-2-7b-chat-hf/blob/main/LICENSE.txt}})
\item Llama 3.1 8B and 70B, Llama 3.2 1B and 3B \citep{meta2024llama3} (Llama 3.1 license\footnote{\url{https://huggingface.co/meta-llama/Llama-3.1-70B-Instruct/blob/main/LICENSE}} and Llama 3.2 license\footnote{\url{https://huggingface.co/meta-llama/Llama-3.2-1B/blob/main/LICENSE.txt}})
\item Llama Krikri 8B \citep{roussis2025krikriadvancingopenlarge} (Llama 3.1 license\footnote{\url{https://huggingface.co/meta-llama/Llama-3.1-70B-Instruct/blob/main/LICENSE}})
\item Meltemi v1.5 7B \citep{voukoutis2024meltemiopenlargelanguage} (Apache 2.0 license)
\item Minerva\footnote{\url{https://huggingface.co/collections/sapienzanlp/minerva-llms}} 1B, 3B, and 7B (Apache 2.0 license)
\item Mistral v0.1 7B, Mistral v0.3 7B, Mistral Small, and Mixtral v0.1 \citep{Jiang2023Mistral7B, jiang2024mixtral} (Apache 2.0 license)
\item PersianMind v1.0 \citep{persianmind} (CC BY-NC-SA 4.0 license)
\item Phi-3 medium and mini instruct \citep{abdin2024phi3}, Phi-3.5 mini instruct, and Phi-4 full and mini instruct \citep{abdin2024phi} (MIT license)
\item Poro 2 8B and 70B \citep{poro2_2025} (Llama 3.1 license)
\item Qwen 2.5 500M 1.5B, 3B, 7B 14B, 32B, and 72B \citep{yang2024qwen2_5}, Qwen 3 600M, 1.7B, 4B, 8B, 14B, and 32B \citep{qwen3technicalreport}, Qwen 3.5 800M, 2B, 4B, 9B, and 27B \citep{qwen3.5} (Apache 2.0 license)
\item Sailor2 1B, 8B, and 20B \citep{sailor2report} (Apache 2.0 license)
\item Sahabat AI 70B \citep{koto-etal-2023-large} (Llama 3.1 license)
\item Salamandra 2B and 7B \citep{gonzalezagirre2025salamandratechnicalreport} (Apache 2.0 license)
\item Sarvam-m \citep{SarvamM2025} (Apache 2.0 license)
\item SeaLLMs v3 1.5B and 7B \citep{zhang-etal-2025-seallms} (SeaLLM license\footnote{\url{https://huggingface.co/SeaLLMs/SeaLLM-13B-Chat/blob/main/LICENSE}})
\item Tiny Aya \citep{salamanca2026tinyayabridgingscale} (CC BY-NC 4.0 license)
\item TowerBase and TowerInstruct v0.1 7B and 13B \citep{tower_llm_2024} (CC BY-NC 4.0 license)
\item Tucano 1.1B and 2.4B \citep{correa2024tucano} (Apache 2.0 license)
\item vinaLlama 2.7B and 7B \citep{nguyen2023vinallama} (Llama 2 license)
\item Viking 7B\footnote{\url{https://huggingface.co/LumiOpen/Viking-7B}} and 13B\footnote{\url{https://huggingface.co/LumiOpen/Viking-13B}} (Apache 2.0 license)
\item XGLM 564M, 1.7B, 2.9B, 4.5B, and 7.5B \citep{lin2021few} (MIT license)
\end{itemize}

\clearpage

\section{Additional Results}
\label{app:additional-results}

\subsection{Parallel Split Pareto Frontier Plot}
\label{app:pareto-frontier-parallel}
\begin{figure}[h!]
    \centering
    \includegraphics[width=0.95\linewidth]{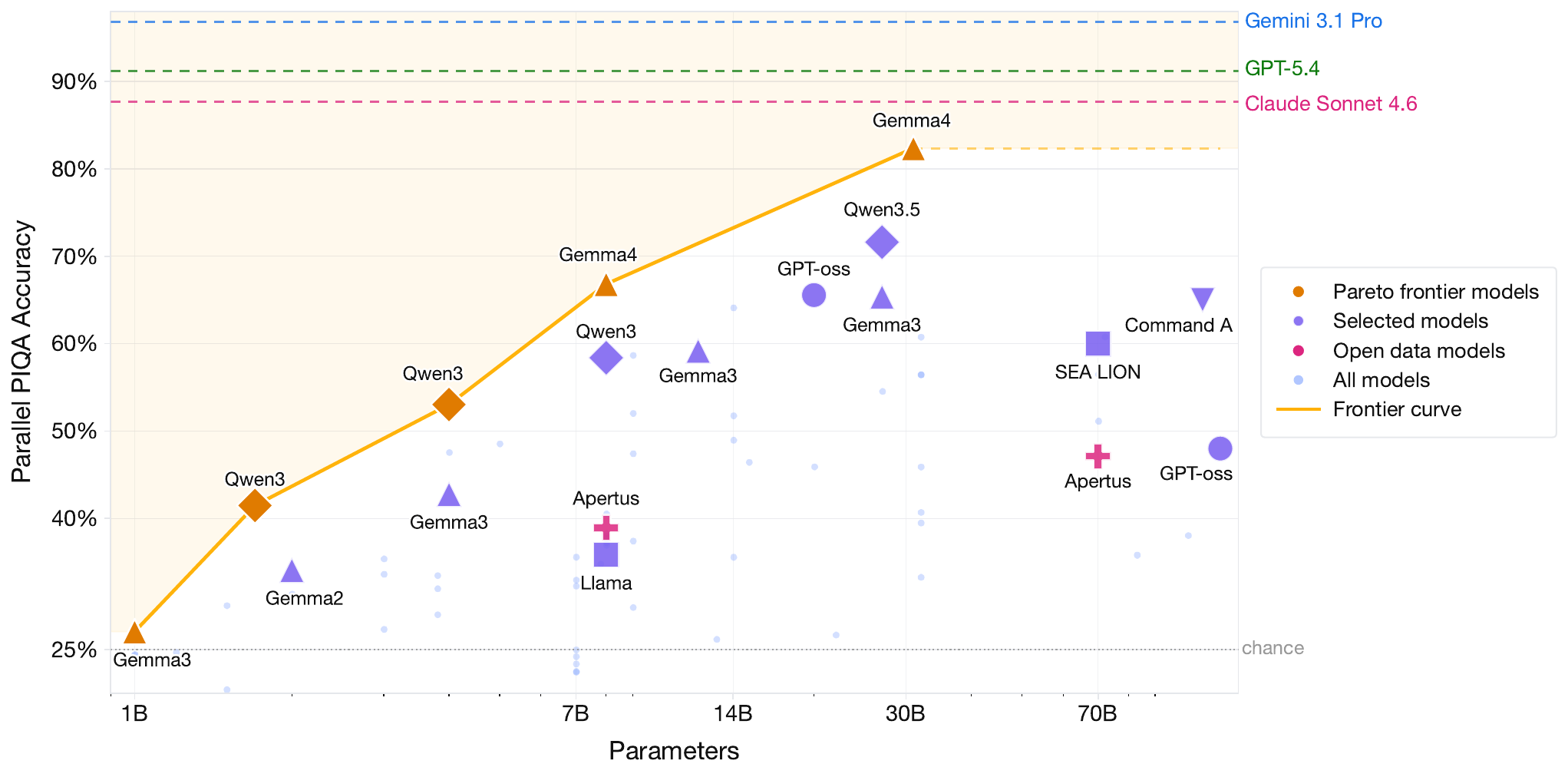}
    \caption{Accuracy averaged across all languages vs. parameter count for open-weight models, on the parallel split of Global PIQA. All evaluations here use the generation-style task format. We display names of top-performing models. Shape indicates model family, and color indicates openness (open-weight in purple vs. fully open in pink, including open data). All other models are plotted as dots. Chance performance (25\%) and performance of top closed systems are plotted as dashed lines.}
    \vspace{-0.5em}
    \label{fig:pareto-frontier-parallel}\end{figure}

The Pareto frontier plot for the Global PIQA parallel split (analogous to Figure~\ref{fig:pareto-frontier} for the non-parallel split) is in Figure~\ref{fig:pareto-frontier-parallel}.

\subsection{Results Per Region for All Models} \label{app:by_region}

\begin{figure}[t]
    \centering
    \includegraphics[width=0.95\linewidth]{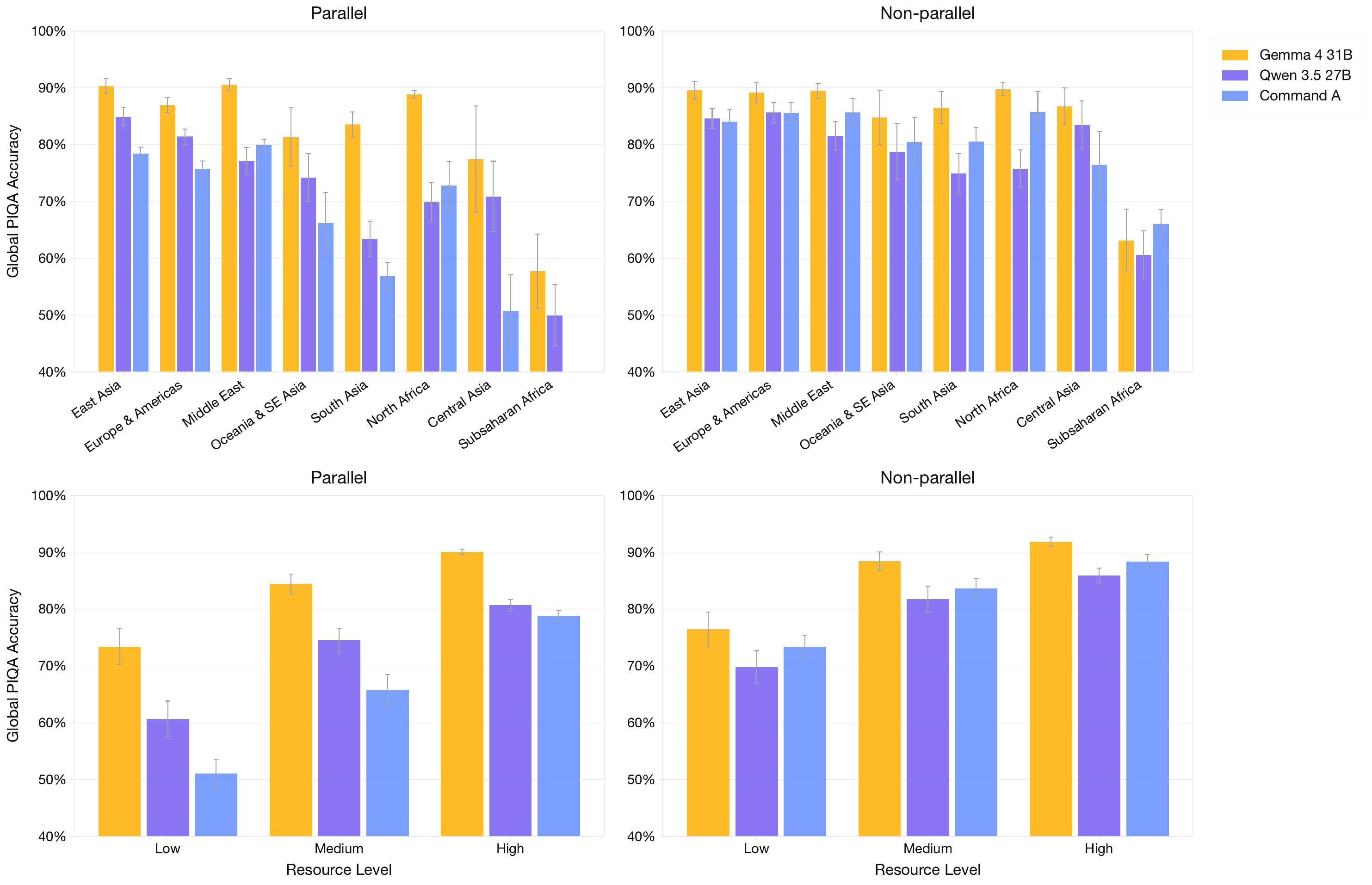}
    \caption{Parallel (left) and non-parallel (right) performance for top open-weight models, aggregating languages by geographic region (\S\ref{app:by_region}; top) and resource level (\citealp{joshi-etal-2020-state}; bottom). Error bars indicate one standard error of the mean.}
    \label{fig:open_2x2}
\end{figure}

\begin{figure}[t]
    \centering
    \includegraphics[width=0.95\linewidth]{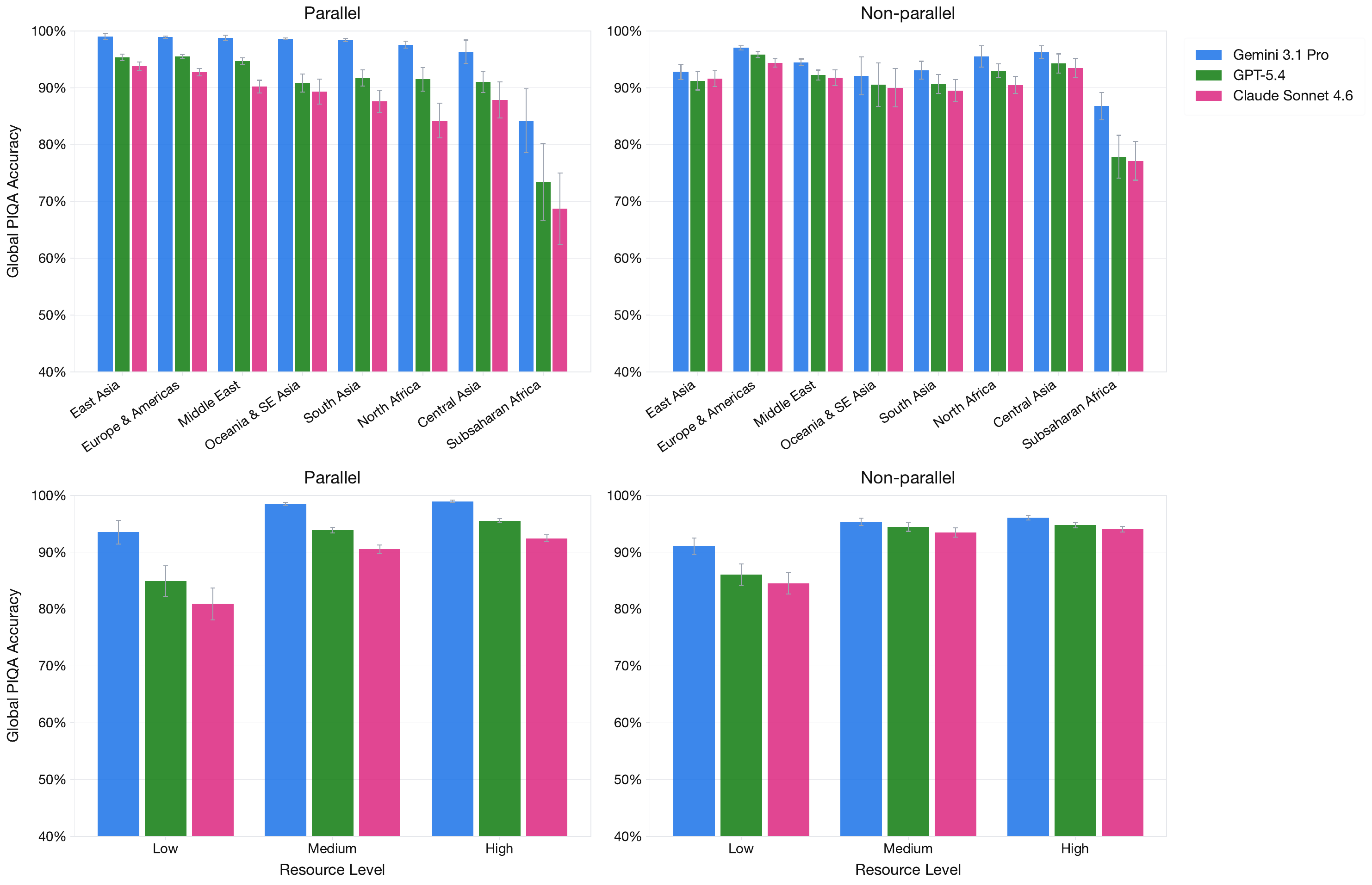}
    \caption{Parallel (left) and non-parallel (right) performance for top closed systems, aggregating languages by geographic region (\S\ref{app:by_region}; top) and resource level (\citealp{joshi-etal-2020-state}; bottom). Error bars indicate one standard error of the mean.}
    \label{fig:closed_2x2}
\end{figure}

Figures \ref{fig:open_2x2} and \ref{fig:closed_2x2} show average performance per region and per resource level for all models we evaluated.
For resource levels, we use the language resource level classifications from \citet{joshi-etal-2020-state}, mapping 0 and 1 to ``low-resource'', 2 and 3 to ``medium-resource'', and 4 and 5 to ``high-resource''.
For regions, we group the North American and South American languages with the European languages in the figures, because all languages from the Americas in Global PIQA are originally
European languages that are now spoken in the Americas.
In our results, the languages in each region are defined as:

\begin{itemize}[leftmargin=0.5cm,itemsep=0.0cm,topsep=0.02cm]
\item \textbf{Central Asia}: \texttt{bxr\_cyrl} (Russian Buryat), \texttt{kaz\_cyrl} (Kazakh), \texttt{kir\_cyrl} (Kyrgyz), \texttt{uig\_arab} (Uighur), \texttt{uzn\_latn} (Northern Uzbek).

\item \textbf{East Asia}: \texttt{cmn\_hans} (Mandarin Chinese), \texttt{cmn\_hant} (Mandarin Chinese), \texttt{jpn\_jpan} (Japanese), \texttt{kor\_hang} (Korean), \texttt{yue\_hant} (Yue Chinese, Cantonese).

\item \textbf{Eastern Europe}: \texttt{als\_latn} (Tosk Albanian), \texttt{azj\_latn} (North Azerbaijani), \texttt{bel\_cyrl} (Belarusian), \texttt{bul\_cyrl} (Bulgarian), \texttt{ces\_latn} (Czech), \texttt{ckm\_latn} (Chakavian), \texttt{ell\_grek} (Greek), \texttt{ekk\_latn} (Estonian), \texttt{hrv\_latn} (Croatian), \texttt{hun\_latn} (Hungarian), \texttt{hye\_armn} (Eastern Armenian), \texttt{kat\_geor} (Georgian), \texttt{lit\_latn} (Lithuanian), \texttt{mkd\_cyrl} (Macedonian), \texttt{pol\_latn} (Polish), \texttt{ron\_latn} (Romanian), \texttt{slk\_latn\_sari} (Šariš Slovak), \texttt{slv\_latn\_cerk} (Slovenian, Cerkno), \texttt{slv\_latn\_prle} (Slovenian, Prlekija), \texttt{srp\_cyrl} (Serbian), \texttt{srp\_cyrl\_torl} (Serbian Torlak), \texttt{srp\_latn} (Serbian), \texttt{srp\_latn\_torl} (Serbian Torlak), \texttt{tur\_latn} (Turkish), \texttt{ukr\_cyrl} (Ukrainian), \texttt{bos\_latn} (Bosnian), \texttt{rus\_cyrl} (Russian), \texttt{slk\_latn} (Slovak), \texttt{slv\_latn} (Slovenian).

\item \textbf{Middle East}: \texttt{acq\_arab} (Yemeni Arabic), \texttt{afb\_arab} (Gulf Arabic), \texttt{apc\_arab\_jord} (Jordanian Arabic), \texttt{apc\_arab\_leba} (Lebanese Arabic), \texttt{apc\_arab\_pale} (Palestinian Arabic), \texttt{apc\_arab\_syri} (Syrian Arabic), \texttt{arb\_arab} (Modern Standard Arabic), \texttt{ars\_arab} (Najdi Arabic), \texttt{ckb\_arab} (Central Kurdish), \texttt{heb\_hebr} (Hebrew), \texttt{pes\_arab} (Western Farsi), \texttt{acm\_arab} (Iraqi Arabic).

\item \textbf{North Africa}: \texttt{aeb\_arab} (Tunisian Arabic), \texttt{arq\_arab} (Algerian Arabic), \texttt{ary\_arab} (Moroccan Arabic), \texttt{arz\_arab} (Egyptian Arabic).

\item \textbf{North America}: \texttt{fra\_latn\_cana} (Canadian French), \texttt{spa\_latn\_mexi} (Mexican Spanish).

\item \textbf{Oceania}: \texttt{haw\_latn} (Hawaiian).

\item \textbf{South America}: \texttt{por\_latn\_braz} (Brazilian Portuguese), \texttt{spa\_latn\_peru} (Peruvian Spanish).

\item \textbf{South Asia}: \texttt{asm\_beng} (Assamese), \texttt{bcc\_arab} (Southern Balochi), \texttt{ben\_beng} (Bengali), \texttt{ben\_latn} (Bengali), \texttt{bgc\_deva} (Haryanvi), \texttt{bho\_deva} (Bhojpuri), \texttt{bra\_deva} (Braj), \texttt{bsk\_arab} (Burushaski), \texttt{cls\_deva} (Classical Sanskrit), \texttt{dhd\_deva} (Dhundari), \texttt{guj\_gujr} (Gujarati), \texttt{hin\_deva} (Hindi), \texttt{hin\_latn} (Hindi), \texttt{kan\_knda} (Kannada), \texttt{kan\_latn} (Kannada), \texttt{mag\_deva} (Magahi), \texttt{mal\_mlym} (Malayalam), \texttt{mar\_deva} (Marathi), \texttt{mni\_beng} (Manipuri), \texttt{mni\_mtei} (Meitei Manipuri), \texttt{nag\_latn} (Nagamese), \texttt{npi\_deva} (Nepali), \texttt{ory\_orya} (Odia), \texttt{pan\_guru} (Eastern Panjabi), \texttt{rwr\_deva} (Marwari), \texttt{sin\_latn} (Sinhala), \texttt{sin\_sinh} (Sinhala), \texttt{snd\_arab} (Sindhi), \texttt{snd\_deva} (Sindhi), \texttt{swv\_deva} (Shekhawati), \texttt{tam\_latn} (Tamil), \texttt{tam\_taml} (Tamil), \texttt{tel\_latn} (Telugu), \texttt{tel\_telu} (Telugu), \texttt{urd\_arab} (Urdu), \texttt{urd\_latn} (Urdu).

\item \textbf{Southeast Asia}: \texttt{btx\_latn} (Batak Karo), \texttt{ceb\_latn} (Cebuano), \texttt{ilo\_latn} (Iloko), \texttt{ind\_latn} (Indonesian), \texttt{jav\_latn} (Javanese), \texttt{sun\_latn} (Sundanese), \texttt{tgl\_latn} (Filipino/Tagalog), \texttt{tha\_thai} (Thai), \texttt{vie\_latn} (Vietnamese), \texttt{zsm\_latn} (Malay).

\item \textbf{Sub-Saharan Africa}: \texttt{amh\_ethi} (Amharic), \texttt{bam\_latn} (Bambara), \texttt{dje\_latn} (Zarma), \texttt{ekp\_latn} (Ekpeye), \texttt{hau\_latn} (Hausa), \texttt{ibo\_latn} (Igbo), \texttt{idu\_latn} (Idoma), \texttt{iso\_latn} (Isoko), \texttt{kin\_latn} (Kinyarwanda), \texttt{lin\_latn} (Lingala), \texttt{luo\_latn} (Luo), \texttt{pcm\_latn} (Nigerian Pidgin), \texttt{plt\_latn} (Plateau Malagasy), \texttt{swh\_latn} (Swahili), \texttt{urh\_latn} (Urhobo), \texttt{xho\_latn} (Xhosa), \texttt{yor\_latn} (Yoruba), \texttt{zul\_latn} (Zulu).

\item \textbf{Western Europe (including Northern Europe)}: \texttt{cat\_latn} (Catalan), \texttt{dan\_latn} (Danish), \texttt{deu\_latn} (German), \texttt{eng\_latn} (English), \texttt{eus\_latn} (Basque), \texttt{fao\_latn} (Faroese), \texttt{fin\_latn} (Finnish), \texttt{fra\_latn\_fran} (French), \texttt{glg\_latn} (Galician), \texttt{isl\_latn} (Icelandic), \texttt{ita\_latn} (Italian), \texttt{nld\_latn} (Dutch), \texttt{nno\_latn} (Norwegian Nynorsk), \texttt{nob\_latn} (Norwegian Bokmål), \texttt{por\_latn\_port} (Portuguese), \texttt{spa\_latn\_spai} (Peninsular Spanish), \texttt{swe\_latn} (Swedish).
\end{itemize}

\subsection{Language Gaps Per Model}
\label{app:acc-gap}
For the Global PIQA parallel split, the largest performance gap between languages within each of the closed LLM systems is reported in Table~\ref{tab:model-performance}. Even the best-performing LLM overall (Gemini 3.1 Pro) has an accuracy gap of 68\% between the best- and worst-performing languages (English and Ekpeye).

\begin{table}[ht]
\centering
\caption{Largest performance gap between languages for each closed LLM system.}
\label{tab:model-performance}
\begin{tabular}{lllrllrr}
\toprule
Model & Best language & Best & Worst language & Worst & Acc. \\
& & acc. & & acc. & gap \\
\midrule
GPT-5.4              & \texttt{spa\_latn\_peru} & 99\% & \texttt{ekp\_latn}      & 28\% & 71\%  \\
GPT-5.4 Mini         & \texttt{sin\_sinh}     & 95\% & \texttt{ekp\_latn}      & 18\% & 77\%  \\
GPT-5.4 Nano         & \texttt{srp\_cyrl}     & 92\% & \texttt{bam\_latn}     & 20\% & 72\% \\
Claude Sonnet 4.6    & \texttt{fin\_latn}   & 99\% & \texttt{ekp\_latn}   & 20\% & 79\%  \\
Gemini 3.1 Pro       & \texttt{eng\_latn} & 100\% & \texttt{ekp\_latn}   & 32\% & 68\% \\
Gemini 3 Flash & \texttt{srp\_cyrl} & 99\% & \texttt{ekp\_latn}   & 33\% & 66\% \\
Gemini 3.1 Flash Lite & \texttt{nob\_latn}     & 96\% & \texttt{ekp\_latn}      & 24\% & 72\% \\
\bottomrule \\
\end{tabular}
\end{table}

\clearpage

\clearpage

\section{Non-Parallel Split: Individual Dataset Descriptions}
\label{app:descriptions}

Here, we provide brief descriptions of the methods that individual groups used to construct their contributions to the non-parallel split of Global PIQA (\S\ref{sec:non-parallel}).
Longer dataset description papers that authors consented to release are at \url{https://github.com/mrlbenchmarks/global-piqa}.
Authors were recruited and organized as described in \S\ref{app:organizing}, and all contributors were offered authorship.
The vast majority have chosen to be authors on this paper. This project would not be possible without the efforts of all authors.

We note that we intentionally do not list authors with their groups and languages. This is to preserve privacy, as some authors would prefer not to be contacted by unaffiliated projects that require expertise in their language.
Skipped group numbers indicate groups whose datasets were not accepted to Global PIQA, often because they did not complete their datasets or did not respond to feedback to correct their datasets (\S\ref{app:organizing}).
\medskip


\hangindent=0.5cm
\hangafter=1
\textbf{Group 0000: Hindi} (\texttt{hin\_deva}: 94 examples)

\setlength{\parskip}{0pt}
\hangindent=0cm
Manually written in English by a native Hindi speaker, machine-translated into Hindi using Google Translate, then checked, corrected, and refined by the dataset author. Approximately 25\% of examples are designed to be culturally-grounded, with references to specific Indian culinary items, musical instruments, common fauna, and social traditions, such as customs within a wedding ceremony.
\setlength{\parskip}{\baselineskip}

\hangindent=0.5cm
\hangafter=1
\textbf{Group 0001: Telugu} (\texttt{tel\_telu}: 131 examples)

\setlength{\parskip}{0pt}
\hangindent=0cm
Manually written by a native Telugu speaker, with examples crafted to reflect realistic scenarios encountered in Telugu households, agriculture,
cooking, transportation, and daily problem-solving. Each question was double-checked, and edge cases and ambiguous situations were discarded to ensure high quality.
\setlength{\parskip}{\baselineskip}

\hangindent=0.5cm
\hangafter=1
\textbf{Group 0002: French (Canadian)} (\texttt{fra\_latn\_cana}: 155 examples)

\setlength{\parskip}{0pt}
\hangindent=0cm
Topic ideas were brainstormed using LLMs, but examples were all written manually. All examples were checked or written by a native speaker.
\setlength{\parskip}{\baselineskip}

\hangindent=0.5cm
\hangafter=1
\textbf{Group 0003: Yoruba} (\texttt{yor\_latn}: 120 examples)

\setlength{\parskip}{0pt}
\hangindent=0cm
Examples from English PIQA were translated and culturally adapted to Yoruba by a native Yoruba speaker. Care was taken to preserve Yoruba idiomatic forms, and for culturally unique contexts, questions were created directly in Yoruba rather than translated. Culturally-specific domains include cooking, clothing, farming, weather, transportation, religion, household practices, and festivals.
\setlength{\parskip}{\baselineskip}

\hangindent=0.5cm
\hangafter=1
\textbf{Group 0004: French} (\texttt{fra\_latn\_fran}: 100 examples)

\setlength{\parskip}{0pt}
\hangindent=0cm
Manually written by a native French speaker, with examples crafted by observing daily life and
social interactions, and by browsing French websites for topics such as furniture, home goods, sports, and news. Many examples were designed to be specific to French culture, e.g. including French food and social norms, or how to take the metro in Paris.
\setlength{\parskip}{\baselineskip}

\hangindent=0.5cm
\hangafter=1
\textbf{Group 0005: Finnish} (\texttt{fin\_latn}: 100 examples)

\setlength{\parskip}{0pt}
\hangindent=0cm
Manually written by a native Finnish speaker, with many examples covering Finnish culture and everyday life. Topics include traditional foods, household chores, log cabin terms, saunas, winter activities, reindeer-related terms, and Finnish sports and traditions.
\setlength{\parskip}{\baselineskip}

\hangindent=0.5cm
\hangafter=1
\textbf{Group 0006: Hungarian, Romanian} (\texttt{hun\_latn}: 100 examples, \texttt{ron\_latn}: 100 examples)

\setlength{\parskip}{0pt}
\hangindent=0cm
Examples were written in English, translated into Hungarian and Romanian (by native speakers of those languages), and reviewed by another translator. All translators and editors were offered authorship.
\setlength{\parskip}{\baselineskip}

\hangindent=0.5cm
\hangafter=1
\textbf{Group 0007: Ukrainian} (\texttt{ukr\_cyrl}: 100 examples)

\setlength{\parskip}{0pt}
\hangindent=0cm
Manually written by a native Ukrainian speaker, and checked by another native speaker, both from Western Ukraine. Topics were inspired by Ukrainian websites and blogs, as well as personal knowledge, covering Ukrainian cuisine, traditions, superstitions, and local Ukrainian festivities.
\setlength{\parskip}{\baselineskip}

\pagebreak

\hangindent=0.5cm
\hangafter=1
\textbf{Group 0008: Mandarin} (\texttt{cmn\_hans}: 100 examples, \texttt{cmn\_hant}: 98 examples)

\setlength{\parskip}{0pt}
\hangindent=0cm
Manually written by a native Mandarin speaker and verified by another native speaker. Examples were balanced across culturally-specific food, clothing and materials, musical instruments, and other objects. Examples were written using Chinese simplified characters, but also translated into traditional characters using Google Translate with human verification.
\setlength{\parskip}{\baselineskip}

\hangindent=0.5cm
\hangafter=1
\textbf{Group 0009: Hebrew} (\texttt{heb\_hebr}: 67 examples)

\setlength{\parskip}{0pt}
\hangindent=0cm
Manually written by a native Hebrew speaker, with examples covering specific Hebrew linguistic constructions, along with Israeli cultural knowledge, such as places, food, climate, and Jewish religion and culture. By design, some items may resist direct translation into other languages, and in some cases, translation may alter the validity of the designated correct answer.
\setlength{\parskip}{\baselineskip}

\hangindent=0.5cm
\hangafter=1
\textbf{Group 0010: Indonesian} (\texttt{ind\_latn}: 204 examples)

\setlength{\parskip}{0pt}
\hangindent=0cm
Examples were generated with the assistance of ChatGPT (GPT-5) using carefully guided prompts to produce PIQA-style examples. All examples were manually reviewed, corrected, and finalized by a native speaker of Indonesian to ensure quality, correctness, and cultural relevance. Because the original LLM-generated examples were often fairly generic, at least 50 examples were manually edited to reflect uniquely Indonesian contexts (e.g. local foods, household practices, and traditional objects). The dataset was written in Standard Indonesian (Bahasa Indonesia).
\setlength{\parskip}{\baselineskip}

\hangindent=0.5cm
\hangafter=1
\textbf{Group 0011: Italian} (\texttt{ita\_latn}: 100 examples)

\setlength{\parskip}{0pt}
\hangindent=0cm
Manually written by a native Italian speaker. ChatGPT was occasionally used to correct typos or to find appropriate words that did not immediately come to mind, but never to generate examples themselves. All final versions of examples were human verified. To include examples reflecting Italian culture, some examples were motivated by online recipes and websites in Italian.
\setlength{\parskip}{\baselineskip}

\hangindent=0.5cm
\hangafter=1
\textbf{Group 0012: Hausa} (\texttt{hau\_latn}: 100 examples)

\setlength{\parskip}{0pt}
\hangindent=0cm
Manually written by a native Hausa speaker, using culturally-relevant themes to motivate example creation. Themes included traveling, food, school, exams, driving, and health.
\setlength{\parskip}{\baselineskip}

\hangindent=0.5cm
\hangafter=1
\textbf{Group 0013: Portuguese (Brazilian)} (\texttt{por\_latn\_braz}: 100 examples)

\setlength{\parskip}{0pt}
\hangindent=0cm
Manually written by a native Brazilian Portuguese speaker, covering food, traditions, regional objects, daily activities, and environmental contexts that are common to Brazil, particularly southern Brazil.
\setlength{\parskip}{\baselineskip}

\hangindent=0.5cm
\hangafter=1
\textbf{Group 0014: Dutch} (\texttt{nld\_latn}: 120 examples)

\setlength{\parskip}{0pt}
\hangindent=0cm
Manually written by a native Dutch speaker, using specific culturally-relevant topics to motivate example creation. Topics include bicycle maintenance techniques, preparation of traditional Dutch foods, managing Dutch rainfall, and navigating Amsterdam's narrow spaces. All examples were verified by another native speaker.
\setlength{\parskip}{\baselineskip}

\hangindent=0.5cm
\hangafter=1
\textbf{Group 0015: Tagalog / Filipino} (\texttt{tgl\_latn}: 103 examples)

\setlength{\parskip}{0pt}
\hangindent=0cm
Manually written by a native Tagalog speaker. A separate Filipino dataset was not included, as many native speakers of Tagalog do not draw a strong distinction between the two. Examples in this dataset were written to be culturally-specific to the Philippines, covering three main topics: (1) cooking and baking, (2) crafts and construction of cultural objects, and (3) art, dances, and literature. The author cross-checked information using websites such as Philippine Wikipedia, Philippine government blogs on culture, and informal verification from fellow native speakers living in the Philippines.
\setlength{\parskip}{\baselineskip}

\hangindent=0.5cm
\hangafter=1
\textbf{Group 0016: Vietnamese} (\texttt{vie\_latn}: 100 examples)

\setlength{\parskip}{0pt}
\hangindent=0cm
Manually written by a native Vietnamese speaker, and examples contain Vietnamese cultural contexts such as everyday objects, weather, clothing, routines, safety, school, simple social norms, and holidays.
\setlength{\parskip}{\baselineskip}

\hangindent=0.5cm
\hangafter=1
\textbf{Group 0017: Russian, Iraqi Arabic (Gelet)} (\texttt{rus\_cyrl}: 100 examples, \texttt{acm\_arab}: 121 examples)

\setlength{\parskip}{0pt}
\hangindent=0cm
Manually written by native Russian and Iraqi Arabic (Gelet) speakers, covering everyday topics such as weather, transportation, home safety, work, hobbies, nature, sports, school, and technology. For a more culturally-specific subset, approximately 20 examples for Iraqi Arabic were translated from the Modern Standard Arabic dataset from Group 0065; a native speaker of Iraqi Arabic selected examples that were culturally relevant to their region.
\setlength{\parskip}{\baselineskip}

\hangindent=0.5cm
\hangafter=1
\textbf{Group 0018: Korean} (\texttt{kor\_hang}: 100 examples)

\setlength{\parskip}{0pt}
\hangindent=0cm
Manually written and verified by three native Korean speakers. Examples were written to cover popular Korean games, food, and mandatory military service.
\setlength{\parskip}{\baselineskip}

\hangindent=0.5cm
\hangafter=1
\textbf{Group 0019: Mandarin} (\texttt{cmn\_hans}: 93 examples)

\setlength{\parskip}{0pt}
\hangindent=0cm
Manually written by a native Mandarin speaker, covering traditional Chinese culture, food, objects, everyday life, customs, and computer use. Some examples were motivated by reading guidebooks on transportation, cooking, or safety operations. Some examples were also designed to cover recently-developed technologies from within the past five to ten years.
\setlength{\parskip}{\baselineskip}

\hangindent=0.5cm
\hangafter=1
\textbf{Group 0020: Kannada} (\texttt{kan\_knda}: 99 examples)

\setlength{\parskip}{0pt}
\hangindent=0cm
Manually written by a native Kannada speaker, and verified by another native speaker. Examples reflect cultural aspects of Karnataka (an Indian state where Kannada is widely spoken), as well as everyday scenarios.
\setlength{\parskip}{\baselineskip}

\hangindent=0.5cm
\hangafter=1
\textbf{Group 0021: Yoruba} (\texttt{yor\_latn}: 99 examples)

\setlength{\parskip}{0pt}
\hangindent=0cm
Manually written by a native Yoruba speaker, and verified by another native speaker. Examples are written to be relevant to the Yoruba land, including festivals, traditions, foods, and clothing.
\setlength{\parskip}{\baselineskip}

\hangindent=0.5cm
\hangafter=1
\textbf{Group 0022: Slovenian, Croatian, Serbian, Macedonian, Slovenian Cerkno, Chakavian} (\texttt{slv\_latn}: 100 examples, \texttt{hrv\_latn}: 100 examples, \texttt{srp\_latn}: 149 examples, \texttt{srp\_cyrl}: 150 examples, \texttt{mkd\_cyrl}: 100 examples, \texttt{slv\_latn\_cerk}: 100 examples, \texttt{ckm\_latn}: 100 examples)

\setlength{\parskip}{0pt}
\hangindent=0cm
Manually written by native speakers of Slovenian, Croatian, Serbian, Macedonian, and two dialects: Slovenian Cerkno and Croatian Chakavian. Authors attempted to include culturally-relevant examples for their language(s). Examples were motivated by everyday objects, life hacks, recipes, and/or assembly manuals in each language. For each dataset, another co-author with significant understanding of the language or dialect solved the task without access to labels. Human accuracies were 97\%, 100\%, 97\%, and 92\%, excluding the two low-resource dialects. Labels were adjusted based on disagreements from this cross-check.
\setlength{\parskip}{\baselineskip}

\hangindent=0.5cm
\hangafter=1
\textbf{Group 0023: Tagalog} (\texttt{tgl\_latn}: 100 examples)

\setlength{\parskip}{0pt}
\hangindent=0cm
Manually written by a native Tagalog speaker, using both common spoken Tagalog (Northern and Manila dialects) and the Filipino dialect. Writing style varies between street-spoken Tagalog and formal Tagalog, and topics focus on daily life in the agricultural town of Talavera, Nueva Ecija (e.g. fishing and cooking). Some examples were inspired by Instructables posts, adapted to be culturally-relevant.
\setlength{\parskip}{\baselineskip}

\hangindent=0.5cm
\hangafter=1
\textbf{Group 0024: French} (\texttt{fra\_latn\_fran}: 115 examples)

\setlength{\parskip}{0pt}
\hangindent=0cm
Manually written and reviewed by native French speakers, using French as spoken in mainland France. Examples were written by observing everyday actions, with distracting information added to some prompts to make the examples more challenging.
\setlength{\parskip}{\baselineskip}

\hangindent=0.5cm
\hangafter=1
\textbf{Group 0025: Polish} (\texttt{pol\_latn}: 130 examples)

\setlength{\parskip}{0pt}
\hangindent=0cm
Manually written and reviewed by native Polish speakers. Authors drew upon their knowledge of Polish history, culture, customs, and everyday habits.
\setlength{\parskip}{\baselineskip}

\hangindent=0.5cm
\hangafter=1
\textbf{Group 0026: Norwegian Bokmål, Norwegian Nynorsk} (\texttt{nob\_latn}: 117 examples, \texttt{nno\_latn}: 117 examples)

\setlength{\parskip}{0pt}
\hangindent=0cm
Manually written in Norwegian Bokmål by native Norwegian speakers, including examples covering local foods, activities, traditions, folklore, and indigenous culture. Text embedding similarity search and then manual verification were used to ensure that examples were not direct translations of English PIQA. Examples were translated into Norwegian Nynorsk using the Nynorsk dictionary from LEXIN OsloMet, and checked by a Norwegian speaker who used Norwegian Nynorsk in school.
\setlength{\parskip}{\baselineskip}

\hangindent=0.5cm
\hangafter=1
\textbf{Group 0027: Malay} (\texttt{zsm\_latn}: 100 examples)

\setlength{\parskip}{0pt}
\hangindent=0cm
Manually written by a native Malay speaker, using Standard Malay (Bahasa Melayu). Examples were designed to cover local commonsense, social norms, food and drink, religious life, and everyday routines. Examples were written with natural Malay phrasing and colloquial register where appropriate.
\setlength{\parskip}{\baselineskip}

\hangindent=0.5cm
\hangafter=1
\textbf{Group 0028: Faroese} (\texttt{fao\_latn}: 100 examples)

\setlength{\parskip}{0pt}
\hangindent=0cm
Manually written and reviewed by native Faroese speakers. Approximately 35 examples were written to be specific to the Faroe Islands, focusing on Faroese food preparation and preservation techniques, weather patterns, traditional clothing, wool and knitting, and geography.
\setlength{\parskip}{\baselineskip}

\hangindent=0.5cm
\hangafter=1
\textbf{Group 0029: Urdu} (\texttt{urd\_arab}: 115 examples)

\setlength{\parskip}{0pt}
\hangindent=0cm
This dataset was written by native Urdu speakers, using Gemini 2.5 Flash and Claude Sonnet 4 for example clarification and refinement. Local websites such as UrduPoint were used to motivate examples, and examples were designed to reflect everyday life in Pakistan, including Pakistani food preparation, household practices, social customs, and traditional crafts. The dataset is written in Standard Pakistani Urdu, with every example checked by at least two native speakers.
\setlength{\parskip}{\baselineskip}

\hangindent=0.5cm
\hangafter=1
\textbf{Group 0030: Uzbek} (\texttt{uzn\_latn}: 101 examples)

\setlength{\parskip}{0pt}
\hangindent=0cm
Manually written by a native Northern Uzbek speaker, drawing from real-life experiences and commonly-used expressions in Uzbek. Colloquial phrases are used where appropriate. The dataset is written using Latin script, although Cyrillic script is also widely used in Uzbekistan.
\setlength{\parskip}{\baselineskip}

\hangindent=0.5cm
\hangafter=1
\textbf{Group 0031: Icelandic} (\texttt{isl\_latn}: 100 examples)

\setlength{\parskip}{0pt}
\hangindent=0cm
Manually written by native Icelandic speakers, covering culturally-specific topics such as food and cooking, holidays and traditions, civics and culture, folklore, geography, history, and agriculture. Some examples were inspired by browsing the Icelandic science web (\url{https://www.visindavefur.is/}).
\setlength{\parskip}{\baselineskip}

\hangindent=0.5cm
\hangafter=1
\textbf{Group 0032: Bengali} (\texttt{ben\_beng}: 102 examples)

\setlength{\parskip}{0pt}
\hangindent=0cm
Manually written by a native Bengali speaker, with culturally grounded examples reflecting daily life in Bangladesh and West Bengal, India. Examples were written to reflect everyday topics such as household chores, seasonal weather, agriculture, cooking, storage, and material interactions.
\setlength{\parskip}{\baselineskip}

\hangindent=0.5cm
\hangafter=1
\textbf{Group 0033: Tunisian Arabic} (\texttt{aeb\_arab}: 100 examples)

\setlength{\parskip}{0pt}
\hangindent=0cm
This dataset was created using a mix of manual writing and LLM generation, with all examples verified by two native speakers of Tunisian Arabic. The examples are written to reflect everday life in Tunisia, including cooking practices, traditional music and instruments, household activities, local customs, and everyday objects. Because Tunisian Arabic is primarily a spoken dialect with no standardized orthography, some linguistic variation may appear across examples.
\setlength{\parskip}{\baselineskip}

\hangindent=0.5cm
\hangafter=1
\textbf{Group 0034: Marathi} (\texttt{mar\_deva}: 119 examples)

\setlength{\parskip}{0pt}
\hangindent=0cm
Manually written by native Marathi speakers, using Marathi as spoken in Pune City, Maharashtra, India (i.e. Puneri dialect). Examples were written to cover culturally-specific everyday topics such as education and exams, cooking and household activities, sports and games, and shopping and technology.
\setlength{\parskip}{\baselineskip}

\hangindent=0.5cm
\hangafter=1
\textbf{Group 0035: Japanese} (\texttt{jpn\_jpan}: 101 examples)

\setlength{\parskip}{0pt}
\hangindent=0cm
One subset of this dataset was created by native Japanese speakers using ChatGPT to translate English PIQA examples and to replace lexical elements with Japanese-specific counterparts. Another subset prompted ChatGPT to generate novel Japanese examples that required knowledge of Japanese cultural norms and conventions. Of the translated subset, 35 out of 145 passed quality checks by the native speakers, and of the novel generations, 66 out of 300 generated examples passed quality checks. All examples were verified by two native Japanese speakers.
\setlength{\parskip}{\baselineskip}

\hangindent=0.5cm
\hangafter=1
\textbf{Group 0036: Italian} (\texttt{ita\_latn}: 120 examples)

\setlength{\parskip}{0pt}
\hangindent=0cm
Manually written by native Italian speakers, covering household, cuisine, and entertainment domains, focusing on everyday scenarios reflecting local Italian practices. All examples were validated for fluency, correctness, and adherence to the task description by another native speaker.
\setlength{\parskip}{\baselineskip}

\pagebreak

\hangindent=0.5cm
\hangafter=1
\textbf{Group 0037: Indonesian} (\texttt{ind\_latn}: 120 examples)

\setlength{\parskip}{0pt}
\hangindent=0cm
Manually written and verified by native Indonesian speakers, with examples motivated by the authors' general knowledge, past experiences, and daily life activities. By design, some prompts incorporated culturally specific Indonesian elements, such as food and traditional musical instruments. All examples were checked by at least two native speakers.
\setlength{\parskip}{\baselineskip}

\hangindent=0.5cm
\hangafter=1
\textbf{Group 0038: Vietnamese} (\texttt{vie\_latn}: 120 examples)

\setlength{\parskip}{0pt}
\hangindent=0cm
Manually written and verified by native Vietnamese speakers, highlighting both Kinh Vietnamese culture and minority ethnic culture (e.g. from the 50+ ethnic minority groups in present-day Vietnam). Examples cover culturally-specific knowledge such as cooking and farming methods, folklore, traditions, well-known cultural events, and minority ethnic culture. All examples were checked by at least two native speakers.
\setlength{\parskip}{\baselineskip}

\hangindent=0.5cm
\hangafter=1
\textbf{Group 0039: Korean} (\texttt{kor\_hang}: 441 examples)

\setlength{\parskip}{0pt}
\hangindent=0cm
Korean questions were collected from Naver Knowledge iN1, a popular Korean Q\&A platform, covering diverse everyday scenarios where Korean users seek practical advice on physical tasks and problem-solving. Qwen3-4B, Qwen3-32B, and HCX-14B were used to identify PIQA-style questions, keeping only questions where all three models unanimously agreed that the question fit the task description (less than 1\% of the originally collected examples). Then, GPT-4o was used to refine questions and generate incorrect solutions. Two native Korean speakers independently validated each question, improving question clarity, calibrating difficulty levels, and verifying cultural appropriateness. KoSentenceBERT was used to removed near-duplicate questions. Of the final dataset, approximately 85 questions contain elements specific to Korean culture such as traditional foods and cooking methods, clothing care, housing systems, specialized appliances, and cultural practices.
\setlength{\parskip}{\baselineskip}

\hangindent=0.5cm
\hangafter=1
\textbf{Group 0040: Urdu} (\texttt{urd\_arab}: 99 examples, \texttt{urd\_latn}: 100 examples)

\setlength{\parskip}{0pt}
\hangindent=0cm
Manually written by a native Urdu speaker using Latin script, in line with the way many Pakistanis communicate on social media platforms. Examples were transliterated into Urdu script using Gemini 2.5 Flash and then manually verified.
\setlength{\parskip}{\baselineskip}

\hangindent=0.5cm
\hangafter=1
\textbf{Group 0041: Hebrew} (\texttt{heb\_hebr}: 209 examples)

\setlength{\parskip}{0pt}
\hangindent=0cm
Manually written by native Hebrew speakers, with each example verified by another native speaker. Approximately 55 examples cover everyday Israeli life or Jewish religious practices, including recipes, household cleaning techniques, cultural traditions, and religious customs. For some examples, motivation for topics came from Wikipedia articles or from lists of everyday objects obtained by prompting LLMs.
\setlength{\parskip}{\baselineskip}

\hangindent=0.5cm
\hangafter=1
\textbf{Group 0042: Catalan, Peninsular Spanish} (\texttt{cat\_latn}: 100 examples, \texttt{spa\_latn\_spai}: 90 examples)

\setlength{\parskip}{0pt}
\hangindent=0cm
Manually written in Catalan by a native Catalan and Spanish speaker, covering everyday topics such as clothing, festivity, folklore, food, literature, music, and sports. Many examples include concepts and situations that are specific to Catalan-speaking communities, and some examples do not translate well into other languages. The Catalan dataset underwent human evaluation by three native speakers, who achieved accuracies of 94\%, 95\%, and 98\% respectively; examples were then adjusted based on this cross-checking. The dataset was translated into Spanish using Google Translate, then post-edited by the same native speaker, keeping examples for Spanish only if they remained valid after translation.
\setlength{\parskip}{\baselineskip}

\hangindent=0.5cm
\hangafter=1
\textbf{Group 0043: Polish} (\texttt{pol\_latn}: 103 examples)

\setlength{\parskip}{0pt}
\hangindent=0cm
Manually written by a native Polish speaker based on physics topics, including fundamental laws of physics, material properties, and principles governing interactions between materials. Online materials describing at-home basic experiments were used to motivate some examples, and several Polish-specific words (e.g. cooking and food items) were used.
\setlength{\parskip}{\baselineskip}

\hangindent=0.5cm
\hangafter=1
\textbf{Group 0045: Belarusian} (\texttt{bel\_cyrl}: 183 examples)

\setlength{\parskip}{0pt}
\hangindent=0cm
Manually written in conversational Belarusian by native Belarusian speakers, inspired by household situations, local customs, and guides on Belarusian life. LLMs were then used for paraphrasing, lengthening examples, and normalizing style, and then all examples were checked again by two native speakers.
\setlength{\parskip}{\baselineskip}

\hangindent=0.5cm
\hangafter=1
\textbf{Group 0046: Swedish} (\texttt{swe\_latn}: 98 examples)

\setlength{\parskip}{0pt}
\hangindent=0cm
Manually written by a native Swedish speaker, and checked by another native speaker. Roughly half of examples include Swedish slang, traditions, or foods, or hard-to-translate Swedish words.
\setlength{\parskip}{\baselineskip}

\hangindent=0.5cm
\hangafter=1
\textbf{Group 0047: Bulgarian} (\texttt{bul\_cyrl}: 122 examples)

\setlength{\parskip}{0pt}
\hangindent=0cm
Manually written by a native Bulgarian speaker, and checked by another native speaker. Examples are designed to test specific types of physical commonsense reasoning, with distractors (incorrect solutions) that are still semantically related to the prompts. Examples are interwoven with Bulgarian cultural elements and require knowledge of Bulgarian morphological cues (e.g. word inflections).
\setlength{\parskip}{\baselineskip}

\hangindent=0.5cm
\hangafter=1
\textbf{Group 0048: Mandarin, Cantonese} (\texttt{cmn\_hans}: 407 examples, \texttt{yue\_hant}: 223 examples)

\setlength{\parskip}{0pt}
\hangindent=0cm
Manually written and reviewed by native Mandarin and Cantonese speakers, based on online encyclopedias and guidebooks in Mandarin and Cantonese. Example domains include activities (e.g. sports), food, geography, and art.
\setlength{\parskip}{\baselineskip}

\hangindent=0.5cm
\hangafter=1
\textbf{Group 0049: Yoruba, Igbo, Naija (Nigerian Pidgin), Hausa, Isoko, Urhobo, Idoma} (\texttt{yor\_latn}: 974 examples, \texttt{ibo\_latn}: 432 examples, \texttt{pcm\_latn}: 247 examples, \texttt{hau\_latn}: 213 examples, \texttt{iso\_latn}: 107 examples, \texttt{urh\_latn}: 119 examples, \texttt{idu\_latn}: 101 examples)

\setlength{\parskip}{0pt}
\hangindent=0cm
Manually written by native speakers of Yoruba, Hausa, Igbo, Idoma, Urhobo, Naija (Nigerian Pidgin English), and Isoko, as part of a community effort by the Linguistics Island community of linguists. Examples cover specific linguistic structures, and topics include food, culture, education, and technology.
\setlength{\parskip}{\baselineskip}

\hangindent=0.5cm
\hangafter=1
\textbf{Group 0050: Bengali, Mandarin, Greek, Korean, Turkish} (\texttt{ben\_beng}: 50 examples, \texttt{cmn\_hans}: 48 examples, \texttt{cmn\_hant}: 17 examples, \texttt{ell\_grek}: 50 examples, \texttt{kor\_hang}: 50 examples, \texttt{tur\_latn}: 49 examples)

\setlength{\parskip}{0pt}
\hangindent=0cm
Manually written by native speakers of Bengali, Mandarin (Taiwanese using traditional characters, mainland using simplified characters), Greek, Korean, and Turkish. All examples were checked by another native speaker of the language. Many examples were written by first thinking of a culturally-specific item, then brainstorming physical properties of that item that could be incorporated into a PIQA-style example.
\setlength{\parskip}{\baselineskip}

\hangindent=0.5cm
\hangafter=1
\textbf{Group 0051: Uyghur} (\texttt{uig\_arab}: 132 examples)

\setlength{\parskip}{0pt}
\hangindent=0cm
Manually written by a native speaker of Uyghur, with each example proofread by five native speakers and using a Uyghur spell-checker. Examples were inspired by Uyghur literary materials, including cultural and traditional texts, proverbs and sayings, folklore collections, and instructional manuals.
\setlength{\parskip}{\baselineskip}

\hangindent=0.5cm
\hangafter=1
\textbf{Group 0053: Bengali} (\texttt{ben\_latn}: 100 examples)

\setlength{\parskip}{0pt}
\hangindent=0cm
Manually written by a native Bengali speaker using ``Banglish'', or Bengali language written in Latin script, often used by Bengali speakers in online settings and informal communication. Examples cover culturally-specific topics such as Bengali religious festivals and practices, traditional foods and cooking, household objects and tools, traditional games and activities, seasonal practices and nature, and folk traditions and customs. ChatGPT was used to brainstorm additional cultural topics, but not to generate examples.
\setlength{\parskip}{\baselineskip}

\hangindent=0.5cm
\hangafter=1
\textbf{Group 0055: Estonian} (\texttt{ekk\_latn}: 100 examples)

\setlength{\parskip}{0pt}
\hangindent=0cm
Manually written by native Estonian speakers, covering culturally relevant elements such as traditional Estonian foods, local materials, and region-specific practices. Inspiration for some examples was drawn from the ``Maybe I’m Lucky'' feature of Sõnaveeb, the language portal maintained by the Institute of the Estonian Language, generating randomly-selected Estonian words. Examples were each tested on six randomly-selected LLMs, and examples that all models got correct were dropped or edited. For human evaluation, another native speaker achieved an accuracy of 95\%; examples were then adjusted based on this cross-checking.
\setlength{\parskip}{\baselineskip}

\pagebreak

\hangindent=0.5cm
\hangafter=1
\textbf{Group 0056: Dutch} (\texttt{nld\_latn}: 100 examples)

\setlength{\parskip}{0pt}
\hangindent=0cm
This dataset was constructed by a native Dutch speaker using a hybrid LLM and manual approach, then reviewed by another native speaker. It includes culturally-relevant topics such as chocolate sprinkles on bread, ice skating, dikes, local sports, and specific dishes. LLMs, including GPT-5, Gemini 2.5 Pro, and Claude Sonnet 4, were used in drafting samples, suggesting topics, and proofreading, but overall, their performance was found to be severely lacking in understanding the task and generating suitable examples without significant refinement.
\setlength{\parskip}{\baselineskip}

\hangindent=0.5cm
\hangafter=1
\textbf{Group 0057: Estonian, Persian (Farsi), Swedish} (\texttt{ekk\_latn}: 105 examples, \texttt{pes\_arab}: 123 examples, \texttt{swe\_latn}: 100 examples)

\setlength{\parskip}{0pt}
\hangindent=0cm
The Estonian part of this dataset was manually written by a native Estonian speaker, and reviewed by another native speaker. Topics include Estonian food, companies, places, cultural events and holidays, and typical activities and phenomena during different seasons of the year. The Farsi part of this dataset was manually written and reviewed by native Farsi speakers, covering six thematic categories: cooking and food, housekeeping and cleaning, daily life and social customs, driving and travel, health and safety, and life hacks and tools. The dataset emphasizes cultural and contextual knowledge, and inspiration was drawn from online articles in Farsi. The Swedish part of this dataset was manually written by a native Swedish speaker, and reviewed by another native speaker, drawing inspiration from online sources that cover everyday physical activities (e.g. sports, gardening, household life, traditional festivities, and traffic-related scenarios).
\setlength{\parskip}{\baselineskip}

\hangindent=0.5cm
\hangafter=1
\textbf{Group 0058: Hindi, Sindhi, Punjabi, Manipuri, Bengali, Gujarati, Marathi, Nepali, Bhojpuri, Marwari, Dhundhari, Nagamese} (\texttt{hin\_deva}: 117 examples, \texttt{snd\_deva}: 116 examples, \texttt{pan\_guru}: 117 examples, \texttt{mni\_beng}: 117 examples, \texttt{bho\_deva}: 117 examples, \texttt{guj\_gujr}: 94 examples, \texttt{mar\_deva}: 117 examples, \texttt{npi\_deva}: 93 examples, \texttt{ben\_beng}: 117 examples, \texttt{rwr\_deva}: 117 examples, \texttt{dhd\_deva}: 116 examples, \texttt{nag\_latn}: 117 examples)

\setlength{\parskip}{0pt}
\hangindent=0cm
Examples in this dataset were primarily adapted from reasoning textbooks in English and Hindi that are widely used for preparation for competitive exams. Examples were written to reflect India-specific cultural contexts. Each example was manually or semi-automatically (i.e. machine-translated with human verification) translated into the 12 target languages, with careful preservation of meaning, cultural familiarity, and syntactic naturalness. All examples were independently labeled by two native speakers to ensure validity.
\setlength{\parskip}{\baselineskip}

\hangindent=0.5cm
\hangafter=1
\textbf{Group 0059: Lingala} (\texttt{lin\_latn}: 102 examples)

\setlength{\parskip}{0pt}
\hangindent=0cm
Manually written by a native Lingala speaker, covering culturally-specific everyday contexts and daily life.
\setlength{\parskip}{\baselineskip}

\hangindent=0.5cm
\hangafter=1
\textbf{Group 0060: Greek} (\texttt{ell\_grek}: 206 examples)

\setlength{\parskip}{0pt}
\hangindent=0cm
Manually written and reviewed by native Greek speakers. Some prompts are adapted from a variety of online material, including government and non-governmental organization (NGO) publications, academic theses, course presentations, commercial product brochures, and Wikipedia. Approximately 40\% of the final examples are annotated by the authors as culturally specific.
\setlength{\parskip}{\baselineskip}

\hangindent=0.5cm
\hangafter=1
\textbf{Group 0061: Sindhi} (\texttt{snd\_arab}: 139 examples)

\setlength{\parskip}{0pt}
\hangindent=0cm
Manually written by a native Sindhi speaker, using Standard Sindhi (Vicholi Sindhi) in the Perso-Arabic script. Examples are culturally grounded in folklore, history, literature, foods, festivals, traditions, and everyday life in Sindh, Pakistan.
\setlength{\parskip}{\baselineskip}

\hangindent=0.5cm
\hangafter=1
\textbf{Group 0062: Swahili, Dhuluo, Lingala} (\texttt{swh\_latn}: 220 examples, \texttt{luo\_latn}: 101 examples, \texttt{lin\_latn}: 188 examples)

\setlength{\parskip}{0pt}
\hangindent=0cm
The dataset was manually written and reviewed by native speakers of Swahili, Dholuo, and Lingala, covering topics such as food, agriculture, transportation, and household practices. The Swahili examples are split between Kenyan and Tanzanian Swahili; these two varieties are structurally similar, but Tanzanian contributions emphasize domestic and rural practices, while Kenyan contributions highlight more urban contexts. The Lingala examples focus on rural life in Central Africa, including cassava preparation, termite cooking, fishing, river transport, market trading, and home construction.
\setlength{\parskip}{\baselineskip}

\pagebreak

\hangindent=0.5cm
\hangafter=1
\textbf{Group 0063: Albanian} (\texttt{als\_latn}: 106 examples)

\setlength{\parskip}{0pt}
\hangindent=0cm
Manually written by a linguist specializing in Albanian and a native speaker of Albanian. Topics cover domains such as cooking, cleaning, object construction, Albanian traditional activities (e.g. music, dances, weddings), cultural practices, and agricultural tasks. The authors note that both dataset creators primarily reside outside the main Albanian-speaking continuum, potentially affecting the representativeness of the selected topics.
\setlength{\parskip}{\baselineskip}

\hangindent=0.5cm
\hangafter=1
\textbf{Group 0064: Indonesian} (\texttt{ind\_latn}: 228 examples)

\setlength{\parskip}{0pt}
\hangindent=0cm
This dataset was created by native Indonesian speakers using GPT-4o with careful prompting to generate culturally-specific examples. Topics include agriculture, art, daily activities, family relationships, fisheries and trade, food, religious holidays, traditional games, and wedding traditions. Examples were filtered for fluency, correctness, and adherence to the task format, and SentenceBERT was used to filter out near-duplicate examples. All examples were reviewed and edited by two native Indonesian speakers, using Standard Indonesian (Bahasa Indonesia). The filtering stages (including filtering for ambiguous solutions) resulted in removing 85.4\% of the original LLM-generated examples.
\setlength{\parskip}{\baselineskip}

\hangindent=0.5cm
\hangafter=1
\textbf{Group 0065: Modern Standard Arabic,  Syrian Arabic, Emirati Arabic, Tunisian Arabic, Algerian Arabic, Moroccan Arabic, Egyptian Arabic, Palestinian Arabic} (\texttt{arb\_arab}: 115 examples, \texttt{apc\_arab\_syri}: 115 examples, \texttt{afb\_arab}: 115 examples, \texttt{aeb\_arab}: 115 examples, \texttt{arq\_arab}: 115 examples, \texttt{ary\_arab}: 115 examples, \texttt{arz\_arab}: 114 examples, \texttt{apc\_arab\_pale}: 115 examples)

\setlength{\parskip}{0pt}
\hangindent=0cm
Manually written by native speakers of eight Arabic dialects (including Modern Standard Arabic). Examples were written by all of the authors to be balanced across locales, and the resulting dataset was translated into each Arabic dialect by the respective native speaker. Domains covered include household, clothing, cooking, hospitality, events, and religion.
\setlength{\parskip}{\baselineskip}

\hangindent=0.5cm
\hangafter=1
\textbf{Group 0066: Galician} (\texttt{glg\_latn}: 109 examples)

\setlength{\parskip}{0pt}
\hangindent=0cm
Manually written and reviewed by native Galician speakers. Approximately half of the dataset covers Galician traditions and seasonal festivities, local customs and folklore, or traditional instruments. Galician websites (e.g. Galician Wikipedia, or local websites) were used to motivate some examples, but none of the content on these sites was used directly.
\setlength{\parskip}{\baselineskip}

\hangindent=0.5cm
\hangafter=1
\textbf{Group 0067: Malayalam} (\texttt{mal\_mlym}: 101 examples)

\setlength{\parskip}{0pt}
\hangindent=0cm
Manually written and reviewed by native Malayalam speakers from different regions of Kerala: one from Muvattupuzha (Idukki and Kottayam dialects), and one from Ottappalam (Palakkad and Thrissur dialects). Examples were written to cover topics specific to Kerala, such as local weather, traditional food recipes, regional flora and fauna, cultural flair, and religious traditions.
\setlength{\parskip}{\baselineskip}

\hangindent=0.5cm
\hangafter=1
\textbf{Group 0068: Persian (Farsi)} (\texttt{pes\_arab}: 100 examples)

\setlength{\parskip}{0pt}
\hangindent=0cm
This dataset was created by native Farsi speakers using a hybrid LLM and manual approach. LLMs were prompted to propose high-level categories and illustrative examples, spanning both everyday knowledge and culturally-specific practices. Based on these examples, the authors either created new samples from scratch inspired by the proposed categories or edited the LLM-generated examples. All examples were reviewed and edited by two native speakers.
\setlength{\parskip}{\baselineskip}

\hangindent=0.5cm
\hangafter=1
\textbf{Group 0069: Hindi, Telugu} (\texttt{hin\_deva}: 179 examples, \texttt{tel\_telu}: 97 examples)

\setlength{\parskip}{0pt}
\hangindent=0cm
This dataset was created by native Hindi and Telugu speakers, using a hybrid LLM and manual approach. First, native speakers wrote a small set of seed examples which were used to prompt Gemini to expand the dataset. Each generated example was reviewed and edited by native speakers. The Hindi portion of the dataset uses Standard Hindi, which is widely understood across Northern India, with many prompts inspired by cultural practices such as food preparation, household activities, and regional crafts. The Telugu portion is based on Standard Telugu, spoken in Telangana and Andhra Pradesh, and it reflects daily life in those regions, from traditional agricultural practices to the handling of clay utensils.
\setlength{\parskip}{\baselineskip}

\pagebreak

\hangindent=0.5cm
\hangafter=1
\textbf{Group 0070: Yemeni Arabic, Egyptian Arabic, Tunisian Arabic, Saudi Arabic, Jordanian Arabic, Lebanese Arabic} (\texttt{acq\_arab}: 100 examples, \texttt{arz\_arab}: 100 examples, \texttt{aeb\_arab}: 99 examples, \texttt{ars\_arab}: 100 examples, \texttt{apc\_arab\_jord}: 100 examples, \texttt{apc\_arab\_leba}: 100 examples)

\setlength{\parskip}{0pt}
\hangindent=0cm
Manually written by native speakers of six Arabic dialects. Examples cover culturally-specific topics such as food, locations, religion, art, games, cultural items, and clothing.
\setlength{\parskip}{\baselineskip}

\hangindent=0.5cm
\hangafter=1
\textbf{Group 0071: Gujarati} (\texttt{guj\_gujr}: 100 examples)

\setlength{\parskip}{0pt}
\hangindent=0cm
This dataset was created by a native Gujarati speaker, using a hybrid LLM and manual approach. ChatGPT was prompted to generate examples, and a native Gujarati speaker manually filtered and edited all examples. Topics include household activities, local festivals, food, school settings, kitchen tools, farm life, animals, seasons, games, common objects, and geography, all reflective of Gujarati customs and environments.
\setlength{\parskip}{\baselineskip}

\hangindent=0.5cm
\hangafter=1
\textbf{Group 0072: Norwegian Bokmål} (\texttt{nob\_latn}: 100 examples)

\setlength{\parskip}{0pt}
\hangindent=0cm
Manually written by a native Norwegian speaker, using Norwegian Bokmål. The dataset covers Norwegian-specific activities, such as the preparation of traditional food dishes and the use of traditional objects.
\setlength{\parskip}{\baselineskip}

\hangindent=0.5cm
\hangafter=1
\textbf{Group 0073: Nepali} (\texttt{npi\_deva}: 201 examples)

\setlength{\parskip}{0pt}
\hangindent=0cm
Manually written and reviewed by native speakers of Nepali, based on topics including household tasks, personal care, outdoor activities, crafts, sports, and recreational pursuits. Another split of this dataset was generated with LLMs and human-verified, but only the human-written examples are included in Global PIQA.
\setlength{\parskip}{\baselineskip}

\hangindent=0.5cm
\hangafter=1
\textbf{Group 0074: Tamil} (\texttt{tam\_taml}: 235 examples)

\setlength{\parskip}{0pt}
\hangindent=0cm
Manually written by native Tamil speakers, focusing on Tamil cooking, including traditional Indian food preparation, ingredients, and terminology.
\setlength{\parskip}{\baselineskip}

\hangindent=0.5cm
\hangafter=1
\textbf{Group 0075: Tamil} (\texttt{tam\_taml}: 113 examples)

\setlength{\parskip}{0pt}
\hangindent=0cm
Manually written and reviewed by native Tamil speakers. Examples cover cultural and traditional dimensions of Sri Lankan life, including food practices, health and safety, religious traditions, rituals and customs, literature and arts, and traditional dress and identity.
\setlength{\parskip}{\baselineskip}

\hangindent=0.5cm
\hangafter=1
\textbf{Group 0076: Malayalam} (\texttt{mal\_mlym}: 100 examples)

\setlength{\parskip}{0pt}
\hangindent=0cm
Manually written by a native Malayalam speaker, and checked by other native speakers. Topics include local culture, cuisine, etiquette, superstitions, religion, and life hacks. Motivation for examples was often drawn from everyday objects in the author's household. Several prompts intentionally illustrate linguistic features unique to Malayalam.
\setlength{\parskip}{\baselineskip}

\hangindent=0.5cm
\hangafter=1
\textbf{Group 0077: Russian} (\texttt{rus\_cyrl}: 92 examples)

\setlength{\parskip}{0pt}
\hangindent=0cm
Manually written by a native Russian speaker, covering topics such as cooking, safety measures, basic physics, and basic computer use. Some questions are designed to be based on Russian culture.
\setlength{\parskip}{\baselineskip}

\hangindent=0.5cm
\hangafter=1
\textbf{Group 0078: Marathi} (\texttt{mar\_deva}: 103 examples)

\setlength{\parskip}{0pt}
\hangindent=0cm
This dataset was created by native Marathi speakers, using a hybrid LLM and manual approach. ChatGPT was prompted to generate examples, and native Marathi speakers manually filtered and edited all examples. Topics include household activities, local festivals, food, school settings, kitchen tools, farm life, animals, seasons, games, common objects, and geography, all reflective of Marathi customs and environments.
\setlength{\parskip}{\baselineskip}

\hangindent=0.5cm
\hangafter=1
\textbf{Group 0079: Bengali, Hindi, Kannada, Tamil, Malayalam} (\texttt{ben\_beng}: 100 examples, \texttt{hin\_deva}: 100 examples, \texttt{kan\_knda}: 97 examples, \texttt{tam\_taml}: 100 examples, \texttt{mal\_mlym}: 100 examples)

\setlength{\parskip}{0pt}
\hangindent=0cm
This dataset was created using LLM generation with human verification by native speakers of Bengali, Hindi, Kannada, Tamil, and Malayalam. LLMs (Gemini 2.5 Pro and Qwen 3) and translation models (MADLAD-400) were used in a multi-stage pipeline to identify topic clusters in English PIQA, to generate localized examples in English (localized to specific Indian states where the respective languages are widely spoken), to translate examples to the respective languages, then to correct any errors in the translations. After this pipeline, native speakers validated all examples.
\setlength{\parskip}{\baselineskip}

\hangindent=0.5cm
\hangafter=1
\textbf{Group 0081: Telugu} (\texttt{tel\_telu}: 93 examples)

\setlength{\parskip}{0pt}
\hangindent=0cm
Manually written and reviewed by native Telugu speakers, using occasional Godavari regional slang. Topics include household activities, food preparation, natural phenomena, and cultural practices.
\setlength{\parskip}{\baselineskip}

\hangindent=0.5cm
\hangafter=1
\textbf{Group 0082: Telugu, Nepali, Hindi} (\texttt{tel\_telu}: 191 examples, \texttt{npi\_deva}: 192 examples, \texttt{hin\_deva}: 192 examples)

\setlength{\parskip}{0pt}
\hangindent=0cm
Manually written and reviewed by native Telugu, Nepali, and Hindi speakers. Embeddings of English translations were used to ensure that no examples were duplicates of English PIQA examples, and Gemini 2.5 Flash was used to verify the correctness of some examples. Posthoc, some examples were modified to incorporate more culturally-specific elements.
\setlength{\parskip}{\baselineskip}

\hangindent=0.5cm
\hangafter=1
\textbf{Group 0083: Hindi} (\texttt{hin\_deva}: 101 examples)

\setlength{\parskip}{0pt}
\hangindent=0cm
Manually written and reviewed by native Hindi speakers, focusing on everyday scenarios. Topics include food and cooking, household chores, health and safety, festivals and traditions, travel, technology and gadgets, environment and hygiene, personal care, and emergency situations.
\setlength{\parskip}{\baselineskip}

\hangindent=0.5cm
\hangafter=1
\textbf{Group 0085: Hindi, Kannada, Telugu, Malayalam} (\texttt{hin\_deva}: 97 examples, \texttt{kan\_knda}: 120 examples, \texttt{tel\_telu}: 100 examples, \texttt{mal\_mlym}: 111 examples)

\setlength{\parskip}{0pt}
\hangindent=0cm
Manually written and reviewed by native speakers of Hindi, Kannada, Telugu, and Malayalam. Examples were written to be relevant to speakers of the respective language, covering topics such as food, clothing, household items, everyday life, festivals, and traditions. GPT-4 was used initially to generate examples for inspiration, but all examples in the final dataset are manually written.
\setlength{\parskip}{\baselineskip}

\hangindent=0.5cm
\hangafter=1
\textbf{Group 0086: Greek} (\texttt{ell\_grek}: 92 examples)

\setlength{\parskip}{0pt}
\hangindent=0cm
This dataset was manually constructed by a native Greek speaker, by navigating Greek websites on the internet, searching for sentences about a given topic, then adapting the sentences for the task. Topics include puzzles and riddles, household, cooking and recipes, driving, gardening, DIY, sports, construction, vacation, spatiotemporal orientation, and dance.
\setlength{\parskip}{\baselineskip}

\hangindent=0.5cm
\hangafter=1
\textbf{Group 0087: Turkish} (\texttt{tur\_latn}: 141 examples)

\setlength{\parskip}{0pt}
\hangindent=0cm
Manually written by native Turkish speakers, motivated by Turkish content such as food blogs, household advice websites, and health institution pages. All examples were manually verified by several Turkish speakers.
\setlength{\parskip}{\baselineskip}

\hangindent=0.5cm
\hangafter=1
\textbf{Group 0088: Yoruba, Nigerian Pidgin (Naijá)} (\texttt{yor\_latn}: 199 examples, \texttt{pcm\_latn}: 191 examples)

\setlength{\parskip}{0pt}
\hangindent=0cm
Manually written and reviewed by native Yoruba and Nigerian Pidgin speakers. First, the authors compiled a list of everyday physical items relevant to both cultures, inspired by online videos, language dictionaries, and social media. Then, realistic scenarios were manually written for different items, and these prompts were used as the basis for examples.
\setlength{\parskip}{\baselineskip}

\hangindent=0.5cm
\hangafter=1
\textbf{Group 0089: Marwari, Marathi} (\texttt{mar\_deva}: 124 examples, \texttt{rwr\_deva}: 124 examples)

\setlength{\parskip}{0pt}
\hangindent=0cm
Manually written and reviewed by native Marathi and Marwari speakers, covering culturally-specific topics such as home, cooking, farming and rural contexts, weather, and desert travel.
\setlength{\parskip}{\baselineskip}

\hangindent=0.5cm
\hangafter=1
\textbf{Group 0090: Telugu} (\texttt{tel\_telu}: 43 examples)

\setlength{\parskip}{0pt}
\hangindent=0cm
Manually written and reviewed by native Telugu speakers, using Kosta Andhra Telugu, a dialect spoken in coastal Andhra Pradesh, India. Examples in the dataset cover local festivals and traditional foods.
\setlength{\parskip}{\baselineskip}

\hangindent=0.5cm
\hangafter=1
\textbf{Group 0091: Tamil} (\texttt{tam\_taml}: 226 examples)

\setlength{\parskip}{0pt}
\hangindent=0cm
Manually written and reviewed by native Tamil speakers, after an initial attempt to use LLMs produced examples that were often generic, obvious, or culturally inaccurate. In the final dataset, all examples are either entirely manually written or substantially rewritten and refined from a primitive LLM-generated example. Culturally-specific topics include traditional rituals, literature and history, agrarian and folk wisdom, and art.
\setlength{\parskip}{\baselineskip}

\pagebreak

\hangindent=0.5cm
\hangafter=1
\textbf{Group 0092: Bengali} (\texttt{ben\_beng}: 79 examples)

\setlength{\parskip}{0pt}
\hangindent=0cm
Manually written by a native Bengali speaker, and reviewed by other native speakers. The dataset uses standard colloquial Bengali as commonly spoken in Kolkata, India, and it includes references to local customs, food, holidays and traditions, and household objects.
\setlength{\parskip}{\baselineskip}

\hangindent=0.5cm
\hangafter=1
\textbf{Group 0093: Slovak, Šariš Slovak} (\texttt{slk\_latn}: 100 examples, \texttt{slk\_latn\_sari}: 100 examples)

\setlength{\parskip}{0pt}
\hangindent=0cm
Manually written by native speakers of Slovak and the Šariš dialect of Slovak. Examples were inspired by content on DIY and home improvement sites in Slovak, but no content was copied directly.
\setlength{\parskip}{\baselineskip}

\hangindent=0.5cm
\hangafter=1
\textbf{Group 0094: Assamese, Bengali, Hindi, Malayalam, Manipuri} (\texttt{asm\_beng}: 195 examples, \texttt{ben\_beng}: 498 examples, \texttt{hin\_deva}: 376 examples, \texttt{mal\_mlym}: 212 examples, \texttt{mni\_mtei}: 114 examples)

\setlength{\parskip}{0pt}
\hangindent=0cm
Manually written and reviewed by native speakers of Assamese, Bengali, Hindi, Malayalam, and Manipuri, covering everyday topics such as food, rituals, tools, climate, and household practices. Additional manual verification is in progress for Maithili, Orya, and Telugu datasets.
\setlength{\parskip}{\baselineskip}

\hangindent=0.5cm
\hangafter=1
\textbf{Group 0095: Italian} (\texttt{ita\_latn}: 105 examples)

\setlength{\parskip}{0pt}
\hangindent=0cm
Manually written and reviewed by native Italian speakers, covering culturally-specific topics such as local foods, artisanal products, domestic practices, and folklore.
\setlength{\parskip}{\baselineskip}

\hangindent=0.5cm
\hangafter=1
\textbf{Group 0096: Thai} (\texttt{tha\_thai}: 97 examples)

\setlength{\parskip}{0pt}
\hangindent=0cm
Manually written by a native Thai speaker. Inspired by browsing the internet in Thai, some examples cover local landmarks, art, cooking, and customs that are unique to Thailand.
\setlength{\parskip}{\baselineskip}

\hangindent=0.5cm
\hangafter=1
\textbf{Group 0097: Hindi, Marathi, Tamil} (\texttt{hin\_deva}: 67 examples, \texttt{mar\_deva}: 85 examples, \texttt{tam\_taml}: 150 examples)

\setlength{\parskip}{0pt}
\hangindent=0cm
Manually written and reviewed by native speakers of Hindi, Marathi, and Tamil, covering culturally-relevant everyday scenarios in Indic contexts, such as food preparation, household chores, and electronic device usage. Examples underwent extensive validation and rewriting, including reading examples aloud to parents, grandparents, and younger relatives.
\setlength{\parskip}{\baselineskip}

\hangindent=0.5cm
\hangafter=1
\textbf{Group 0098: Hindi} (\texttt{hin\_deva}: 83 examples)

\setlength{\parskip}{0pt}
\hangindent=0cm
Manually written by a native Hindi speaker, and reviewed by another native speaker. Examples were drawn from
diverse domains such as traditional Indian games, handicrafts, festivals, musical instruments, and everyday life.
\setlength{\parskip}{\baselineskip}

\hangindent=0.5cm
\hangafter=1
\textbf{Group 0099: Czech} (\texttt{ces\_latn}: 195 examples)

\setlength{\parskip}{0pt}
\hangindent=0cm
Manually written and reviewed by native Czech speakers, covering domains such as everyday activities, cooking, household tasks, and activities related to traditional Czech customs or sayings. Some examples use Moravian and Silesian dialects, or contemporary Gen Z and Gen Alpha slang (e.g. ``skibidi'' and ``6-7''). For examples using slang or dialects, the authors consulted external collaborators from those demographic groups to ensure correct usage. Examples were passed into GPT-5 and Claude Opus 4.1 for edits, and a small number of examples were generated directly by the LLMs themselves; all examples underwent human validation by multiple native speakers.
\setlength{\parskip}{\baselineskip}

\hangindent=0.5cm
\hangafter=1
\textbf{Group 0100: Thai} (\texttt{tha\_thai}: 97 examples)

\setlength{\parskip}{0pt}
\hangindent=0cm
Manually written by a native Thai speaker, using the central Thai dialect. Examples cover specific Thai knowledge, such as Muay Thai movements.
\setlength{\parskip}{\baselineskip}

\hangindent=0.5cm
\hangafter=1
\textbf{Group 0101: Sinhala} (\texttt{sin\_sinh}: 110 examples)

\setlength{\parskip}{0pt}
\hangindent=0cm
Manually written and reviewed by native Sinhala speakers, covering domains such as literature, religion, mythology, sports, food, and history, primarily in a Sri Lankan context.
\setlength{\parskip}{\baselineskip}

\hangindent=0.5cm
\hangafter=1
\textbf{Group 0102: Turkish, Azerbeijani, Kyrgyz} (\texttt{tur\_latn}: 135 examples, \texttt{azj\_latn}: 119 examples, \texttt{kir\_cyrl}: 113 examples)

\setlength{\parskip}{0pt}
\hangindent=0cm
This dataset was written and reviewed by native speakers of Turkish, Azerbaijani, and Kyrgyz. Topics include household routines, cooking, driving, and seasonal conditions, along with everyday and culturally-specific items. Some examples in Turkish were initially generated using GPT-5, but many Turkish examples are fully original, and all examples were verified by native speakers. LLMs were not used for Azerbaijani or Kyrgyz; for example, for Azerbaijani, trials with GPT-5 and Gemini 2.5 Pro produced poor quality samples.
\setlength{\parskip}{\baselineskip}

\hangindent=0.5cm
\hangafter=1
\textbf{Group 0103: Tamil} (\texttt{tam\_taml}: 688 examples)

\setlength{\parskip}{0pt}
\hangindent=0cm
Manually written and reviewed by native speakers of Tamil, using Sri Lankan Tamil and covering domains such as domestic chores, culinary practices, agriculture, and traditional artifacts. Examples were deduplicated with n-grams and SBERT embeddings. When evaluated by humans, four  native speakers agreed unanimously on the label for 95\% of examples.
\setlength{\parskip}{\baselineskip}

\hangindent=0.5cm
\hangafter=1
\textbf{Group 0104: Korean} (\texttt{kor\_hang}: 181 examples)

\setlength{\parskip}{0pt}
\hangindent=0cm
This dataset was constructed by native Korean speakers using a hybrid LLM and manual approach. Using a multi-stage pipeline, LLMs were given Korean-specific seed scenarios to (1) generate examples, (2) validate the questions, (3) validate the solutions, (4) generate distractor solutions, and (5) validate distractors. Finally, examples were deduplicated, and biased answers (e.g. examples that could be solved with simple heuristics) were removed. All final examples were validated by a native Korean speaker.
\setlength{\parskip}{\baselineskip}

\hangindent=0.5cm
\hangafter=1
\textbf{Group 0105: Kinyarwanda} (\texttt{kin\_latn}: 108 examples)

\setlength{\parskip}{0pt}
\hangindent=0cm
Manually written by a native Kinyarwanda speaker, and reviewed by another native speaker, using the standard dialect spoken in education and media. Examples cover everyday scenarios such as household activities, tools and objects, food, transportation, and weather.
\setlength{\parskip}{\baselineskip}

\hangindent=0.5cm
\hangafter=1
\textbf{Group 0106: Swahili} (\texttt{swh\_latn}: 172 examples)

\setlength{\parskip}{0pt}
\hangindent=0cm
Manually written by a native Swahili speaker, covering a variety of everyday contexts.
\setlength{\parskip}{\baselineskip}

\hangindent=0.5cm
\hangafter=1
\textbf{Group 0107: Central Kurdish} (\texttt{ckb\_arab}: 100 examples)

\setlength{\parskip}{0pt}
\hangindent=0cm
Manually written by a native Kurdish speaker, using Central Kurdish (also known as Sorani). Examples focus on village life and traditional practices (e.g. cooking, handicrafts, agriculture, animal husbandry, and customs), domains where Kurdish possesses a rich and nuanced vocabulary.
\setlength{\parskip}{\baselineskip}

\hangindent=0.5cm
\hangafter=1
\textbf{Group 0108: Hungarian} (\texttt{hun\_latn}: 105 examples)

\setlength{\parskip}{0pt}
\hangindent=0cm
Manually written and reviewed by native Hungarian speakers, covering a variety of physical phenomena and incorporating Hungarian cultural context.
\setlength{\parskip}{\baselineskip}

\hangindent=0.5cm
\hangafter=1
\textbf{Group 0109: Turkish} (\texttt{tur\_latn}: 99 examples)

\setlength{\parskip}{0pt}
\hangindent=0cm
Manually written by a native Turkish speaker, with some sentences adapted from online food recipes.
\setlength{\parskip}{\baselineskip}

\hangindent=0.5cm
\hangafter=1
\textbf{Group 0110: Russian} (\texttt{rus\_cyrl}: 100 examples)

\setlength{\parskip}{0pt}
\hangindent=0cm
Manually written and reviewed by two native Russian speakers, covering everyday scenarios. Some examples cover culturally-specific holidays or foods.
\setlength{\parskip}{\baselineskip}

\hangindent=0.5cm
\hangafter=1
\textbf{Group 0112: Javanese} (\texttt{jav\_latn}: 120 examples)

\setlength{\parskip}{0pt}
\hangindent=0cm
One native Javanese speaker contracted five other annotators through Prolific at a rate of 8 GBP per hour, which is significantly above the minimum hourly wage in Indonesia. Many examples were written to be culturally-specific, covering local music, food, nature, and daily life. Generally, this dataset uses the Ngoko register, or casual language in Javanese. Although a standardized writing guideline exists for Javanese, it is not universally followed, and there is substantial variation in orthography and spelling. Annotators were allowed to write in the form they naturally used, to better capture authentic language use. The final examples were reviewed by the primary author of this dataset.
\setlength{\parskip}{\baselineskip}

\hangindent=0.5cm
\hangafter=1
\textbf{Group 0113: Georgian} (\texttt{kat\_geor}: 100 examples)

\setlength{\parskip}{0pt}
\hangindent=0cm
Manually written and reviewed by native Georgian speakers, covering everyday knowledge and activities. Some examples drew inspiration from the Georgian book, ``Imagination and Skillful Hands'' by Neli Okropiridze, which offers tips and tricks for a range of DIY projects and was once widely used in the Georgian community.
\setlength{\parskip}{\baselineskip}

\pagebreak

\hangindent=0.5cm
\hangafter=1
\textbf{Group 0114: Burushaski} (\texttt{bsk\_arab}: 100 examples)

\setlength{\parskip}{0pt}
\hangindent=0cm
Manually written by a native Burushaski speaker, using the Yasin dialect. All examples were checked for grammatical correctness, cultural relevance, and physical commonsense validity.
\setlength{\parskip}{\baselineskip}

\hangindent=0.5cm
\hangafter=1
\textbf{Group 0115: Peruvian Spanish} (\texttt{spa\_latn\_peru}: 102 examples)

\setlength{\parskip}{0pt}
\hangindent=0cm
This dataset was manually compiled by native Spanish speakers. Sentences were adapted from naturally occurring speech among the dataset authors' family and friends. Some examples were drawn from public-interest topics in Lima, Peru, including local traditions or the conduct of public officials. Any names, addresses, or direct identifiers were removed and replaced with more generic placeholders. To capture authentic language usage, tense and punctuation were not standardized but instead left reflective of colloquial speech.
\setlength{\parskip}{\baselineskip}

\hangindent=0.5cm
\hangafter=1
\textbf{Group 0116: Russian} (\texttt{rus\_cyrl}: 53 examples)

\setlength{\parskip}{0pt}
\hangindent=0cm
This small dataset was manually written and reviewed by native Russian speakers from the South Ural Mountains region of Russia. Several examples are designed to test local commonsense knowledge.
\setlength{\parskip}{\baselineskip}

\hangindent=0.5cm
\hangafter=1
\textbf{Group 0117: Hawaiian (`Ōlelo Hawai`i)} (\texttt{haw\_latn}: 100 examples)

\setlength{\parskip}{0pt}
\hangindent=0cm
Manually written by second-language `ōlelo Hawai`i speakers, and verified by native speakers. Examples cover a wide range of scenarios, including contexts specific to Hawai`i, the Hawaiian language, and Hawaiian culture, as well as everyday situations. All Hawaiian text was written in modern orthography, including both the `okina and kahakō. Relevant to anyone using this dataset, the dataset authors note the distinction between no`ono`o Hawai`i (Hawaiian ways of thinking) and no`ono`o Haole (foreign ways of thinking) as applied to NLP, where ``data representation choices risk importing external frameworks. Preserving no`ono`o Hawai`i ensures that datasets and computational models reflect culturally grounded perspectives, maintaining authenticity and integrity in the development of Hawaiian language technologies''.
\setlength{\parskip}{\baselineskip}

\hangindent=0.5cm
\hangafter=1
\textbf{Group 0118: Portuguese (European)} (\texttt{por\_latn\_port}: 105 examples)

\setlength{\parskip}{0pt}
\hangindent=0cm
Manually written and reviewed by native European Portuguese speakers, with many examples covering Portuguese culture (e.g. references to festivities, holidays, and the preparation of traditional dishes). Two native speakers evaluated the dataset without access to labels, achieving accuracies of 90.7\% and 95.4\% respectively.
\setlength{\parskip}{\baselineskip}

\hangindent=0.5cm
\hangafter=1
\textbf{Group 0119: Algerian Arabic, Moroccan Arabic} (\texttt{arq\_arab}: 209 examples, \texttt{ary\_arab}: 198 examples)

\setlength{\parskip}{0pt}
\hangindent=0cm
This dataset was crowdsourced from native Algerian and Moroccan (Darija) Arabic speakers. All examples were checked by other native speakers for naturalness, correctness, and cultural relevance. Contributors and annotators participated voluntarily without monetary compensation. Recruitment occurred via open community channels; participants gave informed consent, could withdraw at any time, and were not subject to coercion or undue influence. No personally identifiable information was collected. Across three annotators, average pairwise agreement on labels was over 95\% (Cohen's kappa > 0.90 for all pairs).
\setlength{\parskip}{\baselineskip}

\hangindent=0.5cm
\hangafter=1
\textbf{Group 0120: Amharic} (\texttt{amh\_ethi}: 141 examples)

\setlength{\parskip}{0pt}
\hangindent=0cm
Approximately half of this dataset was manually written by a native Amharic speaker; the other half was generated by using Gemini 2.5 to expand the size of the dataset. All examples were then verified by multiple native speakers. Examples focus on the topics of sports, culture, history, politics, and education.
\setlength{\parskip}{\baselineskip}

\hangindent=0.5cm
\hangafter=1
\textbf{Group 0121: German} (\texttt{deu\_latn}: 100 examples)

\setlength{\parskip}{0pt}
\hangindent=0cm
Manually written by a native German speaker, covering culturally-specific topics such as food and customs that might not be well known outside of Germany. ChatGPT was used to help double-check grammar and spelling, but not to generate examples.
\setlength{\parskip}{\baselineskip}

\hangindent=0.5cm
\hangafter=1
\textbf{Group 0122: German} (\texttt{deu\_latn}: 26 examples)

\setlength{\parskip}{0pt}
\hangindent=0cm
This small dataset was manually written by a native German speaker, covering topics such as sports, household, gardening, and entertainment.
\setlength{\parskip}{\baselineskip}

\pagebreak

\hangindent=0.5cm
\hangafter=1
\textbf{Group 0123: English (USA and UK)} (\texttt{eng\_latn}: 104 examples)

\setlength{\parskip}{0pt}
\hangindent=0cm
This dataset was obtained by filtering the English PIQA test set to approximately 100 high-quality examples. Examples were excluded if they contained typos or nonsensical answer choices; some examples were modified to correct these errors. Many examples were selected based on cultural relevance to English-speaking contexts in the United States of America or the United Kingdom (e.g. US Thanksgiving, or American football). The resulting dataset was validated by another native English speaker.
\setlength{\parskip}{\baselineskip}

\hangindent=0.5cm
\hangafter=1
\textbf{Group 0124: Amharic} (\texttt{amh\_ethi}: 119 examples)

\setlength{\parskip}{0pt}
\hangindent=0cm
Manually written by a native Amharic speaker, and validated by other native speakers. Examples cover everyday contexts in Ethiopian society, including traditions, customs, food, history, and proverbs.
\setlength{\parskip}{\baselineskip}

\hangindent=0.5cm
\hangafter=1
\textbf{Group 0125: Bambara} (\texttt{bam\_latn}: 111 examples)

\setlength{\parskip}{0pt}
\hangindent=0cm
This dataset was compiled by native Bambara speakers. Some examples were based on content from French quizzes on technical knowledge, translated into Bambara by professional translators. Other examples were written to be culturally-specific to Bambara-speaking contexts. All examples were refined and validated by native Bambara speakers.
\setlength{\parskip}{\baselineskip}

\hangindent=0.5cm
\hangafter=1
\textbf{Group 0126: Peninsular Spanish} (\texttt{spa\_latn\_spai}: 101 examples)

\setlength{\parskip}{0pt}
\hangindent=0cm
Manually written by a native Spanish speaker, using central-northern Peninsular Spanish (e.g. as spoken in Madrid and the interior of Castilla y León). Examples cover culturally-specific foods, customs, and domestic practices.
\setlength{\parskip}{\baselineskip}

\hangindent=0.5cm
\hangafter=1
\textbf{Group 0127: Eastern Armenian} (\texttt{hye\_armn}: 102 examples)

\setlength{\parskip}{0pt}
\hangindent=0cm
Manually written by an Armenian speaker, and checked by a native speaker. Prompts were first outlined in English then translated to Eastern Armenian. Topics include cutlery and tableware, fabrics and clothing, laundry, and cooking. A small number of examples are specific to Armenian culture.
\setlength{\parskip}{\baselineskip}

\hangindent=0.5cm
\hangafter=1
\textbf{Group 0128: Lithuanian} (\texttt{lit\_latn}: 74 examples)

\setlength{\parskip}{0pt}
\hangindent=0cm
Manually written by a native speaker of Lithuanian, with examples constructed using a mix of domain expertise and simple Lithuania-related questions. GPT-5 was used to brainstorm ideas, but not to generate examples.
\setlength{\parskip}{\baselineskip}

\hangindent=0.5cm
\hangafter=1
\textbf{Group 0129: Lithuanian} (\texttt{lit\_latn}: 100 examples)

\setlength{\parskip}{0pt}
\hangindent=0cm
Examples in this dataset were generated based on Wikipedia articles using GPT-5, then manually rephrased and checked by two native speakers of Lithuanian. Topics include traditional Lithuanian food, traditions, places, and literature.
\setlength{\parskip}{\baselineskip}

\hangindent=0.5cm
\hangafter=1
\textbf{Group 0130: Zulu} (\texttt{zul\_latn}: 100 examples)

\setlength{\parskip}{0pt}
\hangindent=0cm
Manually written by a native speaker of isiZulu, with examples written to reflect everyday scenarios and local cultural practices.
\setlength{\parskip}{\baselineskip}

\hangindent=0.5cm
\hangafter=1
\textbf{Group 0131: Kazakh} (\texttt{kaz\_cyrl}: 100 examples)

\setlength{\parskip}{0pt}
\hangindent=0cm
Manually written by a native speaker of Kazakh, using the Northeastern Kazakh dialect, and including some specific words that are commonly used in Karaganda city. Examples cover culturally-specific topics, including food, drinks, music, customs, animals, games, history, architecture and monuments, weather, nature, clothing, and jewelry.
\setlength{\parskip}{\baselineskip}

\hangindent=0.5cm
\hangafter=1
\textbf{Group 0132: Bosnian} (\texttt{bos\_latn}: 145 examples)

\setlength{\parskip}{0pt}
\hangindent=0cm
Manually written by a native Bosnian speaker, using the Ijekavian standard. The dataset covers regionally salient vocabulary and scenarios, including cooking, household tasks, nature, and religious and social customs.
\setlength{\parskip}{\baselineskip}

\hangindent=0.5cm
\hangafter=1
\textbf{Group 0133: Kinyarwanda} (\texttt{kin\_latn}: 99 examples)

\setlength{\parskip}{0pt}
\hangindent=0cm
Manually written and reviewed by native Kinyarwanda speakers. Examples cover everyday domains such as everyday objects, weather, folklore, and literature. The dataset is written in standard Kinyarwanda, without dialectal variations such as those spoken in the northern and southern provinces of Rwanda.
\setlength{\parskip}{\baselineskip}

\hangindent=0.5cm
\hangafter=1
\textbf{Group 0134: Peninsular Spanish, Mexican Spanish} (\texttt{spa\_latn\_spai}: 100 examples, \texttt{spa\_latn\_mexi}: 100 examples)

\setlength{\parskip}{0pt}
\hangindent=0cm
Manually written by native Spanish speakers, covering a variety of subtypes of physical commonsense reasoning. Examples reference local foods, places, traditions, architecture, and everyday objects and tasks in Spain and Mexico (for Peninsular and Mexican Spanish respectively). The Peninsular and Mexican Spanish datasets differ at the topic, lexical, and syntactic levels, to reflect differences between the two dialects. All examples in the two datasets were verified and edited by a native Spanish speaker living in Spain or Mexico respectively.
\setlength{\parskip}{\baselineskip}

\hangindent=0.5cm
\hangafter=1
\textbf{Group 0135: Ekpeye} (\texttt{ekp\_latn}: 100 examples)

\setlength{\parskip}{0pt}
\hangindent=0cm
Manually written by a native Ekpeye speaker, with topics covering everyday life, local Nigerian foods, and local customs.
\setlength{\parskip}{\baselineskip}

\hangindent=0.5cm
\hangafter=1
\textbf{Group 0136: Serbian Torlak} (\texttt{srp\_cyrl\_torl}: 100 examples, \texttt{srp\_latn\_torl}: 100 examples)

\setlength{\parskip}{0pt}
\hangindent=0cm
Manually written by a native speaker of Torlak Serbian, in collaboration with Group 0022. Authors attempted to include culturally-relevant examples, often motivated by everyday objects, life hacks, recipes, and/or assembly manuals in the language.
\setlength{\parskip}{\baselineskip}

\hangindent=0.5cm
\hangafter=1
\textbf{Group 0137: Zarma} (\texttt{dje\_latn}: 102 examples)

\setlength{\parskip}{0pt}
\hangindent=0cm
Examples in this dataset were generated using Mistral and Gemini, following the InstructLR pipeline \citep{instructlr-2026}, then verified and cross-checked by native Zarma speakers. The InstructLR pipeline consists of topic generation and example generation in French, machine translation to Zarma, then validation and correction by native Zarma speakers. Topics include local foods, customs, and everyday life.
\setlength{\parskip}{\baselineskip}

\hangindent=0.5cm
\hangafter=1
\textbf{Group 0138: Odia} (\texttt{ory\_orya}: 100 examples)

\setlength{\parskip}{0pt}
\hangindent=0cm
Manually written by a native Odia speaker. The dataset covers Odisha-centric contexts, including landscape and tourism, religious practices, storm-impacted agriculture and flora, traditional food preparation, folk theater instruments and equipment, and Odissi dance and drama.
\setlength{\parskip}{\baselineskip}

\hangindent=0.5cm
\hangafter=1
\textbf{Group 0139: Southern Balochi} (\texttt{bcc\_arab}: 100 examples)

\setlength{\parskip}{0pt}
\hangindent=0cm
Manually written and reviewed by native speakers of Southern Balochi. Over half of examples are grounded in Balochi cultural, geographic, and social contexts, such as local livelihoods and coastal/desert environments.
\setlength{\parskip}{\baselineskip}

\hangindent=0.5cm
\hangafter=1
\textbf{Group 0140: Classical Sanskrit} (\texttt{cls\_deva}: 111 examples)

\setlength{\parskip}{0pt}
\hangindent=0cm
Manually written and reviewed by contributors for Sanskrit. Examples cover themes such as Vedic astrology, Ayurvedic science for traditional medicine, Hindu epics, military history of the Maratha Empire, classical dance forms such as Bharatanatyam and Garba, temple music (e.g. instruments such as the Veena and Mridangam), and traditional agricultural village life. All examples use formal classical Sanskrit, checked by multiple contributors for Sanskrit.
\setlength{\parskip}{\baselineskip}

\hangindent=0.5cm
\hangafter=1
\textbf{Group 0141: Odia} (\texttt{ory\_orya}: 116 examples)

\setlength{\parskip}{0pt}
\hangindent=0cm
Manually written by a native Odia speaker. The majority of examples cover contexts relevant to everyday life and society in Odisha, including household life, social interactions, art, cuisine, and cultural events such as festivals.
\setlength{\parskip}{\baselineskip}

\hangindent=0.5cm
\hangafter=1
\textbf{Group 0142: Danish} (\texttt{dan\_latn}: 102 examples)

\setlength{\parskip}{0pt}
\hangindent=0cm
Manually written and reviewed by native Danish speakers from the Central Denmark Region and the Capital Region of Denmark. Examples were written to include hard-to-translate terms, unique phrases, cultural objects, and biking culture.
\setlength{\parskip}{\baselineskip}

\hangindent=0.5cm
\hangafter=1
\textbf{Group 0143: Romanian} (\texttt{ron\_latn}: 99 examples)

\setlength{\parskip}{0pt}
\hangindent=0cm
Manually written by a native Romanian speaker. Examples cover Romanian traditions, dancing, and culinary specifics, with an emphasis on the Oltenia region of Romania.
\setlength{\parskip}{\baselineskip}

\pagebreak

\hangindent=0.5cm
\hangafter=1
\textbf{Group 0144: Slovenian Prlekija} (\texttt{slv\_latn\_prle}: 100 examples)

\setlength{\parskip}{0pt}
\hangindent=0cm
Manually written by a native speaker of the Prleščina (Prlekija) dialect of Slovenian, in collaboration with Group 0022. Authors attempted to include culturally-relevant examples, often motivated by everyday objects, life hacks, recipes, and/or assembly manuals in the language.
\setlength{\parskip}{\baselineskip}

\hangindent=0.5cm
\hangafter=1
\textbf{Group 0145: Basque} (\texttt{eus\_latn}: 100 examples)

\setlength{\parskip}{0pt}
\hangindent=0cm
Manually written by a native Basque speaker, using standard Basque (``Euskara Batua''). Examples cover culturally relevant domains such as gastronomy, local traditions, regional festivities, rural sports, geography, music, and folklore. Some examples were curated and adapted from online sources via targeted searches (e.g. rural sports and recipes). LLMs were used only for rephrasing a small number of examples, and for correcting typos.
\setlength{\parskip}{\baselineskip}

\hangindent=0.5cm
\hangafter=1
\textbf{Group 0146: Plateau Malagasy} (\texttt{plt\_latn}: 100 examples)

\setlength{\parskip}{0pt}
\hangindent=0cm
Manually written by a native Malagasy speaker, with influence from Merina and Betsileo Malagasy dialects. Examples were written to cover everyday life in Madagascar, including traditions (e.g. Famadihana), food, and customs.
\setlength{\parskip}{\baselineskip}

\hangindent=0.5cm
\hangafter=1
\textbf{Group 0147: Batak Karo, Sundanese} (\texttt{btx\_latn}: 100 examples, \texttt{sun\_latn}: 102 examples)

\setlength{\parskip}{0pt}
\hangindent=0cm
Manually written and reviewed by native speakers of Batak Karo and Sundanese, with each example written in a natural and idiomatic manner in the language. Topics include local foods, customs, music, and traditions.
\setlength{\parskip}{\baselineskip}

\hangindent=0.5cm
\hangafter=1
\textbf{Group 0148: Hindi} (\texttt{hin\_latn}: 100 examples)

\setlength{\parskip}{0pt}
\hangindent=0cm
This dataset was manually transliterated into Latin script from the Global PIQA non-parallel split for Hindi (Devanagari script), using the style used in colloquial digital communication. English words written in Devanagari were transliterated using their standard English spellings (e.g. ``computer'') to maintain the natural flow of Romanized Hindi text.
\setlength{\parskip}{\baselineskip}

\hangindent=0.5cm
\hangafter=1
\textbf{Group 0149: Odia} (\texttt{ory\_orya}: 104 examples)

\setlength{\parskip}{0pt}
\hangindent=0cm
Manually written by a native Odia speaker. Examples cover activities, tools, and environmental contexts that are characteristic of Odia households, along with local traditions and regional practices.
\setlength{\parskip}{\baselineskip}

\hangindent=0.5cm
\hangafter=1
\textbf{Group 0150: Haryanvi, Shekhawati, Braj} (\texttt{bgc\_deva}: 100 examples, \texttt{swv\_deva}: 100 examples, \texttt{bra\_deva}: 100 examples)

\setlength{\parskip}{0pt}
\hangindent=0cm
This dataset was translated from examples from Group 0058. Each example was machine translated then verified by a native speaker of Haryanvi, Braj, or Shekhawati. The original examples were adapted by Group 0058 from reasoning textbooks in English and Hindi, with examples written to reflect India-specific cultural contexts.
\setlength{\parskip}{\baselineskip}

\hangindent=0.5cm
\hangafter=1
\textbf{Group 0151: Sinhala, Telugu, Tamil, Kannada} (\texttt{sin\_latn}: 100 examples, \texttt{tel\_latn}: 100 examples, \texttt{tam\_latn}: 100 examples, \texttt{kan\_latn}: 100 examples)

\setlength{\parskip}{0pt}
\hangindent=0cm
This dataset was transliterated into Latin script from the Global PIQA non-parallel splits for Sinhala, Telugu, Tamil, and Kannada, using Gemini 3.0 Flash. The transliterations were then manually verified by native speakers of the respective languages.
\setlength{\parskip}{\baselineskip}

\section{Declaration of AI Usage} \label{app:ai-usage}

No AI was used in the writing of this paper.
A minority of contributors used AI in the process of creating examples for the non-parallel split of Global PIQA, which we discuss in \S\ref{sec:construction-methods}.
In the final non-parallel dataset, less than 5\% of examples were originally written with the help of LLMs, and all examples have been human-verified.
No AI was used to generate the original English version of the parallel split of Global PIQA.
AI was used for initial machine translations of the parallel split, but these translations have all been human-verified, as discussed in \S\ref{sec:parallel}.
AI was used to assist in basic tasks when processing and visualizing model evaluation results for \S\ref{sec:results}.

\newpage
\end{document}